\documentclass[sigconf,preprint]{acmart}

\usepackage{amsmath}
\usepackage{algorithm}
\usepackage{algorithmic}
\usepackage{amsfonts}
\usepackage{enumitem}
\usepackage{MnSymbol} 
\usepackage{bm}
\usepackage{dsfont}
\usepackage{multirow}
\usepackage{adjustbox}
\usepackage{diagbox}
\usepackage{pifont}
\usepackage{xcolor}
\usepackage{xspace}
\usepackage{colortbl}
\usepackage{balance} 
\usepackage[normalem]{ulem}
\usepackage{booktabs}
\usepackage{multirow}
\usepackage{makecell}
\usepackage{enumitem}
\usepackage{graphicx}
\usepackage{subcaption}
\usepackage{placeins}
\usepackage[table]{xcolor}
\newcommand{\m}{SynEnergy\xspace}

\AtBeginDocument{%
  }

\copyrightyear{2027}
\acmYear{2027}
\setcopyright{cc}
\setcctype{by}
\acmDOI{}
\begin{document}

\title[SynEnergy]{SynEnergy: Anomaly Semantic-Guided Diffusion for Synthetic Energy Data Generation}

\author{Lin Jiang}
\affiliation{
  \department{Department of Computer Science}
  \institution{Florida State University}
  \city{Tallahassee}
  \state{Florida}
  \country{USA}
}
\email{lin.jiang@fsu.edu}

\author{Dahai Yu}
\affiliation{
  \department{Department of Computer Science}
  \institution{Florida State University}
  \city{Tallahassee}
  \state{Florida}
  \country{USA}
}
\email{dahai.yu@fsu.edu}

\author{Ravikumar Gelli}
\affiliation{
  \department{Department of Electrical and Computer Engineering}
  \institution{Florida State University}
  \city{Tallahassee}
  \state{Florida}
  \country{USA}
}
\email{rgelli@fsu.edu}

\author{Guang Wang}
\affiliation{
  \department{Department of Computer Science}
  \institution{Florida State University}
  \city{Tallahassee}
  \state{Florida}
  \country{USA}
}
\email{guang@cs.fsu.edu}

% \begin{CCSXML}
% <ccs2012>
%    <concept>
%        <concept_id>10002951.10003227.10003236</concept_id>
%        <concept_desc>Information systems~Spatial-temporal systems</concept_desc>
%        <concept_significance>500</concept_significance>
%        </concept>
%    <concept>
%        <concept_id>10002951.10003227.10003351</concept_id>
%        <concept_desc>Information systems~Data mining</concept_desc>
%        <concept_significance>500</concept_significance>
%        </concept>
%  </ccs2012>
% \end{CCSXML}

% \ccsdesc[500]{Information systems~Spatial-temporal data}
% \ccsdesc[500]{Information systems~Data mining}


\begin{abstract}
Fine-grained energy consumption data are essential for applications such as demand forecasting, demand response planning, and grid reliability assessment. However, access to such data is often restricted by privacy concerns and data-sharing constraints, motivating growing interest in synthetic energy data generation. Although existing methods can reproduce overall consumption distributions and recurring temporal patterns, they often smooth out or underrepresent anomalous events caused by extreme weather, infrastructure failures, and behavioral shifts. Preserving these events is challenging because they are sparse, localized in time and space, and shaped by heterogeneous dependencies across geographical proximity and regional attributes.
To address these challenges, we propose \m, a two-stage diffusion-based framework for anomaly-preserving energy consumption data generation. The first stage, Heterogeneous Graph-based Anomaly Semantic Learning (HG-ASL), extracts region-specific anomaly semantics from sparse residual structures by jointly modeling spatial and attribute dependencies across urban regions. The second stage, Anomaly Semantic-guided Diffusion (AS-Diff), injects the learned anomaly semantics into the denoising process to generate realistic consumption sequences while preserving anomalous patterns. This design enables controllable generation for individual regions and scales naturally to city-wide settings.
We evaluate \m on four real-world energy consumption datasets against 11 general-purpose and energy-specific generation baselines. Experimental results show that \m improves anomaly preservation fidelity by an average of 12.21\% and downstream quality by 2.96\%, while maintaining competitive overall generation fidelity compared to baselines.

\end{abstract}
\maketitle
\section{Introduction}

Fine-grained energy consumption data are essential for real-world applications such as demand forecasting, grid infrastructure planning, and grid reliability assessment~\cite{zanocco2022assessing}. However, access to such data remains limited due to privacy concerns and restrictive data-sharing agreements~\cite{quesada2024electricity}. These barriers have motivated growing interest in \textbf{synthetic energy consumption data generation}, with existing approaches spanning simulation-based~\cite{thorve2023high,yuan2023synthetic}, statistical~\cite{kang2023systematic}, GAN-based~\cite{hu2023multiload,razghandi2023smart}, and diffusion-based methods~\cite{fu2024creating,fuest2025cents}.

Despite substantial progress in reproducing overall consumption distributions and recurring temporal patterns, existing methods often struggle to preserve anomalous events faithfully. Energy consumption typically exhibits strong intra-day periodicity, with recurring daytime peaks and nighttime declines~\cite{eia2020hourly}. Because these dominant patterns make up the vast majority of observations, generative models tend to prioritize them when learning the underlying data distribution~\cite{sehwag2022generating}, while sparse and irregular variations receive considerably less attention. Consequently, anomalous events are often smoothed out, attenuated, or underrepresented in the generated data~\cite{galib2024fide,brophy2023generative}. Our empirical analysis in Figure~\ref{fig:zero_consumption_rate} also illustrates this limitation: the prevalence of zero-consumption anomalies drops from 9.14\% in the real data to only 4.84\% in synthetic data generated by a state-of-the-art model, indicating substantial underrepresentation of anomalies. However, preserving such events is critical, as they capture irregular demand changes caused by extreme weather, infrastructure failures, and behavioral changes~\cite{do2023spatiotemporal}.  

Faithfully modeling anomalous events in energy data remains challenging for two key reasons. \textbf{First}, anomalous events are not always random or independent, as many are shaped by heterogeneous dependencies across temporal dynamics, geographical proximity, and attribute similarity. These complex dependencies make the underlying anomaly distributions difficult to characterize, learn, and reproduce in synthetic data. \textbf{Second}, anomalies are typically localized in both time and space and account for only a small fraction of observations, making them easily overwhelmed or distorted by dominant consumption patterns during the generation process. 

% These challenges highlight the need for generation methods that explicitly model anomaly dependencies while preserving localized anomalous signals.

To address these challenges, we propose \textbf{\m}, a two-stage diffusion-based framework for generating synthetic energy consumption data while faithfully preserving anomalous events. Unlike existing methods that implicitly learn sparse anomalies alongside regular consumption patterns, \m centers on the latent representation of anomalous patterns, i.e., \textbf{anomaly semantics}. There are two key novel designs in \m:
(i) \textbf{Heterogeneous Graph-based Anomaly Semantic Learning (HG-ASL)} extracts anomaly semantics from sparse residual structures in real-world energy data. It first constructs region-specific semantic spaces to capture local temporal anomaly patterns, and then integrates them through heterogeneous graphs that explicitly model cross-region dependencies induced by geographical proximity and attribute similarity. This process yields a graph-enhanced global semantic space, from which region-conditioned anomaly semantics are sampled to guide subsequent generation.
(ii) \textbf{Anomaly Semantic-Guided Diffusion (AS-Diff)} consists of a pretrained diffusion Backbone Denoiser and a lightweight AS-Control network. The Backbone Denoiser preserves regular energy consumption patterns, while AS-Control injects sampled anomaly semantics into the denoising process through layer-wise control signals, enabling the generation of anomalous patterns consistent with the learned semantics while maintaining regular consumption patterns. This work integrates expertise in computer science and energy systems, as detailed in Appendix~\ref{appendix:collaboration}. The key contributions of this paper are as follows:
\begin{itemize}
\item \textbf{Conceptually}, we establish anomalous event preservation as a critical objective in synthetic energy consumption data generation. We show that realistic generation requires more than reproducing dominant regular consumption patterns, as anomalous events capture important dynamics of real-world energy systems.

\item \textbf{Technically}, we propose \m, a two-stage diffusion-based framework for
anomaly-preserving energy consumption data generation. The HG-ASL stage learns anomaly semantics from sparse residual structures by jointly modeling spatial and attribute dependencies across geospatial regions through heterogeneous graphs. The AS-Diff stage then injects the sampled anomaly semantics into the denoising process, enabling faithful preservation of anomalous events during synthetic consumption data generation.

\item \textbf{Experimentally}, we evaluate \m on four large-scale energy consumption datasets from Florida, New York, and California against 11 general-purpose and energy-specific generation baselines. Compared with the strongest baselines, \m improves anomaly preservation fidelity by 12.21\% and downstream performance by 2.96\% on average, while maintaining competitive overall generation fidelity.  Code
is available at \href{https://github.com/LinJiang18/SynEnergy}{\textcolor{blue}{\textbf{SynEnergy Code}}}.
\end{itemize}

\vspace{-6pt}
\section{Data-Driven Findings} \label{sec:data-driven}

In this section, we present a data-driven analysis that reveals several key observations motivating the design of \m. The analysis is based on real-world energy consumption data obtained through collaboration with a municipal utility provider in Tallahassee, Florida. The dataset spans more than ten years and includes over one billion household-level consumption records collected at 30-minute intervals from more than 50,000 smart meters. We additionally integrate U.S. Census data~\cite{uscensus_acs} to characterize the socioeconomic attributes of different geospatial regions. Further details about the dataset are provided in Appendix~\ref{appendix:dataset_florida}.

\textbf{Observation 1: Energy consumption anomalies exhibit structured patterns shaped by geographical proximity and attribute similarity.}
Although anomalous events are sparse in individual household consumption time series, their occurrences reveal systematic cross-region patterns within a city. These patterns are primarily associated with two forms of dependency: \emph{geographical proximity} and \emph{attribute similarity}. Geographically proximate regions may experience similar or synchronized anomalies because they share neighborhood-level contexts or are exposed to the same spatially propagating disruptions. Meanwhile, geographically separated regions with similar characteristics, such as socioeconomic conditions, may also exhibit similar anomaly patterns.

\begin{figure}[ht]
\centering
\vspace{-10pt}
\begin{minipage}{0.48\linewidth}
    \centering
    \includegraphics[width=\linewidth]{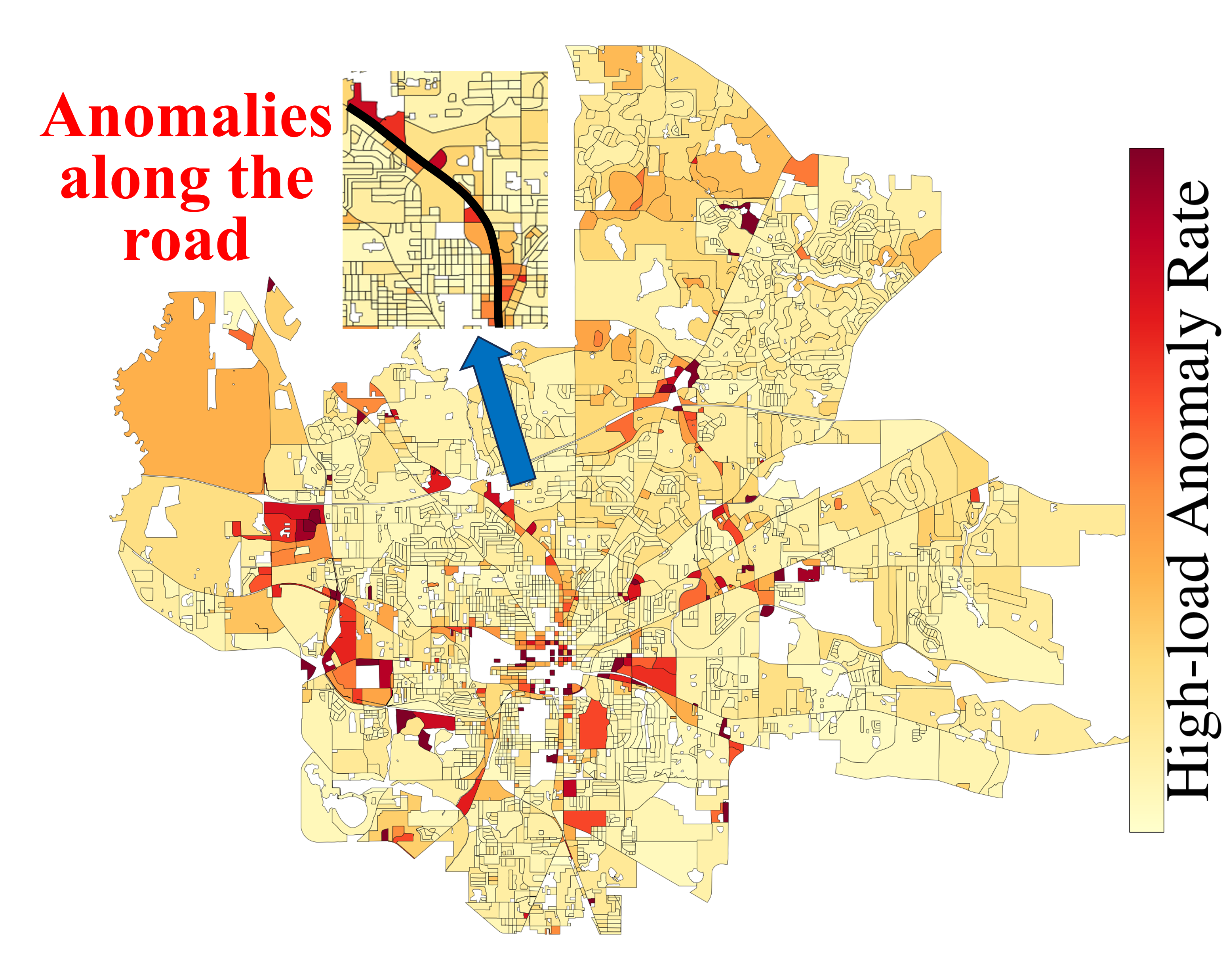}\vspace{-5pt}
    \caption*{(a) Spatial Distribution.}
\end{minipage}
\hfill
\begin{minipage}{0.46\linewidth}
    \centering
    \includegraphics[width=\linewidth]{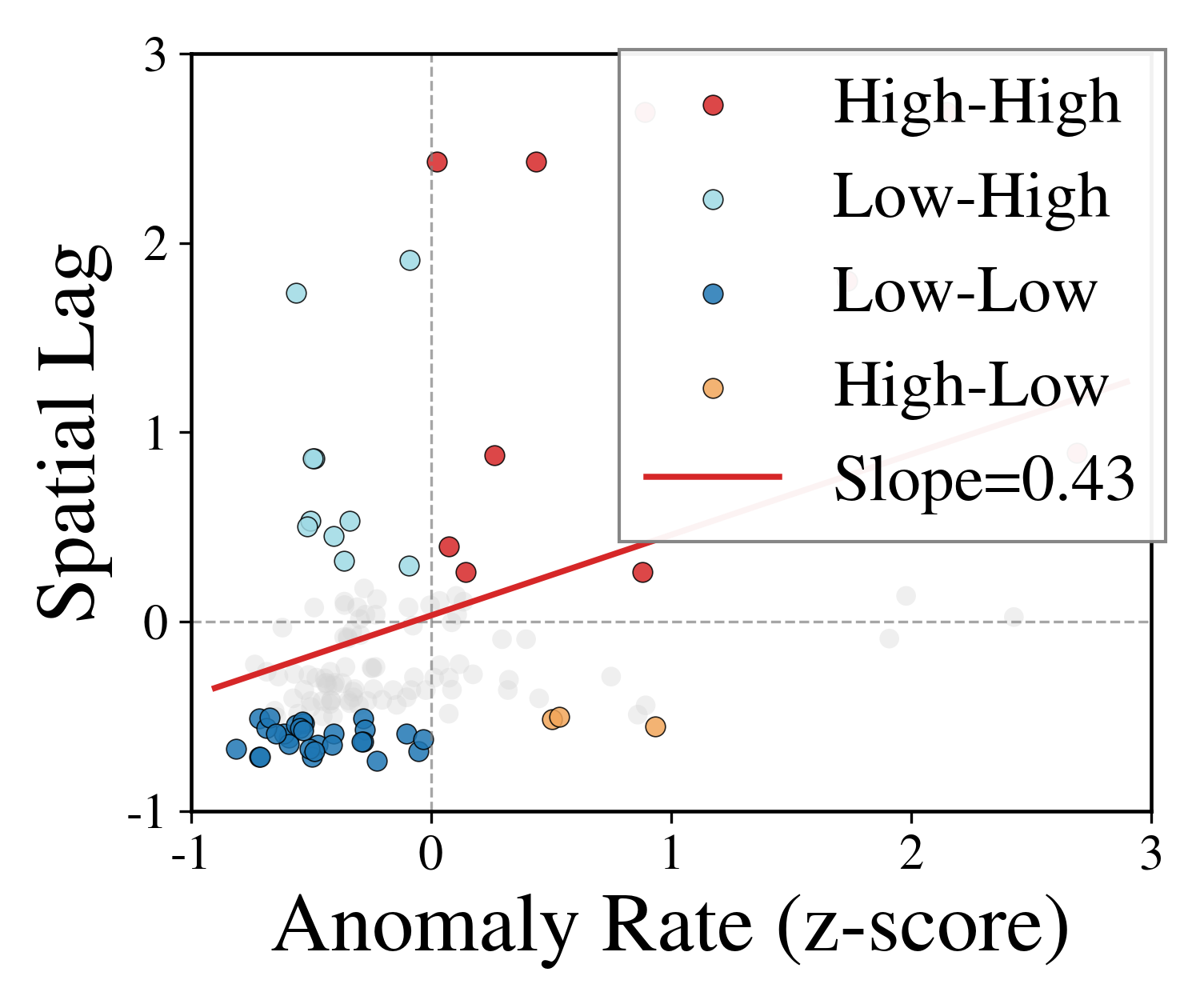}\vspace{-5pt}
    \caption*{(b) Moran’s I Scatter Plot.}
\end{minipage}
\vspace{-10pt}
\caption{Spatial Distribution and Autocorrelation of Anomalies in Tallahassee Energy Data.}
\label{fig:high_load_anomalies}
\end{figure}

We first investigate the influence of geographical proximity through \textbf{shared neighborhood-level contexts}. Figure~\ref{fig:high_load_anomalies}(a) presents the spatial distribution of anomaly rates across census blocks in Tallahassee, with darker shades of red indicating higher rates. Anomalies are visibly concentrated within particular neighborhoods and along both sides of certain roadways, rather than being randomly distributed across the city. The Moran's I scatter plot~\cite{anselin2019moran} in Figure~\ref{fig:high_load_anomalies}(b) further provides quantitative evidence of positive spatial autocorrelation. A positive slope of 0.43 indicates that census blocks with similar anomaly rates tend to cluster geographically. In particular, High--High regions have above-average anomaly rates and are surrounded by neighboring regions that also exhibit above-average rates. Together, these results demonstrate clear spatial dependencies in regional anomaly occurrences.

% \vspace{-5pt}

\vspace{-10pt}
\begin{figure}[b]
\centering
\vspace{-10pt}
\begin{minipage}{0.48\linewidth}
    \centering    \includegraphics[width=\linewidth]{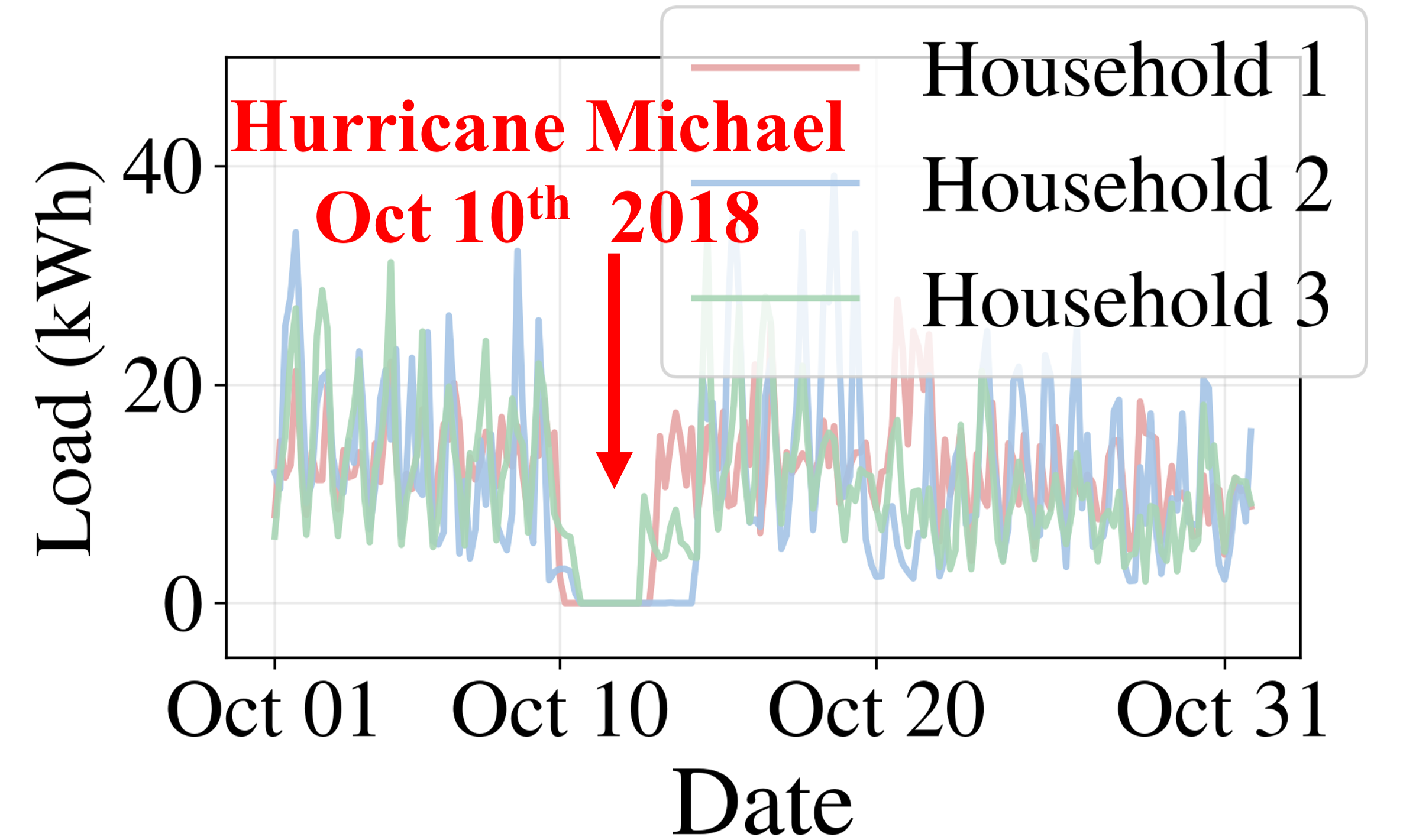}
    \captionof{figure}{Similar Household Consumption Anomalies in Three Neighboring CBGs.}
    \label{fig:low_visualization}
\end{minipage}
\hfill
\begin{minipage}{0.48\linewidth}
    \centering
    \includegraphics[width=\linewidth]{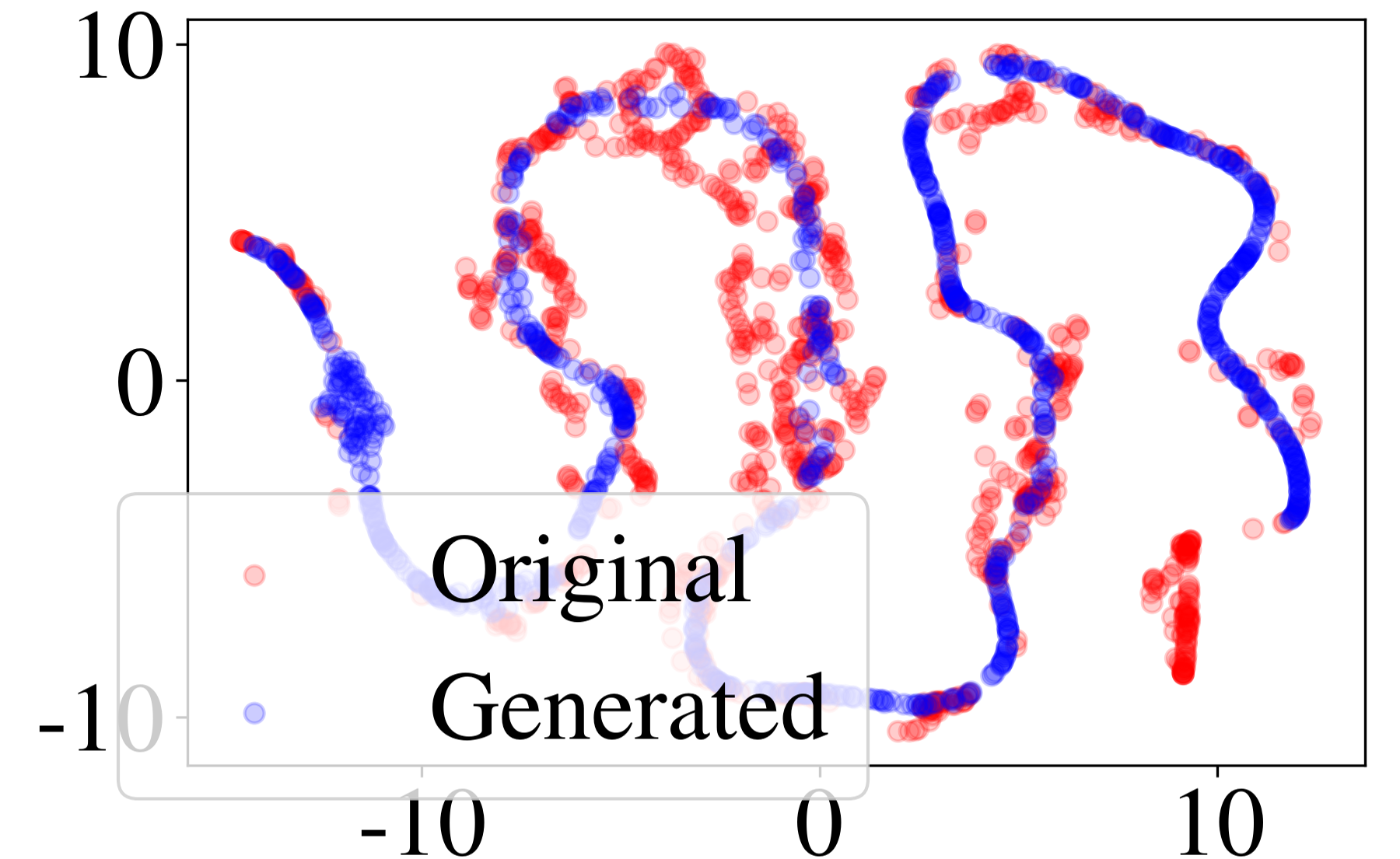}
    \captionof{figure}{Tail Detail Loss in t-SNE Distributions: Original vs. Generated Data. }
    \label{fig:tSNE_distribution}
\end{minipage}
\end{figure}

Next, we examine the influence of geographical proximity under \textbf{spatially propagating disruptions}. Figure~\ref{fig:low_visualization} presents household energy consumption time series from three neighboring census block groups (CBGs) during October 2018, when North Florida was severely affected by Category~5 Hurricane Michael. During this hurricane, households in these regions experienced nearly simultaneous drops in consumption followed by sustained periods of low usage, resulting in synchronized anomaly patterns. This case illustrates how a shared external disruption can induce coordinated anomalies across geographically neighboring regions.

Finally, we investigate how \textbf{attribute similarity} shapes anomaly patterns. We consider four illustrative regional attributes: median household income, property value, higher-education attainment rate, and work-from-home rate. As reported in Appendix~\ref{appendix:attribute}, three of these attributes exhibit significant correlations with regional anomaly rates. This finding suggests that regions with similar anomaly-relevant attributes may exhibit related anomaly patterns even when they are geographically distant.

\textbf{Observation 2: Existing time-series generation methods often overlook fine-grained variations and underrepresent anomalous events.}
To demonstrate this limitation, we use t-SNE~\cite{van2008visualizing} to visualize the original data and synthetic data generated by Diffusion-TS~\cite{yuan2024diffusionts}, a representative state-of-the-art time-series generation model. As shown in Figure~\ref{fig:tSNE_distribution}, the generated samples (blue) reproduce the main distributional structure of the original samples (red), but provide limited coverage of peripheral and outlying regions. This result suggests that less frequent and anomalous patterns are not adequately preserved during generation. More broadly, it indicates that matching the overall data distribution alone is insufficient to preserve anomalous events, motivating the design of \m. Appendix~\ref{appendix:limitation} provides a more comprehensive evaluation of anomaly preservation in existing generative methods, including low-consumption anomaly rates, anomaly-event durations, and anomaly magnitude and deviation distributions.

% \vspace{-3pt}
\section{Problem Formulation} \label{sec:formulation}

\subsection{Energy Consumption Data}

We consider a real-world energy consumption dataset $\mathbf{X}$ collected from $N$ regions. Let $r_n$ denote the $n$-th region, where $n\in\{1,\ldots,N\}$. Region $r_n$ contains $M_n$ households, and
$\mathbf{x}_{n,m}=[x_{n,m}^{1},\ldots,x_{n,m}^{K}]$
denotes the energy consumption sequence of the $m$-th household in that region, where $m\in\{1,\ldots,M_n\}$. Each element $x_{n,m}^{k}$ represents the accumulated energy consumption during the $k$-th time interval, and $K$ is the total number of intervals.
The temporal resolution may range from minutes to days.

\vspace{-4pt}
\subsection{Anomalous Event Definition} \label{sec:Anomalous_event_defintion}

For each energy consumption sequence $\mathbf{x}_{n,m}$, we consider two types of anomalous events: high-consumption and low-consumption events. A high-consumption anomalous event consists of one or more consecutive intervals during which consumption substantially exceeds its expected level, as may occur during extreme heat or unusually intensive appliance use. In contrast, a low-consumption anomalous event consists of one or more consecutive intervals during which consumption is unusually low or near zero relative to its expected pattern, as may occur during a power outage or temporary household absence.

To identify anomalous events, we first compute the residual sequence
$\mathbf{e}_{n,m}=[e_{n,m}^{1},\ldots,e_{n,m}^{K}]$, where
$e_{n,m}^{k}=x_{n,m}^{k}-b_{n}^{k}$ measures the deviation from the expected consumption pattern of region $r_n$ at the $k$-th time interval. The regional background sequence $\mathbf{b}_{n}=[b_{n}^{1},\ldots,b_{n}^{K}]$ is estimated from historical consumption records by averaging consumption across all households in region $r_n$ at each interval. The anomaly threshold $\delta$ is set to a predefined percentile of the absolute residual distribution. For example, the 90th percentile identifies the top 10\% of observations with the largest absolute residuals, and the percentile level is treated as a hyperparameter. A high-consumption anomalous event is then defined as one or more consecutive intervals satisfying $e_{n,m}^{k}>\delta$, whereas a low-consumption anomalous event consists of one or more consecutive intervals satisfying $e_{n,m}^{k}<-\delta$. Appendix~\ref{appendix:event_identification} illustrates the process of identifying anomalous events from raw consumption data.

% , while Section~\ref{xxx} examines the sensitivity to different threshold settings.

\vspace{-2pt}
\subsection{Diffusion-based Energy Data Generation} \label{sec:generation_formulation}
We formulate energy consumption generation using diffusion models, which have emerged as a leading approach to general-purpose time-series generation~\cite{wang2025non}. We utilize the following information as model input: (i) a real-world energy consumption dataset $\mathbf{X}$; (ii) spatial and attribute adjacency matrices $\mathbf{A}^{\mathrm{sp}}$ and $\mathbf{A}^{\mathrm{attr}}$, which serve as cross-region priors encoding geographical proximity and attribute similarity; and (iii) a region condition $\mathbf{c}_n$, which encodes the identity and attributes of region $r_n$. In our implementation, $\mathbf{c}_n$ consists of the region index and two region-level attributes: median household income and median housing value. 
We formulate anomaly-preserving energy consumption generation as a conditional diffusion process jointly guided by the region condition $\mathbf{c}_n$ and the region-conditioned anomaly semantic representation $\bar{\mathbf{u}}_{n}^{a}$. The region condition identifies the target region and characterizes its general consumption context, whereas $\bar{\mathbf{u}}_{n}^{a}$ is learned to capture the region's anomalous patterns. Conditioned on these two signals, the generation process begins with Gaussian noise $\mathbf{s}_T$ and progressively reconstructs a consumption sequence by sampling from the conditional reverse distribution
$p_{\theta}(\mathbf{s}_{t-1}\mid\mathbf{s}_t,\mathbf{c}_n,\bar{\mathbf{u}}_{n}^{a})$ for $t=\{T,\ldots,1\}$.  
The final sample $\mathbf{s}_0$ is taken as the synthetic household consumption sequence $\bar{\mathbf{x}}_{n,m}$ for region $r_n$. Through this dual conditioning mechanism, \m supports controllable generation for a specified region while preserving its characteristic anomalous patterns. The framework can also be readily scaled to city-wide generation across hundreds of regions.
\section{Methodology}\label{sec:method}

\begin{figure*}[t]
    \centering
    \includegraphics[width=\textwidth]{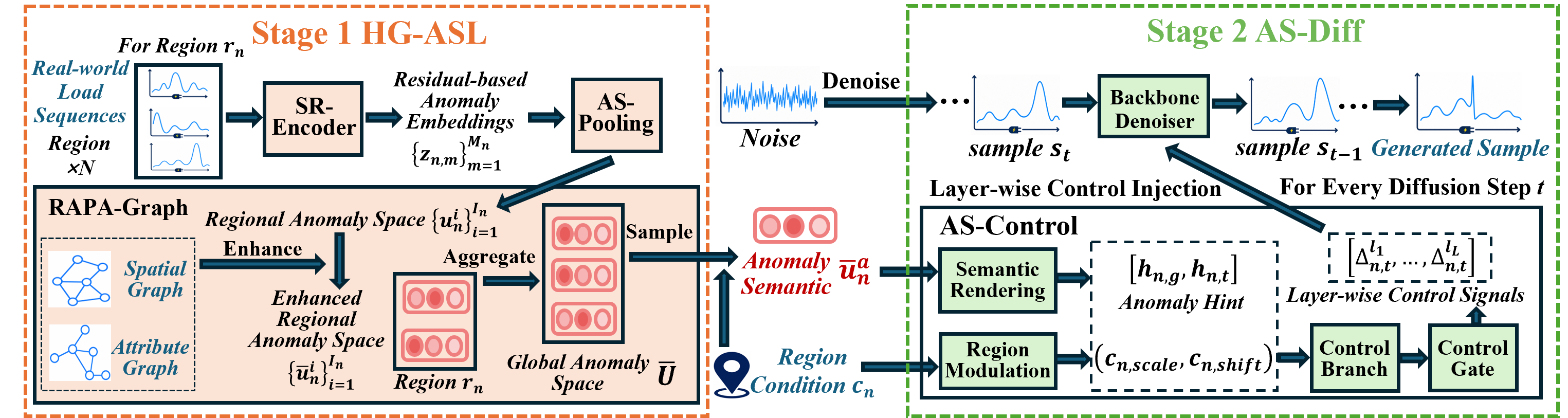}\vspace{-5pt}
    \caption{The Overall Framework of \m.}
    \label{fig:framework}
    \vspace{-10pt}
\end{figure*}

In this paper, we design \m, an anomaly semantic-guided diffusion framework for synthetic
energy data generation. An overall framework of \m is illustrated in Figure~\ref{fig:framework}, which consists of two key components: a Heterogeneous Graph-based
Anomaly Semantic Learning (HG-ASL) module to represent sparse anomalous patterns by extracting anomaly semantics from real-world energy consumption data, and an Anomaly Semantic-Guided Diffu-
sion (AS-Diff) module to leverage these learned semantics to guide anomaly-preserving diffusion generation.

\subsection{Heterogeneous Graph-based Anomaly Semantic Learning}\label{sec:HG-ASL}
We design a HG-ASL module to learn anomaly semantics from real-world energy consumption sequences, which consists of three key components: \textbf{SR-Encoder} first learns \emph{residual-based anomaly embeddings}, \textbf{AS-Pooling} then organizes these embeddings into \emph{regional anomaly spaces}, and \textbf{RAPA-Graph} further models cross-region dependencies to produce \emph{enhanced regional anomaly spaces}. These enhanced spaces are aggregated into a \emph{global anomaly space}, from which anomaly semantics are sampled to guide the subsequent generation process.

\subsubsection{\textbf{SR-Encoder}}
\label{sec:SR-Encoder}

The Sparse Residual Encoder (SR-Encoder) learns residual-based anomaly embeddings that characterize sparse anomalous deviations in consumption sequences. As defined in Section~\ref{sec:Anomalous_event_defintion}, positive and negative residuals represent consumption above and below the expected regional pattern, respectively, making residuals a natural representation of both high- and low-consumption anomalies~\cite{schmidl2022anomaly}.

Given the residual sequence $\mathbf{e}_{n,m}$ defined in Section~\ref{sec:Anomalous_event_defintion}, we first apply robust standardization to account for scale differences across household sequences:
\begin{equation} \label{eq:1}
\bar{\mathbf{e}}_{n,m}
=
\frac{\mathbf{e}_{n,m}}
{\operatorname{MAD}(\mathbf{e}_{n,m})+\epsilon_e},
\end{equation}
where $\operatorname{MAD}(\cdot)$ denotes the median absolute deviation and $\epsilon_e$ ensures numerical stability. Since most residuals are near zero, we use soft weighting to emphasize informative deviations:
\begin{equation} \label{eq:2}
\mathbf{w}^{e}_{n,m}
=
\operatorname{sigmoid}
\left(
\gamma_e
\left(
|\bar{\mathbf{e}}_{n,m}|-\tau_e
\right)
\right),
\qquad
\mathbf{e}^{f}_{n,m}
=
\mathbf{w}^{e}_{n,m}\odot\bar{\mathbf{e}}_{n,m},
\end{equation}
where $\tau_e$ controls the activation position and $\gamma_e$ controls the weighting sharpness. This soft weighting suppresses near-zero residuals while emphasizing anomaly-related information. A visualization for $\mathbf{e}^{f}_{n,m}$ is provided in Figure~\ref{fig:anomaly_identify} in Appendix~\ref{appendix:event_identification}.

The zero-aware encoder $E_{\theta}$ takes the filtered residual $\mathbf{e}^{f}_{n,m}$ and its corresponding weight $\mathbf{w}^{e}_{n,m}$ as inputs:
\begin{equation}  \label{eq:3}
\mathbf{z}_{n,m}
=
E_{\theta}
\left(
\mathbf{e}^{f}_{n,m},
\mathbf{w}^{e}_{n,m}
\right),
\qquad
\mathcal{Z}_{n}
=
\left\{
\mathbf{z}_{n,m}
\right\}_{m=1}^{M_n}.
\end{equation}
Here, $\mathbf{z}_{n,m}$ denotes the residual-based anomaly embedding of household $m$ in region $r_n$, and $\mathcal{Z}_{n}$ collects the embeddings of all households in the region. The encoder employs temporal convolutional layers and complementary pooling operations to encode anomalous information. Its detailed design is provided in Appendix~\ref{appendix:zero_aware_encoder}.

\subsubsection{\textbf{AS-Pooling}}
\label{sec:AS-Pooling}
Given the residual-based anomaly embeddings $\mathcal{Z}_n$ for region $r_n$, Anomaly Semantic Pooling (AS-Pooling) groups similar embeddings to construct a regional anomaly space $\mathcal{U}_n=\{\mathbf{u}_{n,i}\}_{i=1}^{I_n}$. Each $\mathbf{u}_{n,i}$ represents a characteristic anomaly pattern and is defined as an anomaly semantic. Specifically, we compute pairwise cosine similarities and group embeddings with similar directions in the representation space:
\begin{equation}
\{\mathcal{Z}_{n,i}\}_{i=1}^{I_n}
=
\operatorname{Cluster}
\left(
\mathcal{Z}_n,
\mathbf{S}_n
\right),\quad
\mathbf{u}_{n,i}
=
|\mathcal{Z}_{n,i}|^{-1}
\sum\nolimits_{\mathbf{z}\in\mathcal{Z}_{n,i}}
\mathbf{z},
\end{equation}
where $\mathbf{S}_n(m,m')$ denotes the cosine similarity between $\mathbf{z}_{n,m}$ and $\mathbf{z}_{n,m'}$, and $|\mathcal{Z}_{n,i}|$ denotes the number of anomaly embeddings assigned to the $i$-th cluster in region $r_n$.

\subsubsection{\textbf{RAPA-Graph}} \label{sec:RAPA-Graph}
Although AS-Pooling constructs the regional anomaly space $\mathcal{U}_n$, it learns anomaly semantics independently within each region, thereby failing to capture cross-region dependencies. Therefore, we further design Relation-aware Anomaly Pattern Attention Graph (RAPA-Graph), which enhances $\mathcal{U}_n$ by modeling semantic dependencies across regions to capture city-scale anomaly patterns. Taking the spatial graph $\mathbf{A}^{\mathrm{sp}}$ and attribute graph $\mathbf{A}^{\mathrm{attr}}$ introduced in Section~\ref{sec:generation_formulation} as inputs, RAPA-Graph incorporates the regional proximity and attribute similarity encoded by these graphs as structural priors and learns three parameter groups: $\mathbf{W}_{\mathrm{att}}$ for semantic relevance, $\mathbf{W}_{\mathrm{rel}}$ for relation balancing, and $\mathbf{W}_{\mathrm{gate}}$ for gated updating.

\textit{Firstly}, RAPA-Graph captures semantic relevance among anomaly semantics from different regions. For each target semantic $\mathbf{u}_{n,i}\in\mathcal{U}_n$, it evaluates its relevance to semantics from neighboring regions under each relation $\rho\in\{\mathrm{sp},\mathrm{attr}\}$, while incorporating the corresponding prior graph $\mathbf{A}^{\rho}$ as a structural bias. Specifically, for each neighboring region $r_q$, where $q\in\mathcal{N}^{\rho}_{n}$, the relevance between $\mathbf{u}_{n,i}$ and each semantic $\mathbf{u}_{q,j}\in\mathcal{U}_q$ is computed as:
\begin{equation}
e^{\rho}_{n,i,q,j}
=
\left(\mathbf{W}^{\rho}_{\mathrm{att}}\mathbf{u}_{n,i}\right)^{\top}
\left(\mathbf{W}^{\rho}_{\mathrm{att}}\mathbf{u}_{q,j}\right)
+
\log\left(\mathbf{A}^{\rho}_{n,q}+\epsilon\right).
\end{equation}
The resulting scores are normalized to aggregate a relation-specific neighboring representation, which introduces cross-region information to enhance the target semantic $\mathbf{u}_{n,i}$ in region $r_n$:
\begin{equation}
\mathbf{u}^{\rho,\mathrm{nbr}}_{n,i}
=
\sum_{q\in\mathcal{N}^{\rho}_{n}}
\sum_{j=1}^{I_q}
\frac{\exp\left(e^{\rho}_{n,i,q,j}\right)}
{\sum_{q'\in\mathcal{N}^{\rho}_{n}}\sum_{j'=1}^{I_{q'}}\exp\left(e^{\rho}_{n,i,q',j'}\right)}
\mathbf{u}_{q,j}.
\end{equation}

\textit{Second}, to adaptively balance the two heterogeneous relations, RAPA-Graph learns $\mathbf{W}_{\mathrm{rel}}$ to assign relation-specific weights to the spatial- and attribute-neighbor representations, denoted by $\{\mathbf{u}^{\mathrm{sp},\mathrm{nbr}}_{n,i}, \mathbf{u}^{\mathrm{attr},\mathrm{nbr}}_{n,i}\}$. These weights determine their respective contributions to the fused neighboring representation:
\begin{equation}
[\alpha^{\mathrm{sp}}_{n,i}, \alpha^{\mathrm{attr}}_{n,i}]
=
\operatorname{softmax}
\left(
\mathbf{W}_{\mathrm{rel}}\mathbf{u}_{n,i}
\right),
\mathbf{u}^{\mathrm{nbr}}_{n,i}
=
\alpha^{\mathrm{sp}}_{n,i}\mathbf{u}^{\mathrm{sp},\mathrm{nbr}}_{n,i}
+
\alpha^{\mathrm{attr}}_{n,i}\mathbf{u}^{\mathrm{attr},\mathrm{nbr}}_{n,i}.
\end{equation}
\textit{Finally}, $\mathbf{W}_{\mathrm{gate}}$ controls how much neighboring information is injected into the original semantic through a gated residual update:
\begin{equation}
    \bar{\mathbf{u}}_{n,i}
    =
    \mathbf{u}_{n,i}
    +
    \sigma
    \left(
    \mathbf{W}_{\mathrm{gate}}
    [\mathbf{u}_{n,i};\mathbf{u}^{\mathrm{nbr}}_{n,i}]
    \right)
    \odot
    \mathbf{u}^{\mathrm{nbr}}_{n,i}.
\end{equation}
Here, $\odot$ denotes element-wise multiplication, and $\sigma(\cdot)$ denotes the sigmoid gate function. The resulting $\bar{\mathbf{u}}_{n,i}$ is a graph-enhanced anomaly semantic, and all such semantics form the enhanced regional anomaly space $\bar{\mathcal{U}}_n
=\{\bar{\mathbf{u}}_{n,i}\}_{i=1}^{I_n}$ for region $r_n$.

By integrating the enhanced regional anomaly spaces $\{\bar{\mathcal{U}}_n\}_{n=1}^{N}$ across all regions, RAPA-Graph constructs a global anomaly space $\bar{\mathcal{U}}$ that captures city-scale anomaly patterns in real-world energy consumption data. During generation, an anomaly semantic $\bar{\mathbf{u}}_{n}^{a}$ associated with the target region condition $\mathbf{c}_n$ is sampled from $\bar{\mathcal{U}}$ to guide generators toward the corresponding anomaly pattern:
\begin{equation}
\bar{\mathcal{U}}
=
\bigcup\nolimits_{n=1}^{N}\bar{\mathcal{U}}_n,
\qquad
\bar{\mathbf{u}}_{n}^{a}
\sim
P_{\mathrm{sem}}(\bar{\mathcal{U}}\mid \mathbf{c}_n).
\end{equation}
where $\mathbf{c}_n$ denotes the region condition introduced in Section~\ref{sec:generation_formulation}, $P_{\mathrm{sem}}(\bar{\mathcal{U}}\mid \mathbf{c}_n)$ is the region-conditioned sampling distribution, and $\bar{\mathbf{u}}_{n}^{a}$ is the \textbf{sampled anomaly semantic} provided as an additional condition for anomaly-preserving generation in stage 2.

To train RAPA-Graph, we optimize the graph-enhanced semantics to cover real anomaly representations at both the city-wide and individual-region levels while preserving the original regional semantics. The training objectives are defined as follows:
\begin{equation}
\begin{aligned}
&\mathcal{L}_{\mathrm{RAPA}}
=
\mathcal{L}_{\mathrm{glo}}
+
\lambda_{\mathrm{reg}}
\mathcal{L}_{\mathrm{reg}}
+
\lambda_{\mathrm{pre}}
\mathcal{L}_{\mathrm{pre}},
\quad
\mathcal{L}_{\mathrm{glo}}
=
\frac{1}{|\mathcal{Z}|}
\sum_{\mathbf{z}\in\mathcal{Z}}
\min_{\bar{\mathbf{u}}\in\bar{\mathcal{U}}}
\left\|
\mathbf{z}
-
\bar{\mathbf{u}}
\right\|_2^2, \\
&\mathcal{L}_{\mathrm{reg}}
=
\frac{1}{N}
\sum_{n=1}^{N}
\frac{1}{|\mathcal{Z}_n|}
\sum_{\mathbf{z}\in\mathcal{Z}_n}
\min_{\bar{\mathbf{u}}\in\bar{\mathcal{U}}}
\left\|
\mathbf{z}
-
\bar{\mathbf{u}}
\right\|_2^2,
\quad
\mathcal{L}_{\mathrm{pre}}
=
\sum_{n=1}^{N}
\sum_{i=1}^{I_n}
\left\|
\bar{\mathbf{u}}_{n,i}
-
\mathbf{u}_{n,i}
\right\|_2^2.
\end{aligned}
\end{equation}
Here, $\mathcal{Z}=\bigcup_{n=1}^{N}\mathcal{Z}_n$ denotes all residual-based anomaly embeddings produced by SR-Encoder. The cardinalities $|\mathcal{Z}|$ and $|\mathcal{Z}_n|$ denote the numbers of anomaly embeddings across all regions and within region $r_n$, respectively. The global loss $\mathcal{L}_{\mathrm{glo}}$ encourages $\bar{\mathcal{U}}$ to cover the overall anomaly distribution across all regions. In contrast, the region-wise loss $\mathcal{L}_{\mathrm{reg}}$ gives each region balanced consideration, preventing regions with more anomaly embeddings from dominating the training objective. The preservation loss $\mathcal{L}_{\mathrm{pre}}$ limits excessive graph-induced drift from the original regional semantics.

% For each relation, RAPA-Graph first selects the top-$K_r$ neighbors of region $r_n$ from the corresponding relation graph:
% \begin{equation}
%     \mathcal{N}^{\rho}_{n}
%     =
%     \operatorname{TopK}_{K_r}
%     \left(
%     \mathbf{A}^{\rho}_{n,:}
%     \right),
%     \quad
%     \rho\in\{\mathrm{sp},\mathrm{attr}\}.
% \end{equation}

\subsection{Anomaly Semantic-Guided Diffusion}\label{sec:AS-Diff}
Given the anomaly semantics $\bar{\mathbf{u}}_{n}^{a}$ sampled by HG-ASL, we then design AS-Diff to generate realistic energy consumption sequences through diffusion while preserving fine-grained anomalous events. AS-Diff comprises two key components: a \textbf{Backbone Denoiser} that models the overall consumption distribution, and \textbf{AS-Control} that converts the anomaly semantics into control signals and injects them layer by layer into the Backbone Denoiser to enable anomaly-guided generation.

\subsubsection{Backbone Denoiser} \label{sec:backbone denoiser}
As shown in the AS-Diff module in Figure~\ref{fig:framework}, the Backbone Denoiser parameterizes each reverse transition from $\mathbf{s}_t$ to $\mathbf{s}_{t-1}$, progressively transforming the initial Gaussian noise $\mathbf{s}_T$ into the final synthetic sample $\mathbf{s}_0$. Architecturally, it is implemented as a Transformer-based denoising network with an encoder-decoder structure, as illustrated in Figure~\ref{fig:AS-Diff}. The Encoder captures temporal dependencies in the intermediate consumption sequence $\mathbf{s}_t$ through self-attention, while the multi-layer Decoder integrates the Encoder output and decomposes the hidden representations into trend and seasonal components. Further details of the Backbone Denoiser architecture are provided in Appendix~\ref{appendix:backbone_denoiser}.

The output of the multi-layer Decoder is then used in the reverse update function to get the next intermediate sample $s_{t-1}$:
\begin{equation}\label{eq:x_update}
\mathbf{s}_{t-1}
=
\frac{1}{1-\bar{\alpha}_t}
\Big(
  \sqrt{\bar{\alpha}_{t-1}}\,\beta_t\,\hat{\mathbf{s}}_{0}(\mathbf{s}_{t},t)
  +
  \sqrt{\alpha_t}\,(1-\bar{\alpha}_{t-1})\,\mathbf{s}_t
\Big)
+
\tilde{\sigma}_t\,\boldsymbol{\epsilon}_t,
\end{equation}
where $\hat{\mathbf{s}}_{0}(\mathbf{s}_{t},t)$ denotes the $x$-prediction estimate~\cite{yuan2024diffusionts,li2026back} of the clean load sequence at diffusion step $t$. Here, $\beta_t$ denotes the noise schedule at step $t$, $\alpha_t=1-\beta_t$, and $\bar{\alpha}_t=\prod_{\tau=1}^{t}\alpha_\tau$ is the cumulative noise-retention coefficient. The term $\boldsymbol{\epsilon}_t\sim\mathcal{N}(\mathbf{0},\mathbf{I})$ denotes the Gaussian noise injected during sampling, and $\tilde{\sigma}_t$ controls its scale.

\subsubsection{AS-Control} \label{sec:AS-Control}

\begin{figure}[t]
    \centering
    \includegraphics[width=0.9\columnwidth]{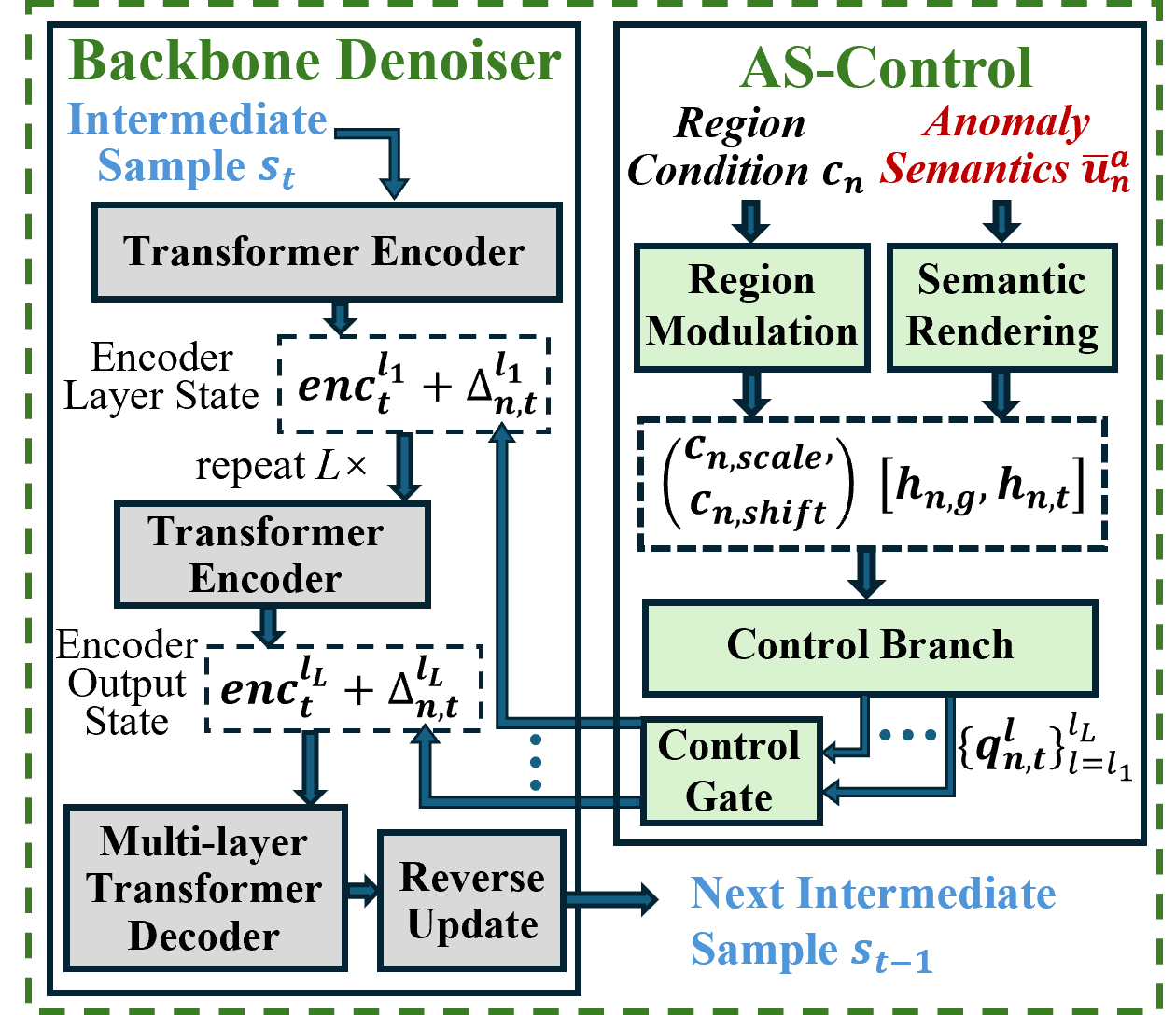}
    \caption{The Layer-wise Control Injection Between Backbone Denoiser and AS-Control in AS-Diff}
    \label{fig:AS-Diff}
    \vspace{-15pt}
\end{figure}

AS-Control extends the Backbone Denoiser with anomaly-preserving guidance. At each diffusion step $t$, it converts the sampled anomaly semantics $\bar{\mathbf{u}}_{n}^{a}$ into control signals and injects them into the denoising process to steer generation toward the specified anomaly pattern. As shown in Figure~\ref{fig:AS-Diff}, AS-Diff is jointly conditioned on the sampled anomaly semantics $\bar{\mathbf{u}}_{n}^{a}$ from HG-ASL and the region condition $\mathbf{c}_n$, which specify the target anomaly pattern and region, respectively. AS-Control incorporates these conditions through four components: Semantic Rendering, Region Modulation, Control Branch, and Control Gate.

\textbf{Semantic Rendering}: 
Given the sampled anomaly semantics $\bar{\mathbf{u}}_{n}^{a}$, Semantic Rendering converts it into time-aligned anomaly hints through two complementary parts. The global anomaly hint $\mathbf{h}_{n,g}\in\mathbb{R}^{T\times d}$ is obtained by normalizing $\bar{\mathbf{u}}_{n}^{a}$ with LayerNorm~\cite{ba2016layer} and passing it through a two-layer MLP, which provides the global anomaly context across the temporal dimension. The temporal anomaly hint $\mathbf{h}_{n,t}\in\mathbb{R}^{T\times d}$ is obtained by decoding $\bar{\mathbf{u}}_{n}^{a}$ with the residual decoder paired with the SR-Encoder in Section~\ref{sec:SR-Encoder}. This decoder produces the anomaly occurrence probability $\hat{\mathbf{p}}_{n}$ and deviation pattern $\hat{\mathbf{v}}_{n}$, which capture when and how anomalies appear over time. The final anomaly hint is formed by concatenating the global and temporal anomaly hints:
\begin{equation}
    \mathbf{h}_{n}^{a}
    =
    [\mathbf{h}_{n,g},\mathbf{h}_{n,t}],
    \qquad
    \mathbf{h}_{n,t}
    =
    \operatorname{MLP}_{t}
    \left(
    \hat{\mathbf{p}}_{n}
    \odot
    \hat{\mathbf{v}}_{n}
    \right).
\end{equation}
where $\mathbf{h}_{n,g}$ provides global anomaly semantics and $\mathbf{h}_{n,t}$ provides temporal anomaly details for the subsequent control branch.

\textbf{Region Modulation}: 
Region Modulation incorporates target region information into the Control Branch. Given the region label $c_n$, we use adaptive layer normalization (AdaLN)~\cite{peebles2023scalable} to generate layer-wise scale and shift parameters:
\begin{equation}
    \mathbf{c}_{n,\mathrm{scale}}^{l}, \mathbf{c}_{n,\mathrm{shift}}^{l}
    =
    \operatorname{AdaLN}^{l}(c_n).
\end{equation}
where the scale parameter adjusts the magnitude of the anomaly hint representation in the $l$-th layer, while the shift parameter adjusts its region-conditioned bias. This layer-wise modulation allows the control signal to reflect regional anomaly patterns.

\textbf{Control Branch}: 
The Control Branch transforms the anomaly hint $\mathbf{h}_{n}^{a}$ and the regional modulation parameters $\mathbf{c}_{n,\mathrm{scale}}^{l}$ and $\mathbf{c}_{n,\mathrm{shift}}^{l}$ into layer-wise control signals. Specifically, it mirrors the Encoder structure of the Backbone Denoiser to maintain layer-wise correspondence with the selected injection layers. The $\mathbf{h}_{n}^{a}$ is first used to initialize the hint representation $\mathbf{h}_{n,t}^{l_1}$ of the Control Branch:
\begin{equation}
    \mathbf{h}_{n,t}^{l_1}
    =
    \operatorname{Fuse}(\mathbf{h}_{n}^{a}).
\end{equation}
At the $l$-th control block, the hint representation from the preceding layer $\mathbf{h}_{n,t}^{l-1}$ is updated with the regional modulation parameters to obtain $\mathbf{h}_{n,t}^{l}$, which is then projected into the control feature $\mathbf{q}_{n,t}^{l}$:
\begin{equation}
    \mathbf{h}_{n,t}^{l}
    =
    \operatorname{CEnc}^{l}
    \left(
    \mathbf{h}_{n,t}^{l-1}, 
    \mathbf{c}_{n,\mathrm{scale}}^{l}, 
    \mathbf{c}_{n,\mathrm{shift}}^{l}
    \right),
    \quad
    \mathbf{q}_{n,t}^{l}
    =
    \operatorname{ZeroP}^{l}(\mathbf{h}_{n,t}^{l}).
\end{equation}

\textbf{Control Gate}: 
As shown in Figure~\ref{fig:AS-Diff}, the layer-wise control features $\{\mathbf{q}_{n,t}^{l}\}_{l=l_1}^{l_L}$ are passed through the Control Gate to produce the final injected signals using two gating mechanisms. The first is a temporal gate derived from the anomaly occurrence probability $\hat{\mathbf{p}}_{n}$ in Semantic Rendering, which determines where control is applied along the consumption sequence. The second is a preset diffusion-step gate $\lambda_{\mathrm{ctrl},t}$, which controls the overall injection strength at step $t$. The final gated control signal is obtained through:
\begin{equation}
    \Delta_{n,t}^{l}
    =
    \lambda_{\mathrm{ctrl},t}
    \cdot
    \mathcal{G}_{n}
    \left(
    \mathbf{q}_{n,t}^{l}, \hat{\mathbf{p}}_{n}
    \right),
    \quad
    l \in \{l_1,\ldots,l_L\}.
\end{equation}
where $\mathcal{G}_{n}(\cdot)$ applies temporal gating to the control feature according to the anomaly occurrence probability $\hat{\mathbf{p}}_{n}$. Finally, as shown in Figure~\ref{fig:AS-Diff}, the layer-wise control signals $\{\Delta_{n,t}^{l}\}_{l=l_1}^{l_L}$ are injected into the corresponding encoder layer states $\mathbf{enc}_{t}^{l}$ of the Backbone Denoiser, enabling layer-wise anomaly guidance during generation.
\section{Evaluation} \label{sec:experiment}
In this section, we conduct a comprehensive experimental evaluation of the proposed \m. Specifically, we address the following nine research questions (RQs):
\begin{itemize}[leftmargin=*]
\item \textbf{RQ 1}: How well does \m preserve overall fidelity?
\item \textbf{RQ 2}: How well does \m preserve anomaly fidelity?
\item \textbf{RQ 3}: How well does \m preserve downstream utility?
\item \textbf{RQ 4}: How can we visualize \m's anomaly-preserving performance from multiple perspectives?
\item \textbf{RQ 5}: How effectively does \m capture specific large-scale anomalous events, such as power outages and heatwaves?
\item \textbf{RQ 6}: How does the generation quality of \m vary across different generation scales and temporal granularities?
\item \textbf{RQ 7}: How sensitive is \m to the anomaly threshold $\delta$?
% \item \textbf{RQ 8}: How does \m perform in the ablation study?
\item \textbf{RQ 8}: How does each component of \m contribute to its overall performance?
\item \textbf{RQ 9}: Is \m computationally efficient?
\end{itemize}

\subsection{Evaluation Setup}\label{sec:setup}
\subsubsection{Datasets} 
We evaluate \m using real-world energy consumption data from three U.S. states. The Florida dataset contains ten years of household-level records from over 50K independently metered households across 147 census block groups (CBGs) in Tallahassee, with measurements recorded at 30-minute intervals, totaling over one billion data points. From this dataset, we construct two subsets centered on representative large-scale anomalous events: \textbf{FL1}, covering Hurricane Michael in October 2018, and \textbf{FL2}, covering a major heatwave in May 2019. To evaluate the generalizability of \m, we further incorporate two public datasets, \textbf{NY} and \textbf{CA}, collected in New York and California in 2024, respectively~\cite{nyiso,caiso}. Both provide hourly energy consumption records at a coarser spatial granularity, covering 11 regions and 4 zones, respectively. The preprocessing procedures and resulting experimental dataset statistics are provided in Appendix~\ref{appendix:dataset}.

\subsubsection{Baselines}
We compare \m with 11 state-of-the-art baselines across seven categories: (1) \textbf{GAN-based}: TimeGAN~\cite{yoon2019time}; (2) \textbf{VAE-based}: TimeVAE~\cite{desai2021timevae} and koVAE~\cite{naimangenerative}; (3) \textbf{Flow-based}: F-Flow~\cite{alaa2021generative}; (4) \textbf{Diffusion-based}: DiffWave~\cite{kongdiffwave} and Diffusion-TS~\cite{yuan2024diffusionts}; (5) \textbf{LLM-based}: SDForger~\cite{rousseau2025forging}; (6) \textbf{Anomaly-aware Generation}: FIDE~\cite{galib2024fide} and HeavyDiff~\cite{pandey2025heavy}; and (7) \textbf{Energy-specific Generation}: CENTS~\cite{fuest2025cents} and Cond-Diff~\cite{fu2024creating}. Additional details of these baselines are provided in Appendix~\ref{Appendix:Baseline}.

\subsubsection{Metrics}
We evaluate generation performance from three complementary dimensions using 10 metrics. First, \textbf{Overall Generation Fidelity} measures the overall similarity between real and generated consumption data using T-Wass., D-Wass., S-Wass., and MMD, following common practice in time-series generation evaluation~\cite{ang2023tsgbench}. Second, \textbf{Anomaly Preservation Fidelity} evaluates how well anomalous events are preserved using A-Rate, A-Count, A-Energy, and A-Tail. Lower values indicate better performance for all metrics in these two dimensions. Third, \textbf{Downstream Quality} measures the utility of generated data for anomaly detection and prediction tasks using Det-PRAUC and Pred-PRAUC, respectively, under the \textit{train-on-synthetic, test-on-real} setting~\cite{esteban2017real}, with higher values indicating better performance. Detailed definitions and implementation details are provided in Appendix~\ref{Appendix:Metric}.

\subsubsection{Parameter Settings}
Across the FL1 and FL2 datasets, we aggregate consecutive readings into \textbf{4-hour} intervals and construct \textbf{one-month} sequences, each comprising six observations per day and 180 time steps. Each dataset contains \textbf{10,000} household samples. Following Section~\ref{sec:Anomalous_event_defintion}, we set the anomaly threshold $\delta$ to the \textbf{90th} percentile of the absolute residual distribution. The corresponding settings for the NY and CA datasets are provided in Appendix~\ref{appendix:dataset}. Complete configurations are available in our code.
\begin{table*}[!h]
\centering
\footnotesize
\renewcommand{\arraystretch}{1.0}
\setlength{\tabcolsep}{2pt}

\caption{Overall generation fidelity, anomaly preservation fidelity, and downstream quality on the FL1 dataset. Best results are highlighted in bold, and second-best results are underlined. $\uparrow$ and $\downarrow$ indicate that higher and lower values are better, respectively.}\vspace{-3pt}
\label{tab:FL1}
\begin{tabular}{cclcccccccccc}
\toprule
\multirow{2}{*}{\textbf{Dataset}} 
& \multirow{2}{*}{\textbf{Type}} 
& \multirow{2}{*}{\textbf{Method}} 
& \multicolumn{4}{c}{\textbf{Overall Generation Fidelity}} 
& \multicolumn{4}{c}{\textbf{Anomaly Preservation Fidelity}} 
& \multicolumn{2}{c}{\textbf{Downstream Quality}} \\
\cmidrule(lr){4-7} \cmidrule(lr){8-11} \cmidrule(lr){12-13}
& & 
& \textbf{T-Wass.} $\downarrow$
& \textbf{D-Wass.} $\downarrow$
& \textbf{S-Wass.} $\downarrow$
& \textbf{MMD} $\downarrow$
& \textbf{A-Rate.} $\downarrow$
& \textbf{A-Count.} $\downarrow$
& \textbf{A-Energy.} $\downarrow$
& \textbf{A-Tail.} $\downarrow$
& \textbf{Det-PRAUC.} $\uparrow$
& \textbf{Pred-PRAUC.} $\uparrow$ \\
\midrule
\multirow{13}{*}{\makecell{\textbf{FL1}\\2018\\-10}}
& \textbf{GAN}
& TimeGAN (2019)~\cite{yoon2019time}
& 0.1075  & 0.0574 & 0.1177 & 1.2467 & 0.1326 & 9.2364 & 1.8747 & 0.0226 & 0.0473 & 0.2273 \\

\arrayrulecolor{gray!60}\cmidrule(lr){2-13}\arrayrulecolor{black}

& \multirow{2}{*}{\textbf{VAE}}
& TimeVAE (2021)~\cite{desai2021timevae}
& 0.1168 & 0.0538 & 0.1389 & 1.2678 & 0.1115 & 7.0844 & 2.0471 & 0.0208 & 0.0502 & 0.2492 \\

&
& koVAE (2024)~\cite{naimangenerative}
& 0.1853 & 0.0803 & 0.2387 & 0.9129 & 0.1895 & 10.922 & 2.2580 & 0.0335 & 0.0483 &  0.1773 \\

\arrayrulecolor{gray!60}\cmidrule(lr){2-13}\arrayrulecolor{black}

& \textbf{Flow}
& F-Flow (2021)~\cite{alaa2021generative}
& 0.2328 & 0.1241 & 0.2945 & 1.5873 & 0.2346 & 12.974 & 2.8355 & 0.0628 & 0.0236 & 0.0884 \\

\arrayrulecolor{gray!60}\cmidrule(lr){2-13}\arrayrulecolor{black}

& \multirow{2}{*}{\textbf{Diffusion}}
& DiffWave (2021)~\cite{kongdiffwave}
& 0.2235 & 0.0993 & 0.1658 & 0.8326 & 0.0997 & 4.4152 & 1.1559 & 0.0373 & 0.0554 & 0.2547 \\

&
& Diffusion-TS (2024)~\cite{yuan2024diffusionts}
& 0.0356 & 0.0181 & 0.0313 & 0.2049 & 0.0427 & 3.9424 & 0.4483 & 0.0118 & \underline{0.0608} & 0.3354  \\

\arrayrulecolor{gray!60}\cmidrule(lr){2-13}\arrayrulecolor{black}

& \textbf{LLM}
& SDForger (2025)~\cite{rousseau2025forging}
 & 0.0386 & 0.0156 & 0.0345 & 0.2200 & 0.0377 & 3.0494 & 0.4218 & 0.0125 & 0.0607 & 0.3106 \\

\arrayrulecolor{gray!60}\cmidrule(lr){2-13}\arrayrulecolor{black}

& \multirow{2}{*}{\makecell{\textbf{Anomaly-}\\\textbf{Aware}}}
& FIDE (2024)~\cite{galib2024fide}
& 0.0473 & 0.0239  & 0.0371 & 0.2874 & 0.0367 & 4.2260 & 0.4877 & \underline{0.0113} & 0.0594 &  0.2855  \\

&
& HeavyDiff (2025)~\cite{pandey2025heavy}
& \underline{0.0234} & \textbf{0.0109} & \textbf{0.0206} & \underline{0.1662} & \underline{0.0161} & \underline{1.8122} & \underline{0.2631} & \textbf{0.0107} & 0.0591 & \underline{0.3411} \\

\arrayrulecolor{gray!60}\cmidrule(lr){2-13}\arrayrulecolor{black}

& \multirow{2}{*}{\makecell{\textbf{Elec.-}\\\textbf{Specific}}}
& Cond-Diff (2024)~\cite{fu2024creating}
& 0.2196 & 0.0842 & 0.3544 & 1.9436 & 0.2039 & 8.2390 & 1.7145 & 0.0349 & 0.0419 & 0.1552 \\
&
& CENTS (2025)~\cite{fuest2025cents}
& 0.0411 & 0.0235 & 0.0377 & 0.2745 & 0.4329 & 4.5110 & 0.5233 & 0.1491 & 0.0585 & 0.2930 \\

\arrayrulecolor{gray!60}\cmidrule(lr){2-13}\arrayrulecolor{black}

& \textbf{Ours}
& \textbf{\m}
& \textbf{0.0228} & \underline{0.0122} & \underline{0.0217} & \textbf{0.1569} & \textbf{0.0135} & \textbf{1.6491} & \textbf{0.2008} & \textbf{0.0107} & \textbf{0.0623} & \textbf{0.3529} \\
\bottomrule
\end{tabular}
\vspace{-5pt}
\end{table*}

\subsection{Generation Fidelity and Utility (RQ1--RQ3)}

To answer RQs 1--3, we evaluate \m from three perspectives: overall generation fidelity, anomaly preservation fidelity, and downstream utility. We report the results on the \textbf{FL1} dataset as a representative example, with additional results on the \textbf{FL2}, \textbf{NY}, and \textbf{CA} datasets provided in Appendix~\ref{appendix:more_result}. As shown in Table~\ref{tab:FL1}, \m achieves the best performance on two of the four overall fidelity metrics, all four anomaly fidelity metrics, and both downstream utility metrics. Compared with the strongest baseline results, \m improves anomaly preservation fidelity by an average of 12.21\% and downstream utility by 2.96\%. These results demonstrate that \m substantially improves anomaly generation quality and the utility of synthetic data for downstream tasks while maintaining competitive overall generation fidelity. 

Specifically, different categories of baselines exhibit clear performance differences. HeavyDiff is the strongest baseline overall, ranking first on two overall fidelity metrics and second on most remaining overall and anomaly fidelity metrics. Diffusion-TS and the LLM-based SDForger also achieve consistently strong results across multiple metrics. These observations suggest that diffusion-based and LLM-based generation are effective at modeling general temporal patterns, while explicit anomaly-aware modeling further improves the preservation of rare anomalous structures. In contrast, TimeGAN, TimeVAE, koVAE, and F-Flow exhibit relatively weak overall performance, suggesting that conventional GAN-, VAE-, and flow-based methods are less effective at modeling complex time-series data with sparse anomalies. Notably, the two energy-specific methods show inconsistent performance, further demonstrating that incorporating domain conditions alone is insufficient to ensure high-quality generation and that explicitly modeling and utilizing anomaly patterns are crucial for preserving both overall fidelity and anomalous patterns.

\vspace{-3pt}
\subsection{Anomaly-Preservation Visualization (RQ4)}
\begin{figure}[htbp]
\centering
% ==================== First row ====================
\begin{minipage}{0.48\linewidth}
    \centering
    \captionsetup{skip=3pt}
    \includegraphics[width=\linewidth]
    {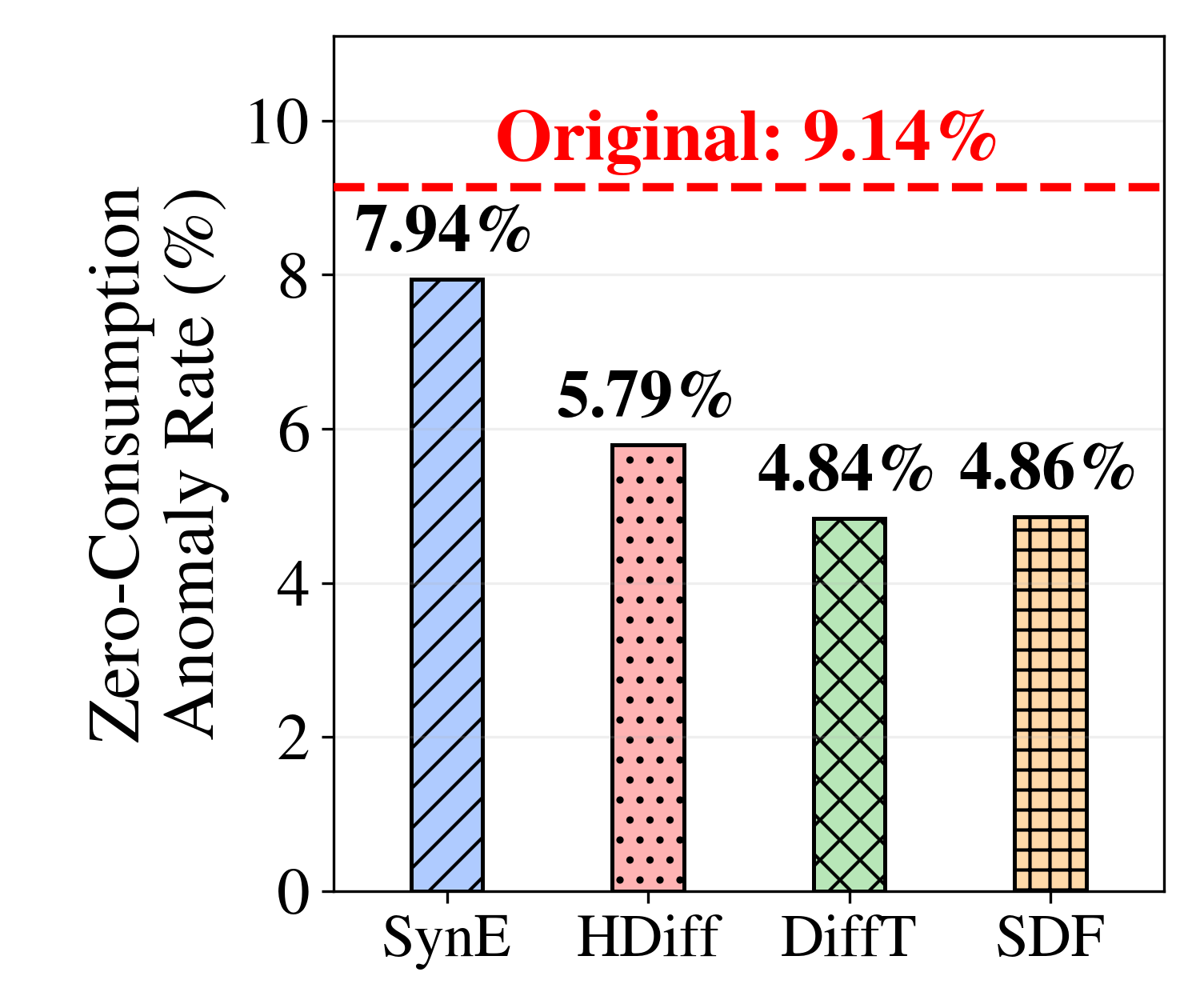}
    \captionof{figure}{Anomaly Rate.}
    \label{fig:zero_consumption_anomaly_rate}
\end{minipage}
\hfill
\begin{minipage}{0.48\linewidth}
    \centering
    \captionsetup{skip=3pt}
    \includegraphics[width=\linewidth]
    {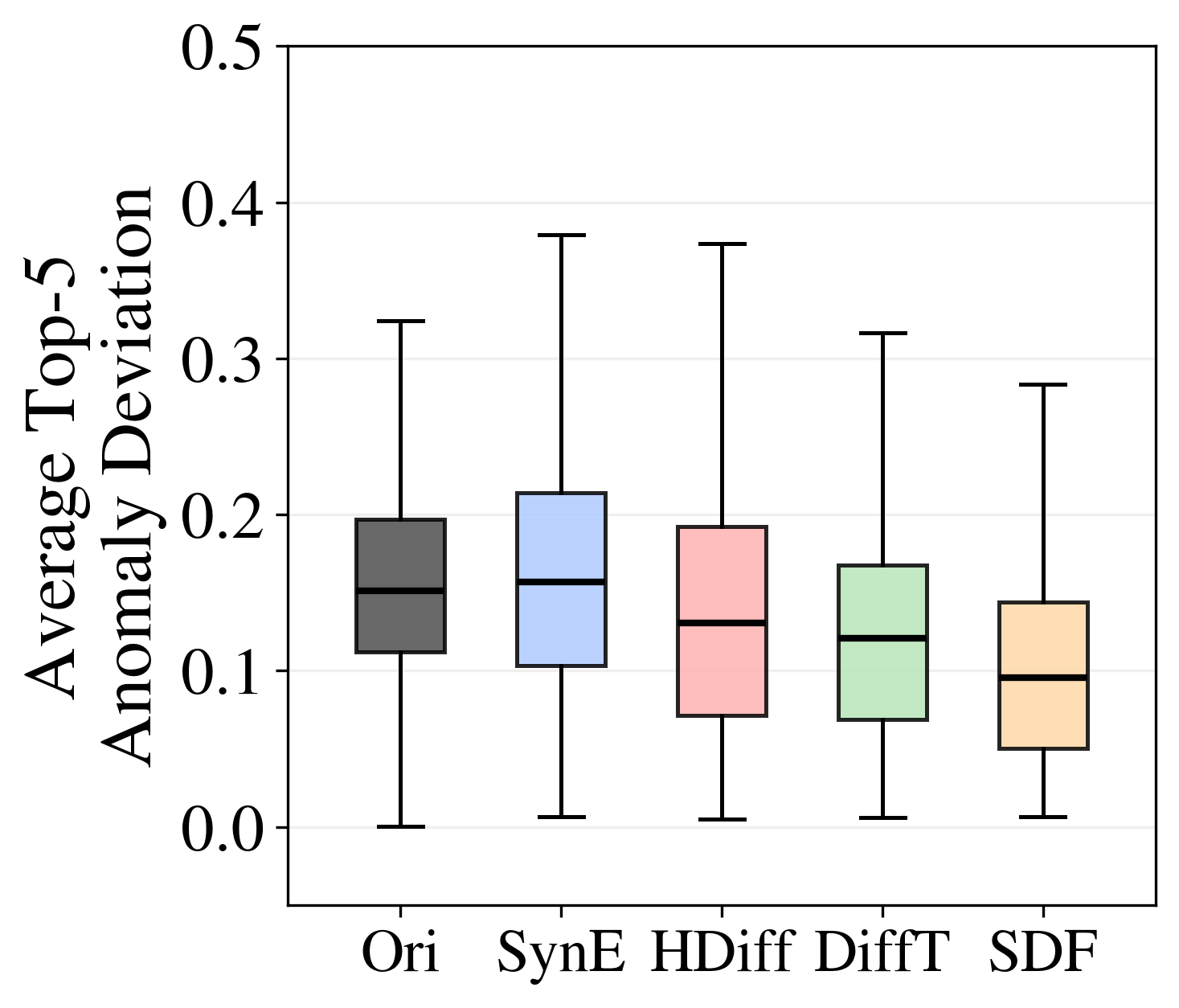}
    \captionof{figure}{Anomaly Deviation.}
    \label{fig:anomaly_deviation_boxplot}
\end{minipage}

\vspace{2pt}

% ==================== Second row ====================
\begin{minipage}{0.48\linewidth}
    \centering
    \captionsetup{skip=3pt}
    \includegraphics[width=\linewidth]
    {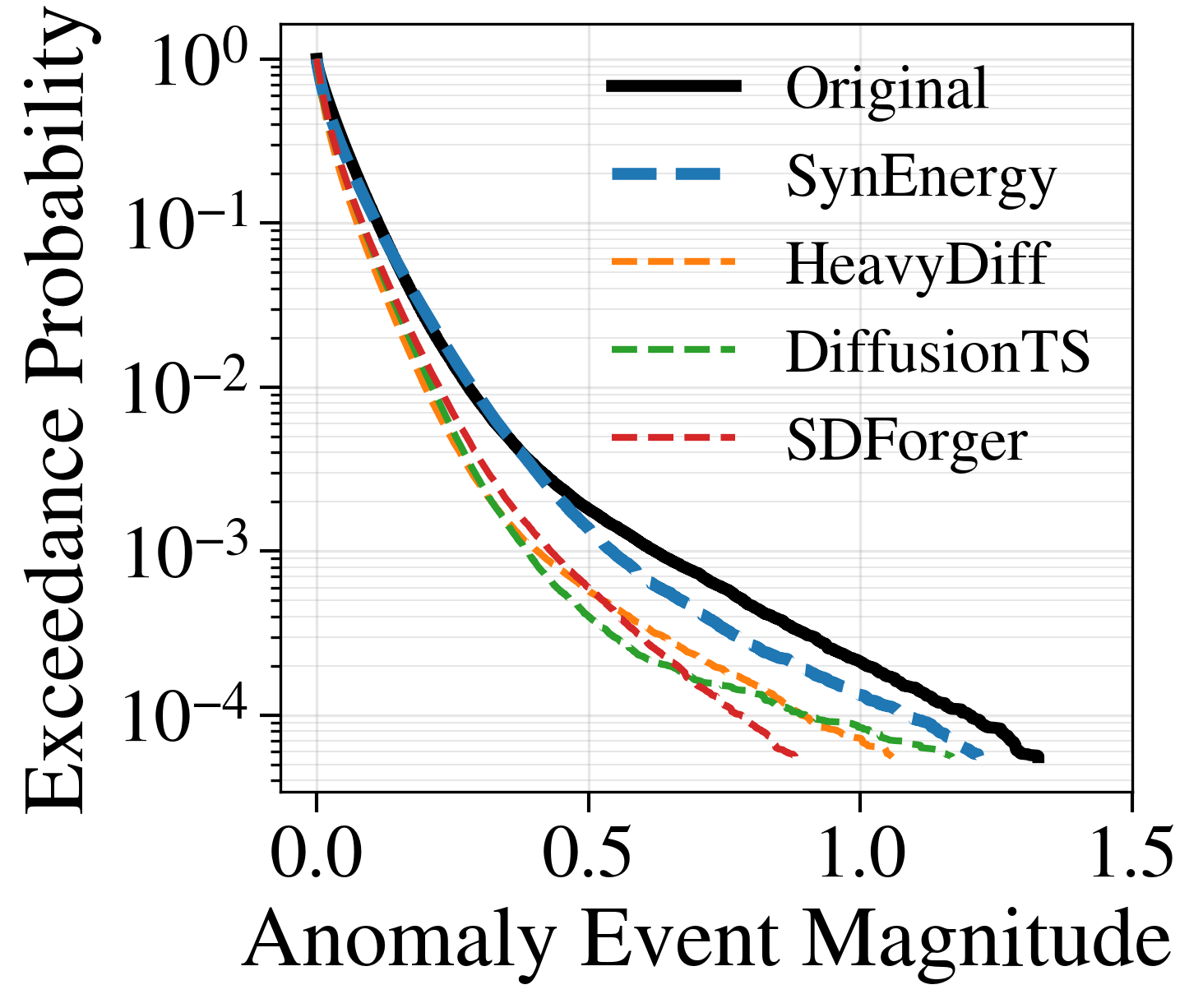}
    \captionof{figure}{CCDF of Anomalous Event Magnitudes.}
    \label{fig:anomaly_magnitude}
\end{minipage}
\hfill
\begin{minipage}{0.48\linewidth}
    \centering
    \captionsetup{skip=3pt}
    \includegraphics[width=\linewidth]
    {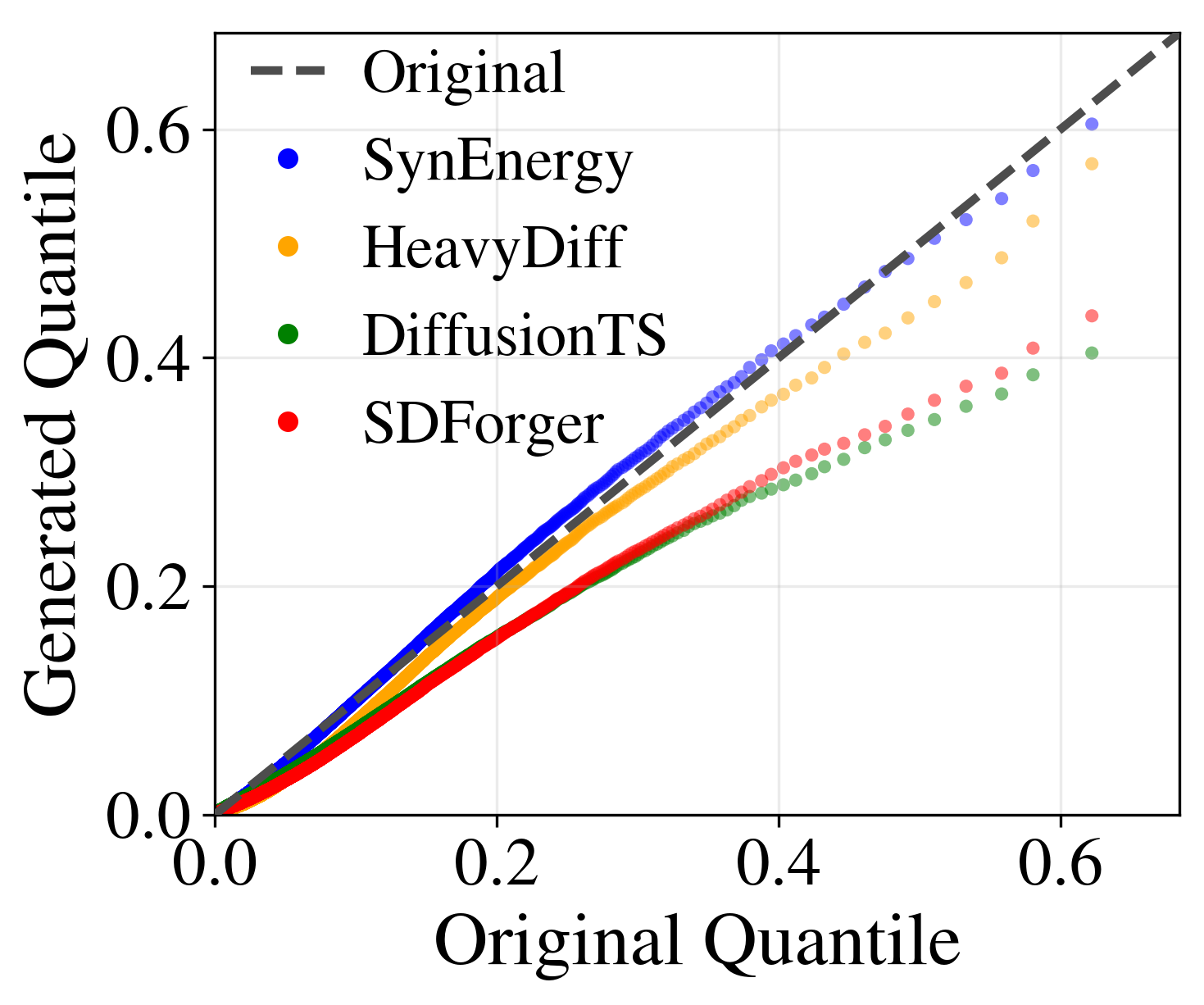}
    \captionof{figure}{Q--Q Comparison of Anomalous Event Energy.}
    \label{fig:qq_anomaly_rate}
\end{minipage}
\vspace{-10pt}
\end{figure}

To answer RQ4, we visualize anomaly-preserving performance from multiple perspectives, including anomaly rate, deviation, magnitude, and energy. We compare \m with the three strongest baselines and use abbreviated names in several figures: SynE for SynEnergy, HDiff for HeavyDiff, DiffT for Diffusion-TS, and SDF for SDForger. Figure~\ref{fig:zero_consumption_anomaly_rate} presents the zero-consumption anomaly rate, defined as the proportion of zero-consumption anomaly points in each sequence. \m achieves a rate of 7.94\%, substantially closer to the original 9.14\% than the baselines.
Figure~\ref{fig:anomaly_deviation_boxplot} shows boxplots of the mean top-5 anomaly deviations. For each anomalous point, the deviation is measured relative to the average of its surrounding points, and the five largest deviations in each sequence are averaged. The distribution generated by \m closely matches the original data, with a nearly identical average deviation.
Figures~\ref{fig:anomaly_magnitude} and~\ref{fig:qq_anomaly_rate} compare anomalous-event magnitude and energy using CCDF~\cite{embrechts2013modelling} and Q--Q plots~\cite{wilk1968probability}, respectively. Event magnitude is the maximum absolute residual within an event, while event energy is the cumulative exceedance beyond the anomaly threshold over its duration. In both cases, \m most closely reproduces the original distributions.
Overall, the baselines consistently underrepresent anomalous patterns, whereas \m preserves them across multiple perspectives. Additional results including anomaly duration and event count are provided in Appendix~\ref{appendix:anomaly_preserving_performance}.

\vspace{-3pt}

\subsection{Anomalous Event Preservation (RQ5)} 
\begin{figure}[t]
    \vspace{-5pt}
    \centering
    \begin{subfigure}[t]{\linewidth}
        \centering
        \includegraphics[width=0.90\linewidth]{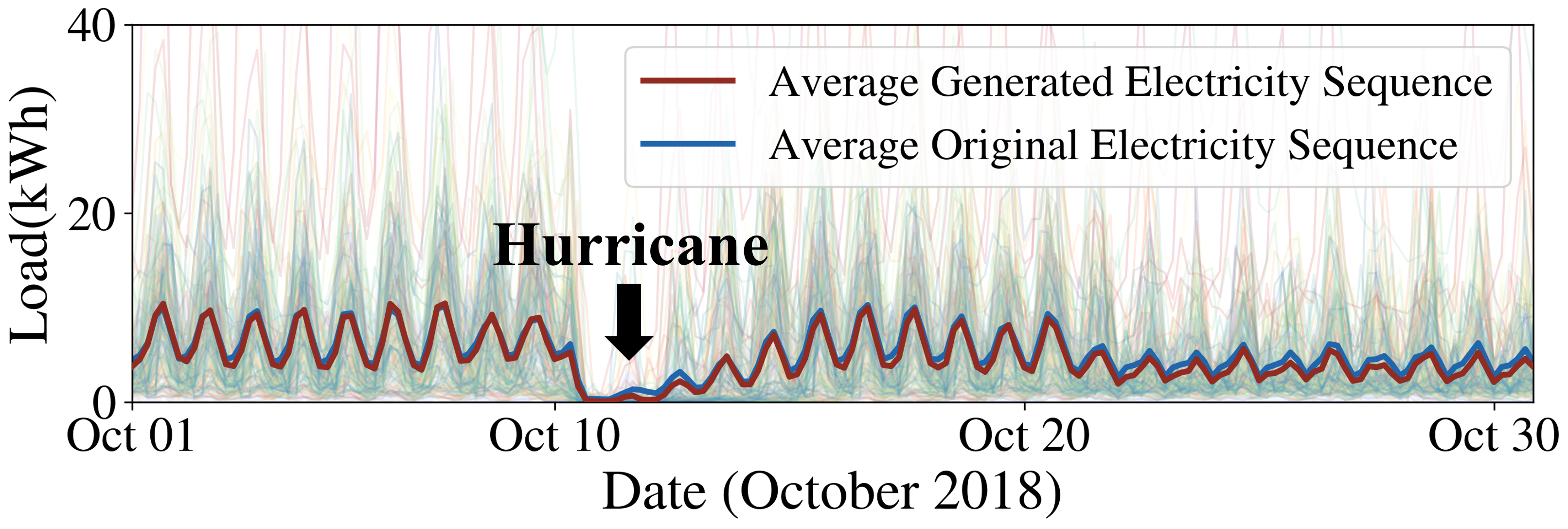}
        \label{fig:F1P}
    \end{subfigure}
    \vspace{-5pt}
    \begin{subfigure}[t]{\linewidth}
        \centering
        \includegraphics[width=0.90\linewidth]{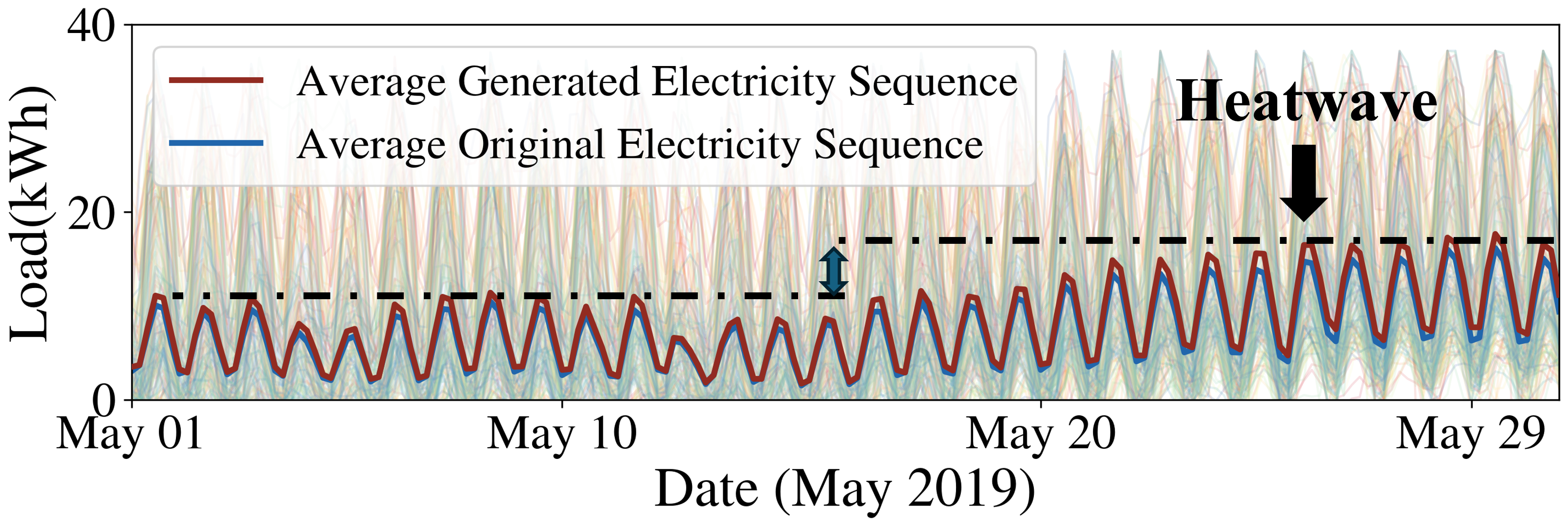}
        \label{fig:F2P}
    \end{subfigure}
    \vspace{-12pt}
    \caption{Visualization of the synthetic data generated by \m for large-scale anomalous event preservation.}
    \label{fig:large_scale_event_preservation}
    \vspace{-15pt}
\end{figure}

To answer RQ5, we examine how well \m preserves two large-scale anomalous events in October 2018 and May 2019. In Figure~\ref{fig:large_scale_event_preservation}, the red and blue lines represent the mean consumption of the generated and original data, respectively. In October 2018, \m reproduces the widespread power outage following Hurricane Michael on October 10, with approximately 87\% of sampled households exhibiting prolonged zero-consumption anomalies, closely matching the original data. In May 2019, \m captures the collective high-consumption pattern associated with the late-May heatwave, as consumption after May 20 is noticeably higher than before May 15. These results demonstrate that \m preserves both low- and high-consumption large-scale anomalous events, including their timing and household-level impact.

\vspace{-5pt}
\subsection{Further Analysis (RQ6--RQ9)}
For RQ6, we investigate the generation quality of \m across different scales and temporal granularities. We consider four scales with 100, 1,000, 10,000, and 50,000 households and three granularities of 1 hour, 4 hours, and 1 day. Across all settings, \m achieves the best performance on at least 7 of the 10 metrics, demonstrating consistent robustness to changes in scale and temporal granularity. Detailed results are provided in Appendix~\ref{appendix:scalability}.

For RQ7, we examine the sensitivity of \m to the anomaly threshold $\delta$, defined using the top 1\%, 5\%, 10\%, and 20\% of absolute residuals. Across all settings, \m achieves the best performance on at least 5 of the 10 metrics. The 5\% and 10\% settings yield the strongest results, ranking first on 9 and 8 metrics, respectively. Performance decreases when the anomaly definition is either too restrictive at 1\% or too broad at 20\%, indicating that moderate thresholds provide more effective guidance for anomaly-preserving generation. Detailed results are provided in Appendix~\ref{appendix:threshold}.

For RQ8, we conduct an ablation study on two key components of \m. First, \textbf{\m w/o HG-ASL} replaces HG-ASL with a simple encoder that directly maps anomalous residuals into diffusion conditions, removing explicit anomaly semantic learning and cross-region modeling. Second, \textbf{\m w/o AG} removes the entire anomaly-guidance pathway and retains only the Backbone Denoiser, where AG denotes Anomaly Guidance. Both variants substantially degrade performance, particularly on anomaly preservation metrics, confirming the importance of anomaly semantic representation and its dedicated injection into the diffusion process. Detailed results are provided in Appendix~\ref{appendix:ablation}.

For RQ9, we evaluate the training and sampling efficiency of \m. On FL1 and FL2, training for 10,000 optimization steps takes approximately 20 minutes, while generating 10,000 samples requires 31.7 minutes. On NY and CA, the corresponding times are approximately 8.3 and 13.3 minutes, respectively, due primarily to their shorter sequence lengths. These results demonstrate the practical efficiency of \m across datasets. Detailed results are provided in Appendix~\ref{appendix:efficiency}.

\section{Related Work}\label{sec:related_work}
We review related work from three perspectives. \textbf{Synthetic energy data generation} has advanced from statistical and behavior-based modeling to deep generative approaches, including GANs and diffusion models. However, existing methods primarily emphasize overall distributional and temporal fidelity rather than anomaly preservation~\cite{kang2023systematic,thorve2023high,yuan2023synthetic,hu2023multiload,razghandi2023smart,fu2024creating,fuest2025cents}. \textbf{Anomaly-aware time-series generation} aims to preserve rare and extreme patterns, but most methods target general time series and do not account for the strong temporal periodicity and spatial correlations inherent in energy consumption data~\cite{allouche2022ev,hasan2022modeling,finzi2023user,shariatian2025denoising,pandey2025heavy,galib2024fide,jiang2026e4gen}. \textbf{Diffusion-based time-series generation} provides effective temporal modeling, but existing methods lack a dedicated design for incorporating anomaly semantics into the generation process~\cite{yuan2024diffusionts,kollovieh2023predict,bilovs2023modeling,narasimhan2024time,yan2024probabilistic,naiman2024utilizing,li2025population}. Collectively, these limitations motivate the design of \m. Detailed discussions are provided in Appendix~\ref{appendix:related_work}.

\section{Conclusion}\label{sec:Conclusion}
In this paper, we propose \m, an anomaly semantic-guided diffusion framework for generating realistic synthetic energy consumption data while preserving rare and complex anomalous events. The design of \m is motivated by empirical observations that energy consumption anomalies exhibit structured patterns shaped by geographical proximity and regional attribute similarity. \m consists of two core components: (i) HG-ASL, which learns structured anomaly semantic representations from sparse residual patterns and cross-region dependencies; and (ii) AS-Diff, which injects these representations into the diffusion denoising process to guide anomaly-preserving generation. Extensive experiments on four real-world datasets from Florida, New York, and California demonstrate the superiority of \m over 11 general-purpose and energy-specific generation baselines. Compared with the strongest baselines, \m improves anomaly preservation fidelity by an average of 12.21\% and downstream quality by 2.96\%, while maintaining competitive overall generation fidelity.

\section*{Limitations and Ethical Considerations}
The current work has two limitations. 
% First, it focuses on energy consumption data generation; future work could extend the framework to other time-series domains in which anomalies are shaped by geographical proximity and attribute similarity. 
First, the regional conditions incorporate only a limited set of attributes and may not fully capture regional heterogeneity. Enriching them with additional contextual information may further improve region-specific generation. Second, although synthetic data can reduce the direct exposure of household records, potential privacy risks should still be carefully evaluated before data release or downstream use.

We adhere to the KDD Code of Ethics. De-identified data were obtained from a municipal utility provider in Florida under a non-disclosure agreement and are stored and processed only on FSU's secure computing facilities. We declare no conflicts of interest.

\section*{GenAI Disclosure}
In the preparation of this work, the authors utilized Generative AI tools solely for the purpose of language refinement and improving readability. No AI tools were used to generate scientific concepts, experimental results, or the intellectual content of this paper. The authors have reviewed all AI-assisted edits and take full responsibility for the final content of the manuscript.

% \section*{GenAI Disclosure}
% In the preparation of this work, the authors utilized Generative AI tools solely for the purpose of language refinement and improving readability. No AI tools were used to generate scientific concepts, experimental results, or the intellectual content of this paper. The authors have reviewed all AI-assisted edits and take full responsibility for the final content of the manuscript.

\bibliographystyle{ACM-Reference-Format}
\balance
\bibliography{reference}

\newpage
\appendix
\section*{Appendix} \label{Sec:Appendix}

\section{Interdisciplinary Collaboration} \label{appendix:collaboration}
This work benefits from the cross-domain expertise of Dr. Ravikumar Gelli, whose contributions were central to the interdisciplinary collaboration between energy systems and artificial intelligence. Dr. Gelli is an Associate Professor in the Department of Electrical and Computer Engineering at the FAMU–FSU College of Engineering, Florida State University, where he leads the GridAI Lab and is affiliated with the Center for Advanced Power Systems. His research focuses on power systems, artificial intelligence for energy applications, and the security of smart-grid cyber-physical systems. In this work, he provided domain expertise in AI-enabled energy systems, clarified the characteristics of energy consumption data and associated anomalous events, and identified the challenges of preserving such events in synthetic data generation. These insights helped shape the research motivation and evaluation design.

% Dr. Guang Wang contributes domain expertise in data-driven urban and energy systems. He is an Assistant Professor in the Department of Computer Science at Florida State University and a collaborator with the Resilient Infrastructure $\&$ Disaster Response Center. His research spans spatiotemporal data mining, cyber-physical systems, and socially relevant applications in energy, urban mobility, and disaster resilience. He has conducted extensive research on city-scale systems, including electric-vehicle networks, urban mobility, electricity-related disruptions, and equitable power restoration, providing expertise in the analysis and modeling of large-scale urban energy data. He also established the collaboration with the municipal utility provider in Tallahassee and facilitated access to the energy consumption data used in this study.

\section{Data-Driven Analysis Details}
\subsection{Empirical Analysis of Regional Attributes and Anomaly Rates} \label{appendix:attribute}

\begin{figure}[htbp]
\centering

\begin{minipage}{0.48\linewidth}
    \centering
    \includegraphics[width=\linewidth]{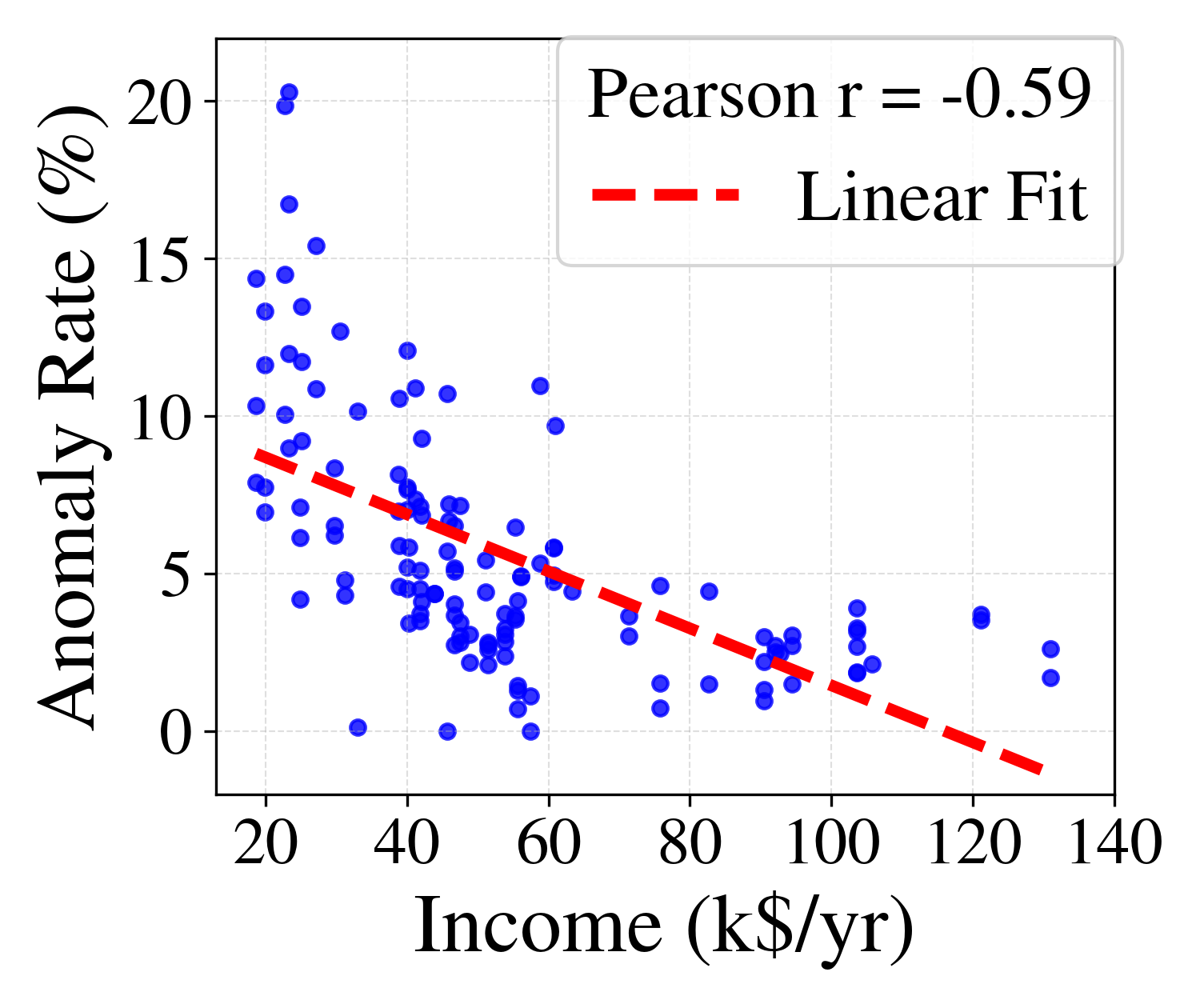}
    \captionof{figure}{Region Median Income.}
    \label{fig:income_anomaly_rate}
\end{minipage}
\hfill
\begin{minipage}{0.48\linewidth}
    \centering
    \includegraphics[width=\linewidth]{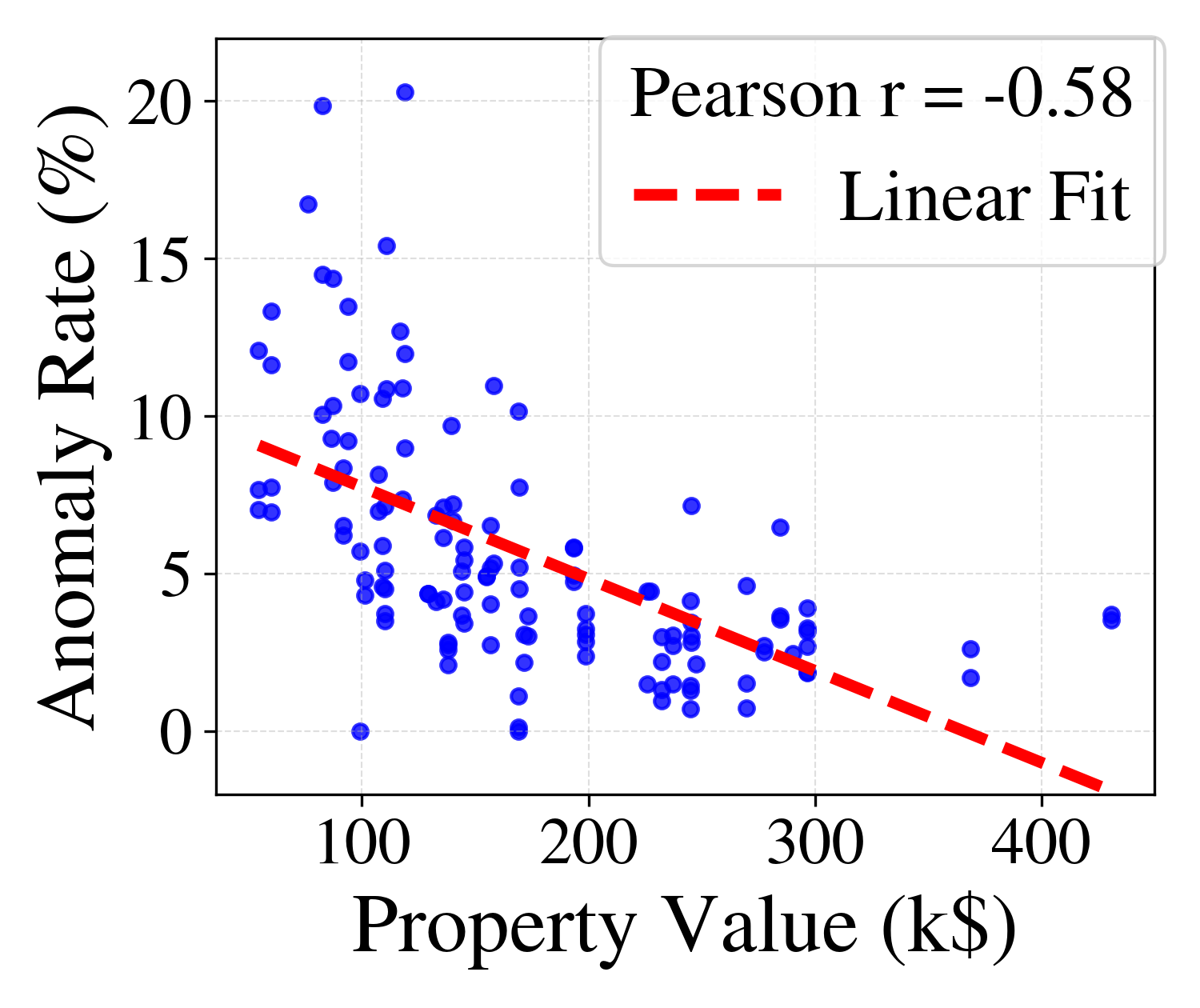}
    \captionof{figure}{Region Median Property Value.}
    \label{fig:property_value_anomaly_rate}
\end{minipage}

\vspace{6pt}

\begin{minipage}{0.48\linewidth}
    \centering
    \includegraphics[width=\linewidth]{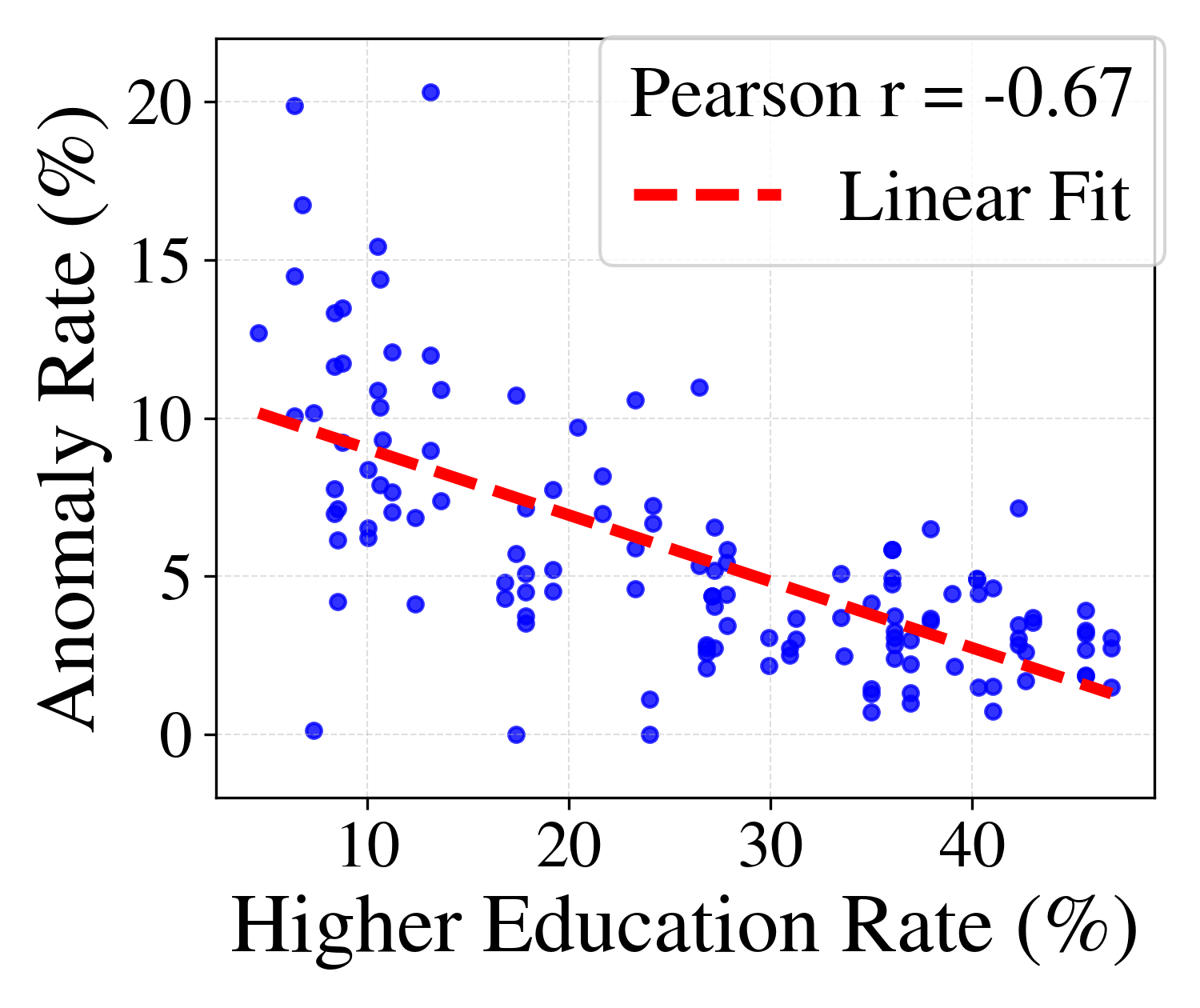}
    \captionof{figure}{Region Higher Education Rate.}
    \label{fig:education_anomaly_rate}
\end{minipage}
\hfill
\begin{minipage}{0.48\linewidth}
    \centering
    \includegraphics[width=\linewidth]{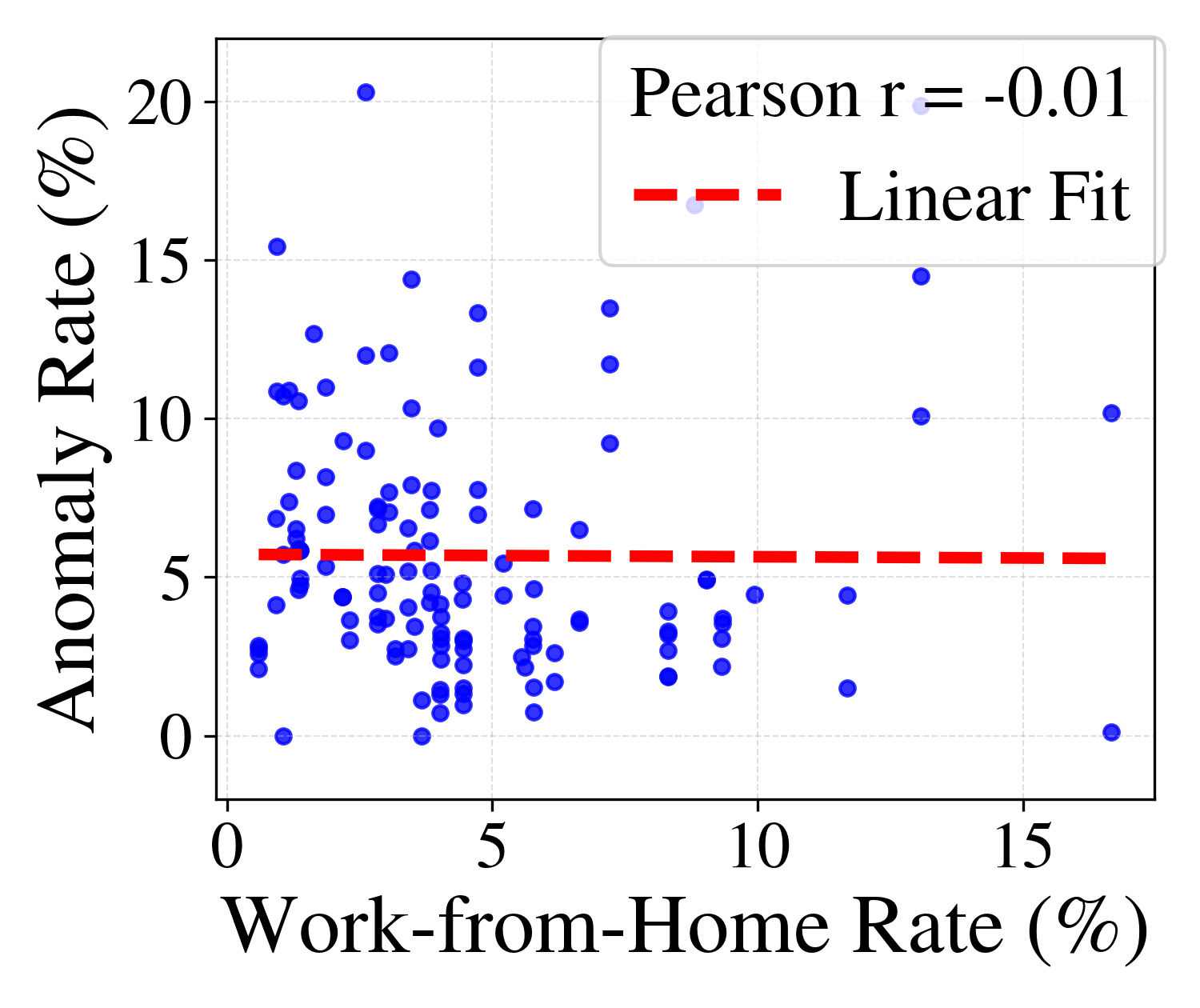}
    \captionof{figure}{Region Remote Work Rate.}
    \label{fig:remote_work_anomaly_rate}
\end{minipage}

\end{figure}

To examine the relationship between regional attributes and energy consumption anomalies, we select four representative socioeconomic characteristics as illustrative examples: median household income, median property value, the share of residents with a bachelor's degree or higher, and the work-from-home rate. In each figure, the horizontal axis represents the corresponding regional attribute, while the vertical axis shows the regional anomaly rate. As shown in Figures~\ref{fig:income_anomaly_rate}--\ref{fig:education_anomaly_rate}, median household income, median property value, and higher-education rate exhibit significant negative correlations with anomaly rates, with Pearson correlation coefficients of $-0.59$, $-0.58$, and $-0.67$, respectively. These results suggest that regions with better socioeconomic conditions generally experience lower energy consumption anomaly rates, potentially reflecting better-maintained electricity infrastructure and more reliable service in affluent areas.

In contrast, the work-from-home rate in Figure~\ref{fig:remote_work_anomaly_rate} shows a negligible correlation with the regional anomaly rate, with a Pearson correlation coefficient of only $-0.01$, indicating that this attribute offers limited explanatory value for regional anomaly patterns. This result suggests that the relationships between regional attributes and energy consumption anomalies are attribute-specific rather than universal. Overall, regions with similar anomaly-related socioeconomic characteristics may exhibit comparable anomaly patterns despite being geographically distant. Motivated by this observation, \m constructs an attribute graph that encodes attribute similarity as structural prior information, guiding the model to place greater emphasis on regions that share anomaly-relevant attributes rather than mere spatial proximity.

\subsection{Limitations of Existing Methods in Preserving Anomalous Events}\label{appendix:limitation}

To motivate the design of \m, we empirically evaluate how well existing time-series generation methods preserve fine-grained temporal variations and anomalous events. We use Diffusion-TS~\cite{yuan2024diffusionts}, a state-of-the-art time-series generation framework published in ICLR 2024, as a representative baseline and examine its performance from four complementary perspectives.

\begin{figure}[t]
\centering
\begin{minipage}{0.48\linewidth}
    \centering
    \includegraphics[width=\linewidth]{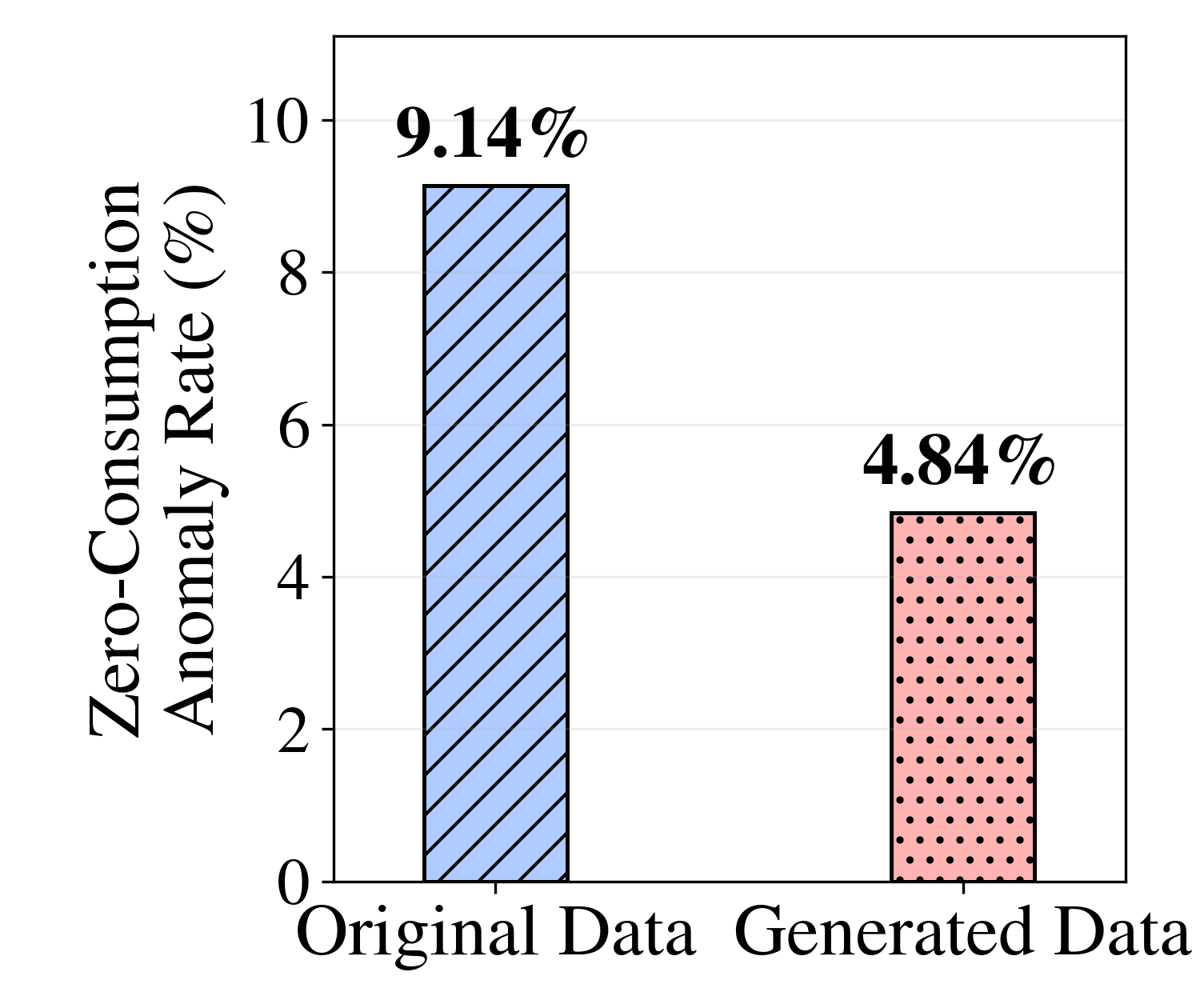}
    \captionof{figure}{Comparison of Zero-Consumption Rates.}
    \label{fig:zero_consumption_rate}
\end{minipage}
\hfill
\begin{minipage}{0.48\linewidth}
    \centering
    \includegraphics[width=\linewidth]{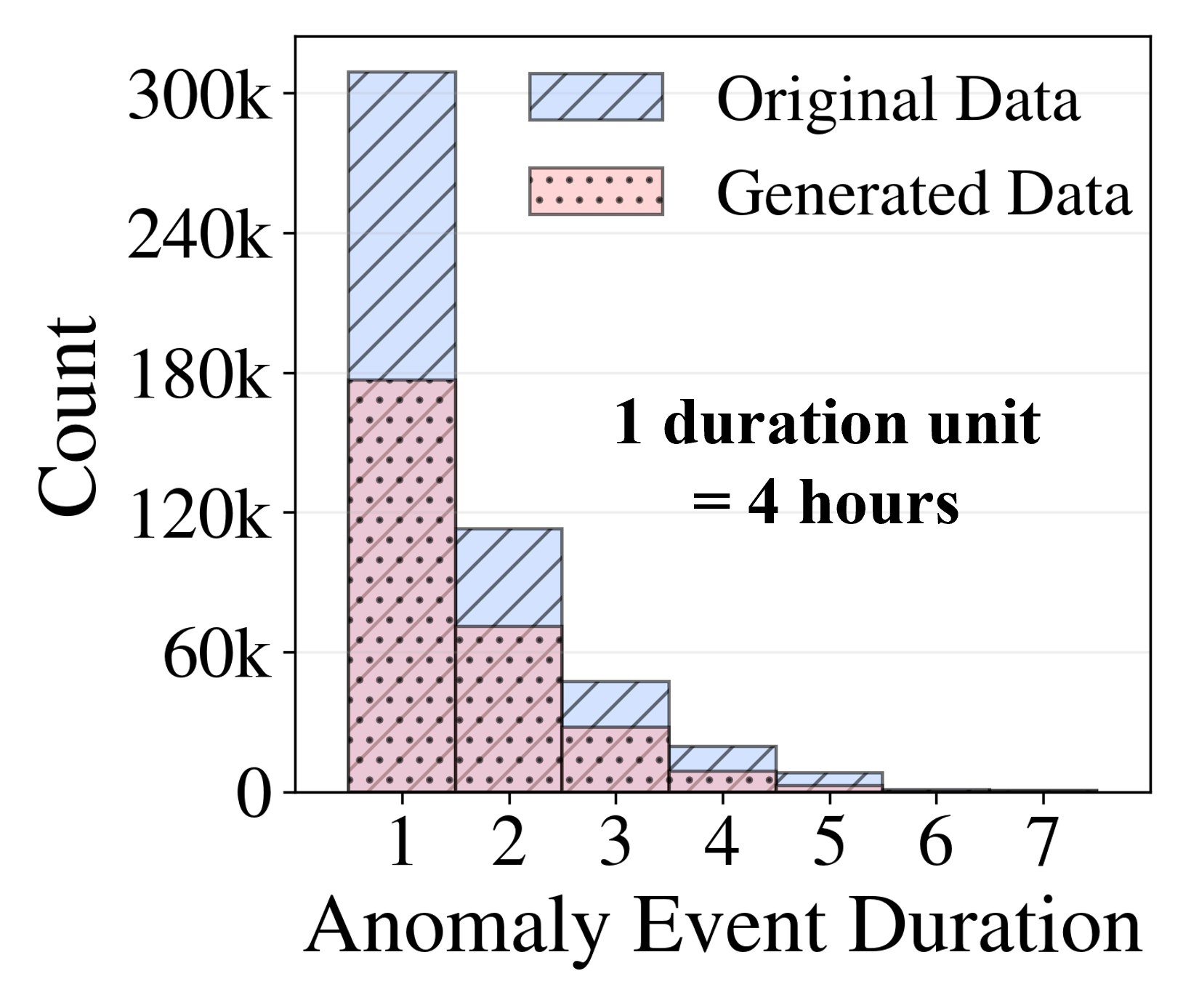}
    \captionof{figure}{Comparison of Anomaly Event Durations.}
    \label{fig:anomaly_duration_distribution}
\end{minipage}
\end{figure}

First, we evaluate the ability of existing time-series generation methods to generate \textbf{zero-consumption anomalies}. In this study, a zero-consumption anomaly refers to a four-hour period during which a household records no energy consumption. Such anomalies often reflect large-scale power outages or temporary household absences, such as business travel. Preserving these events is important for maintaining the realism and diversity of large-scale synthetic energy consumption data. As shown in Figure~\ref{fig:zero_consumption_rate}, the zero-consumption anomaly rate decreases from 9.14\% in the original data to 4.84\% in the generated data, corresponding to a relative reduction of approximately 47.0\%. This substantial decline demonstrates that current time-series generation methods considerably underrepresent zero-consumption anomalies and fail to preserve their occurrence frequency.

Second, we evaluate how well current time-series generation methods preserve \textbf{anomalous event duration}. An anomalous event is defined as a sequence of consecutive anomalous intervals, and its duration is measured by the number of intervals it spans. Since the temporal granularity is four hours, each duration unit corresponds to four hours. As shown in Figure~\ref{fig:anomaly_duration_distribution}, both the original and generated data are dominated by short-duration events. However, the generated data contain substantially fewer events across nearly all duration levels, with the discrepancy becoming more pronounced for longer-lasting anomalies. This result indicates that current generation methods not only reduce the number of anomalous events but also underrepresent their temporal persistence.

\begin{figure}[t]
\centering
\begin{minipage}{0.48\linewidth}
    \centering
    \includegraphics[width=\linewidth]{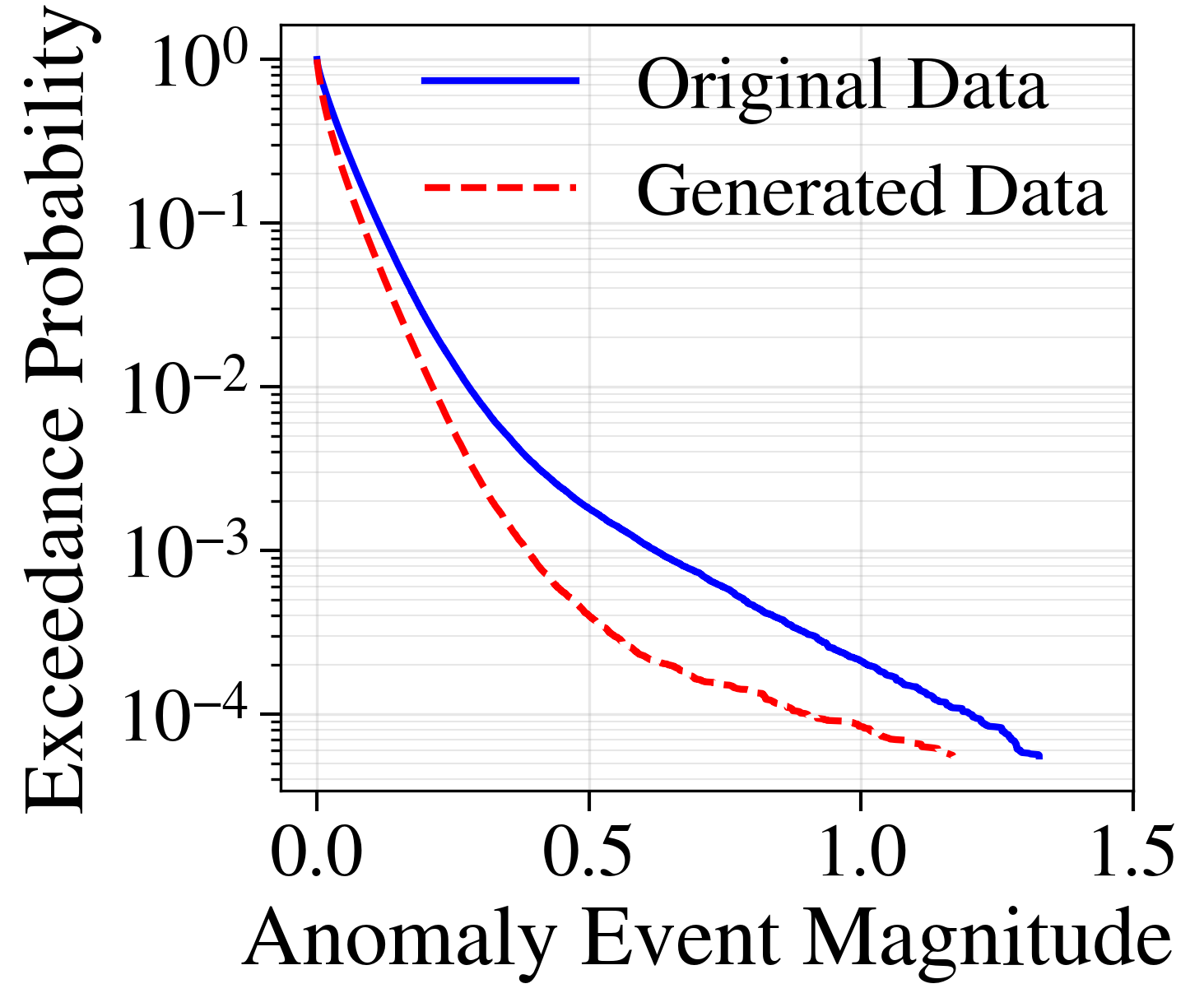}
    \captionof{figure}{CCDF of Anomaly Event Magnitudes.}
    \label{fig:anomaly_magnitude_distribution}
\end{minipage}
\hfill
\begin{minipage}{0.48\linewidth}
    \centering
    \includegraphics[width=\linewidth]{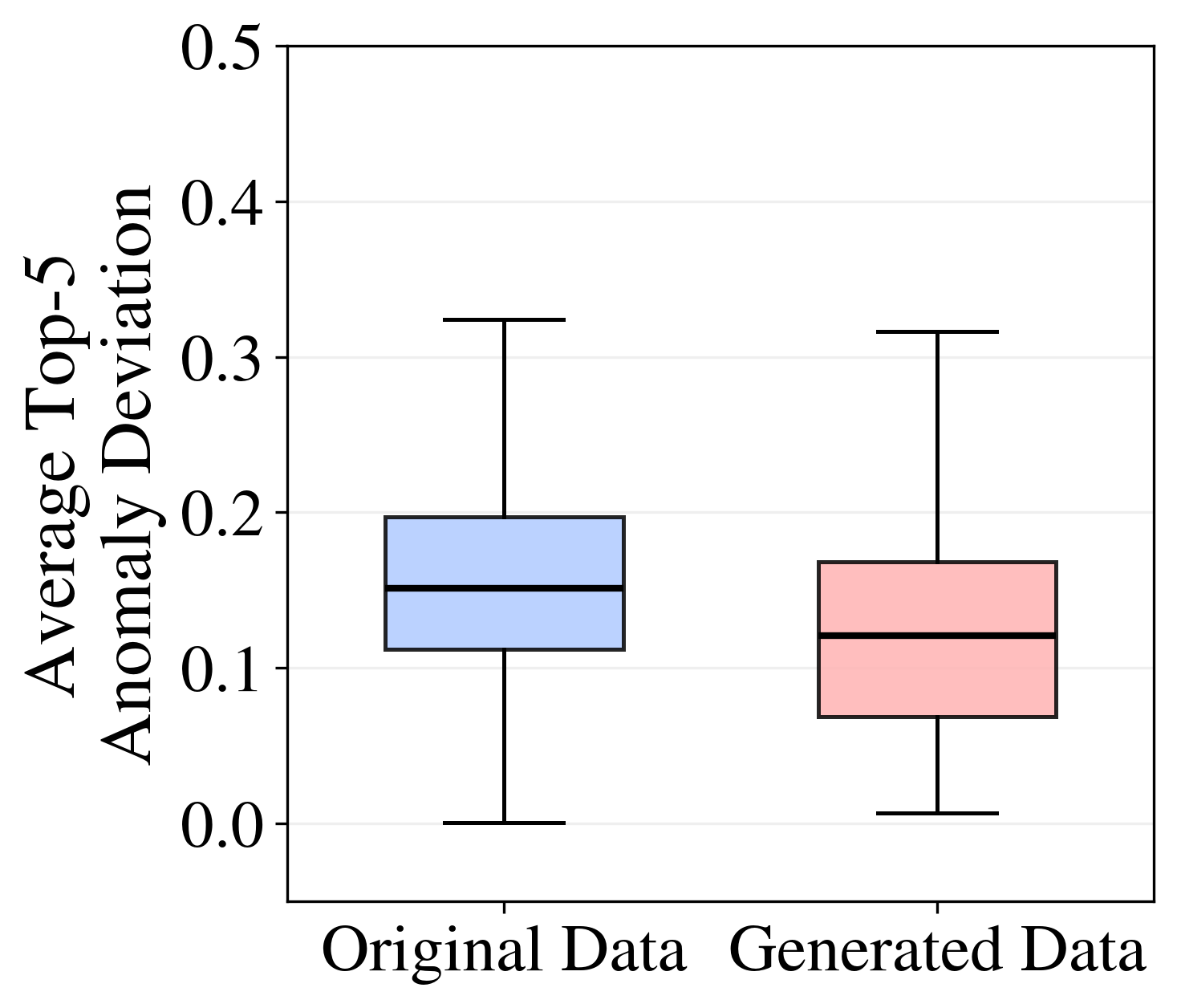}
    \captionof{figure}{Boxplot Comparison of Anomaly Deviations.}
    \label{fig:anomaly_deviation_distribution}
\end{minipage}
\end{figure}

Third, we evaluate how well current time-series generation methods preserve \textbf{anomalous event magnitude}. We quantify anomaly magnitude as the deviation of each anomalous event from its local temporal baseline and compare the resulting distributions using the complementary cumulative distribution function (CCDF). As shown in Figure~\ref{fig:anomaly_magnitude_distribution}, the generated data consistently exhibit lower exceedance probabilities than the original data across the entire magnitude range. This underrepresentation becomes increasingly pronounced toward the tail, where larger-magnitude anomalous events are substantially less frequent in the generated data.

Fourth, we evaluate how well current time-series generation methods preserve \textbf{local anomaly deviations}. For each household consumption sequence, we measure the increase at each time point relative to its surrounding seven-point temporal window and retain the five largest deviations. We then average these deviations to characterize the intensity of local anomalous fluctuations. As shown in Figure~\ref{fig:anomaly_deviation_distribution}, the generated data exhibit a noticeably lower median and a downward-shifted overall distribution compared with the original data. This difference indicates that current generation methods tend to produce weaker local anomalous fluctuations and underrepresent the intensity of pronounced deviations.

Overall, these results show that current time-series generation methods fail to faithfully preserve anomalous events in terms of their occurrence frequency, duration, magnitude, and local deviation intensity. Collectively, these findings demonstrate that such methods often overlook fine-grained temporal variations and systematically underrepresent anomalous events.

\section{Methods Details}
\subsection{Anomalous Event Identification Process} \label{appendix:event_identification}

\begin{figure}[htbp]
    \centering
    \includegraphics[width=\linewidth]{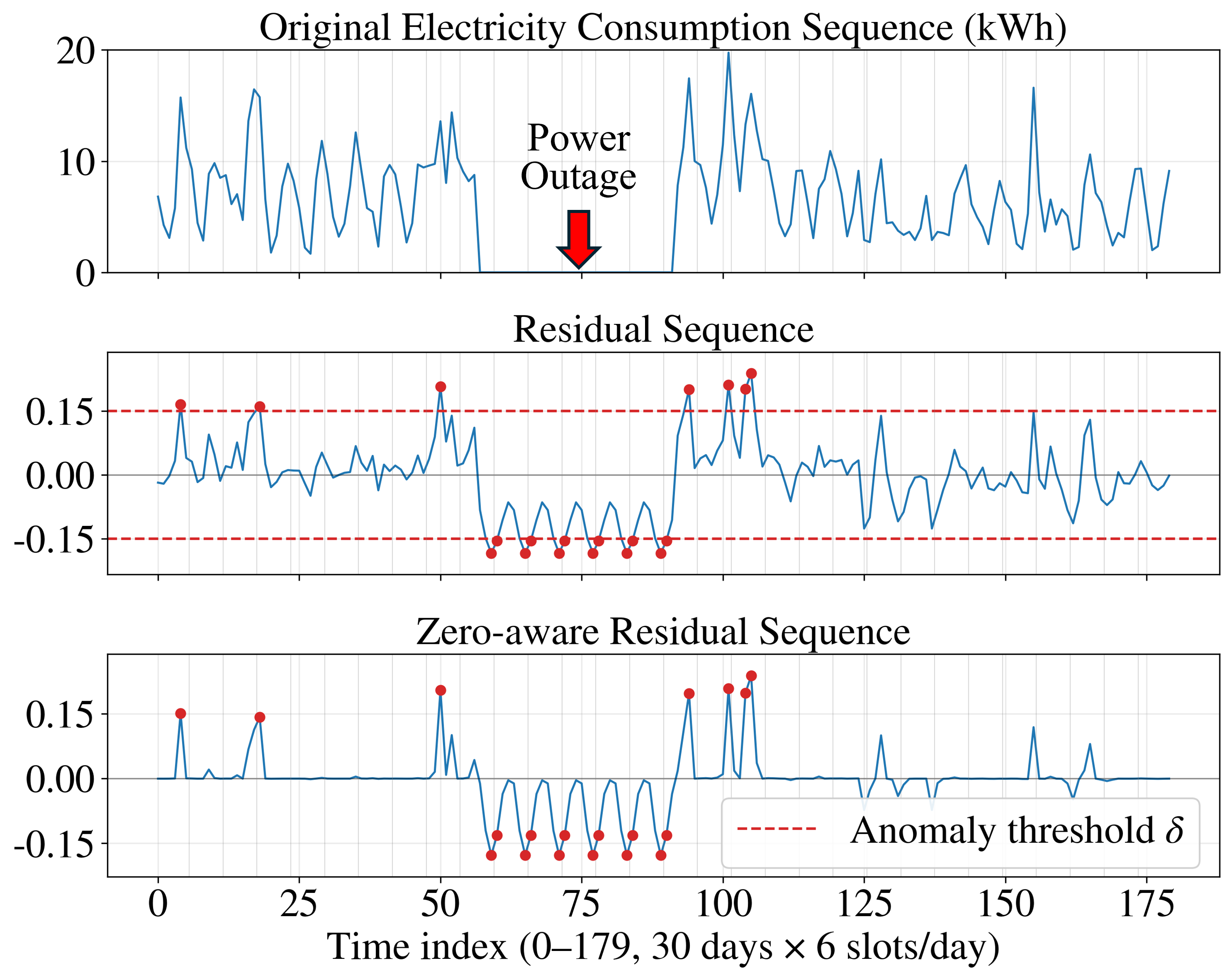}
    \caption{Anomalous Event Identification Process}
    \label{fig:anomaly_identify}
\end{figure}

Figure~\ref{fig:anomaly_identify} illustrates the process of identifying anomalous events from raw energy consumption data. The first panel shows the original consumption sequence of a representative household that experienced a nearly five-day power outage during the hurricane. The second panel presents the residual sequence $\mathbf{e}_{n,m}$ defined in Section~\ref{sec:Anomalous_event_defintion}, obtained by removing the regional background pattern. Using the anomaly threshold $\delta$, intervals with residuals above $\delta$ or below $-\delta$ are identified as high- or low-consumption anomalous events, respectively. The third panel shows the zero-aware filtered residual sequence $\mathbf{e}^{f}_{n,m}$ defined in Equation~\ref{eq:2}, where near-zero residuals are suppressed while informative anomalous deviations are preserved. This increases the relative prominence of anomaly-related information and facilitates subsequent anomaly semantic representation.

\subsection{Design of the Zero-Aware Encoder in SR-Encoder}
\label{appendix:zero_aware_encoder}

The zero-aware encoder $E_{\theta}$ maps each filtered residual sequence $\mathbf{e}^{f}_{n,m}$ to a compact anomaly embedding $\mathbf{z}_{n,m}$. Since anomalous deviations are sparse and most residuals remain close to zero, directly encoding the residual sequence can cause normal intervals to dominate the learned representation. To address this issue, $E_{\theta}$ jointly processes $\mathbf{e}^{f}_{n,m}$ and its corresponding soft-weight sequence $\mathbf{w}^{e}_{n,m}$, thereby emphasizing informative high- and low-consumption deviations while suppressing near-zero residuals associated with normal consumption patterns.

Specifically, we concatenate $\mathbf{e}^{f}_{n,m}$ and $\mathbf{w}^{e}_{n,m}$ along the feature dimension and feed the resulting two-channel sequence into three one-dimensional temporal convolutional layers. Each layer uses a kernel size of 5 and preserves the original sequence length. The convolutional stack captures local anomaly structures, including isolated spikes, sustained deviations, and abrupt sign changes, and produces a temporal feature map $\mathbf{H}_{n,m}\in\mathbb{R}^{d_h\times K}$.

We aggregate $\mathbf{H}_{n,m}$ along the temporal dimension using average, max, and soft-weighted pooling:
\begin{equation}
\mathbf{h}^{\mathrm{avg}}_{n,m}
= \frac{1}{K}\sum_{k=1}^{K}\mathbf{H}_{n,m}^{k},
\qquad
\mathbf{h}^{\mathrm{max}}_{n,m}
= \max_{k\in\{1,\ldots,K\}}\mathbf{H}_{n,m}^{k},
\end{equation}
and
\begin{equation}
\mathbf{h}^{\mathrm{w}}_{n,m}
=
\frac{\sum_{k=1}^{K}w^{e,k}_{n,m}\mathbf{H}_{n,m}^{k}}
{\sum_{k=1}^{K}w^{e,k}_{n,m}+\epsilon}.
\end{equation}
Here, the maximum is computed element-wise over the temporal dimension. Average pooling summarizes the overall residual context, max pooling retains the strongest local responses, and soft-weighted pooling focuses on features from anomaly-relevant intervals.

The three pooled representations are concatenated and projected into the final anomaly embedding:
\begin{equation}
\mathbf{z}_{n,m}
=
\mathbf{W}_{z}
\left[
\mathbf{h}^{\mathrm{avg}}_{n,m}
\Vert
\mathbf{h}^{\mathrm{max}}_{n,m}
\Vert
\mathbf{h}^{\mathrm{w}}_{n,m}
\right]
+
\mathbf{b}_{z},
\end{equation}
where $\Vert$ denotes feature concatenation. In this way, $\mathbf{z}_{n,m}$ jointly captures the global residual context, the strongest anomalous responses, and representative patterns within high-weight intervals.

To preserve anomaly-relevant information under dimensional compression, we pretrain $E_{\theta}$ jointly with a dedicated auxiliary decoder. Given $\mathbf{z}_{n,m}$, the decoder separately estimates anomaly occurrence and signed residual magnitude, whose element-wise product forms the reconstructed filtered residual sequence. The reconstruction objective assigns greater importance to anomaly-relevant intervals while penalizing nonzero reconstructions in near-zero regions. This encourages $\mathbf{z}_{n,m}$ to retain both the occurrence and magnitude of anomalous deviations without being dominated by normal intervals.

After pretraining, the trained encoder $E_{\theta}$ is incorporated into SR-Encoder for anomaly semantic extraction. The auxiliary decoder is excluded from the SR-Encoder forward path but retained separately to reconstruct the temporal anomaly hints used by AS-Diff.

\subsection{Backbone Denoiser Design in AS-Diff}\label{appendix:backbone_denoiser}

Following the Transformer-based design of Diffusion-TS~\cite{yuan2024diffusionts}, the Backbone Denoiser adopts an encoder-decoder architecture to predict the clean consumption sequence from its noisy intermediate state $\mathbf{s}_t$. At diffusion step $t$, it produces $\hat{\mathbf{s}}_{0}(\mathbf{s}_{t},t)$, an $x$-prediction estimate~\cite{li2026back} of the clean consumption sequence, which is subsequently used in Equation~\ref{eq:x_update}.

The Encoder consists of four Transformer blocks that capture temporal dependencies in $\mathbf{s}_t$. Each block operates on 64-dimensional hidden representations and contains a four-head self-attention layer followed by a feed-forward network with an expansion ratio of four. The output of the $l$-th Encoder block is denoted as $\mathbf{enc}_{t}^{l}$, which represents the layer-wise temporal features at diffusion step $t$. Adaptive layer normalization injects the diffusion timestep and region condition into the attention module, while residual connections are applied around both the attention and feed-forward layers. The final Encoder representation $\mathbf{enc}_{t}^{L}$ provides global temporal context for the subsequent Decoder.

Building on the final Encoder output $\mathbf{enc}_{t}^{L}$, the Decoder progressively refines the encoded temporal features through $L_D$ Decoder blocks. Specifically, the first Decoder block takes $\mathbf{enc}_{t}^{L}$ as its input, while each subsequent block operates on the output of the preceding block:
\begin{equation}
    \mathbf{h}_{t}^{(0)}=\mathbf{enc}_{t}^{L},
    \qquad
    \mathbf{h}_{t}^{(l)}
    =
    \operatorname{Dec}^{(l)}
    \left(
    \mathbf{h}_{t}^{(l-1)}
    \right),
    \quad l=1,\ldots,L_D.
\end{equation}
Each Decoder block then decomposes $\mathbf{h}_{t}^{(l)}$ into a trend component and a seasonality component to capture slowly varying structure and dominant periodic patterns~\cite{shen2026cited}, respectively. The resulting layer-wise components are aggregated across all Decoder blocks to obtain $\hat{\mathbf{s}}_{0}^{\mathrm{tr}}$ and $\hat{\mathbf{s}}_{0}^{\mathrm{se}}$, whose sum forms the clean-sequence estimate $\hat{\mathbf{s}}_{0}(\mathbf{s}_t,t)$.

\textbf{Trend Function.}
For the $l$-th Decoder block, the trend branch maps its hidden representation $\mathbf{h}_{t}^{(l)}$ to a layer-wise trend estimate that captures smooth, long-term variations. Direct Fourier-based modeling is less suitable for this purpose because non-periodic long-range trends can be incompletely represented by periodic bases. Inspired by~\cite{desai2021timevae,oreshkin2020nbeats}, we therefore employ a polynomial regressor:
\begin{equation}
\mathbf{s}_{t,l}^{\mathrm{tr}}
=
\operatorname{Trend}\!\left(
\mathbf{h}_{t}^{(l)};
\theta_{\mathrm{tr}}
\right)
=
\boldsymbol{\phi}(\tau)^{\top}
\left(
\mathbf{W}_{\mathrm{tr}}
h\!\left(\mathbf{h}_{t}^{(l)};\mathbf{a}_{\mathrm{tr}}\right)
+
\mathbf{b}_{\mathrm{tr}}
\right),
\end{equation}
where $\mathbf{s}_{t,l}^{\mathrm{tr}}$ denotes the trend component produced by the $l$-th Decoder block, $\tau$ is the temporal index, and
$\boldsymbol{\phi}(\tau)=[1,\tau,\tau^2,\ldots,\tau^p]^{\top}$
is a polynomial basis of degree $p$. The feature extractor
$h(\cdot;\mathbf{a}_{\mathrm{tr}})$ and projection parameters
$\mathbf{W}_{\mathrm{tr}}$ and $\mathbf{b}_{\mathrm{tr}}$
produce the corresponding polynomial coefficients. Since the trend branch is intended to represent slowly varying structure, we use a low polynomial degree, with $p=3$ in our implementation.

\textbf{Seasonality Function.}
In parallel, the seasonality branch maps $\mathbf{h}_{t}^{(l)}$ to a layer-wise seasonality estimate that captures periodic and oscillatory patterns in the frequency domain. Specifically, we apply the discrete Fourier transform, retain the Top-$K$ frequency components with the largest magnitudes, and reconstruct the seasonal component through the inverse transform:
\begin{equation}
\mathbf{s}_{t,l}^{\mathrm{se}}
=
\operatorname{Seasonality}\!\left(
\mathbf{h}_{t}^{(l)};
\theta_{\mathrm{se}}
\right)
=
\mathcal{F}^{-1}
\left(
\mathcal{M}_{K}
\left(
\mathcal{F}\!\left(\mathbf{h}_{t}^{(l)}\right)
\right)
\right),
\end{equation}
where $\mathbf{s}_{t,l}^{\mathrm{se}}$ denotes the seasonality component produced by the $l$-th Decoder block. The operators $\mathcal{F}$ and $\mathcal{F}^{-1}$ denote the discrete Fourier transform and its inverse, respectively. The operator $\mathcal{M}_{K}(\cdot)$ preserves the dominant frequency components together with their conjugate-symmetric counterparts, thereby retaining the main periodic structure while suppressing weak frequency noise.

\textbf{Output Aggregation.}
The layer-wise trend and seasonality components are aggregated across the $L_D$ Decoder blocks:
\begin{equation}
    \hat{\mathbf{s}}_{0}^{\mathrm{tr}}
    =
    \sum_{l=1}^{L_D}
    \mathbf{s}_{t,l}^{\mathrm{tr}},
    \qquad
    \hat{\mathbf{s}}_{0}^{\mathrm{se}}
    =
    \sum_{l=1}^{L_D}
    \mathbf{s}_{t,l}^{\mathrm{se}}.
\end{equation}
The final clean-sequence estimate is then obtained as
\begin{equation}
    \hat{\mathbf{s}}_{0}(\mathbf{s}_t,t)
    =
    \hat{\mathbf{s}}_{0}^{\mathrm{tr}}
    +
    \hat{\mathbf{s}}_{0}^{\mathrm{se}}.
\end{equation}
Thus, $\mathbf{enc}_{t}^{L}$ initializes the Decoder representations $\mathbf{h}_{t}^{(l)}$, each $\mathbf{h}_{t}^{(l)}$ produces a pair of layer-wise trend and seasonality components, and their aggregation forms the final estimate of the clean consumption sequence.

\section{Additional Experimental Details}
\subsection{Dataset Description}\label{appendix:dataset}
\subsubsection{FL1 \& FL2 Datasets} \label{appendix:dataset_florida}

We have access to city-scale household-level energy consumption data collected by a municipal utility provider in Tallahassee, Florida. The complete dataset spans over ten years and contains more than one billion energy consumption readings at a 30-minute granularity from over 50K independently metered households across 147 census block groups (CBGs). Each record contains a meter ID, timestamp, and energy consumption measured in kilowatt-hours (kWh). We construct two datasets centered on representative anomalous events. \textbf{FL1} covers October 1--30, 2018, when a major Category 5 Hurricane Michael made landfall on October 10 and caused widespread power disruptions and substantial low-consumption anomalies. \textbf{FL2} covers May 1--30, 2019, when a major heatwave led to elevated electricity demand and pronounced high-consumption anomalies. For each dataset, we select 10,000 households with the most complete records to ensure data completeness and temporal continuity. Their consumption records are organized into 30-day monthly sequences, and every eight consecutive 30-minute readings are aggregated into a non-overlapping 4-hour interval. Each household sequence therefore contains six observations per day and 180 observations over one month, resulting in a dataset shape of $(10{,}000 \times 180 \times 1)$ for both \textbf{FL1} and \textbf{FL2}.

We further incorporate demographic and socioeconomic attributes from the U.S. Census Bureau's American Community Survey (ACS)~\cite{uscensus_acs}. All region-level analyses and computations in this paper are conducted at the CBG level, except for Figure~\ref{fig:high_load_anomalies}, which is presented at the census block level. The study area comprises 147 CBGs, with attributes covering age, race, household income, educational attainment, housing value, and other demographic and socioeconomic characteristics. 

% The study area contains 56 census tracts, 147 census block groups, and 3,941 census blocks

\subsubsection{NY Dataset}
The New York dataset is obtained from the New York Independent System Operator (NYISO)~\cite{nyiso_load_data}, which operates and monitors New York State’s bulk electric power system. It contains publicly available integrated energy consumption measurements collected through supervisory control and data acquisition (SCADA) systems across the transmission network. Unlike the household-level energy consumption data in the Florida datasets, the New York dataset provides region-level energy consumption, with each region corresponding to an NYISO load zone. Each record includes a timestamp, a region identifier, and the corresponding energy consumption measured in megawatts (MW). The dataset covers the full calendar year of 2024 and includes 11 regions: CAPITL, CENTRL, DUNWOD, GENESE, HUD VL, LONGIL, MHK VL, MILLWD, N.Y.C., NORTH, and WEST. We organize the measurements into region-level daily sequences, each consisting of 24 hourly observations. Incomplete daily sequences are removed to ensure temporal completeness and consistency. After preprocessing, each region contains 365 complete daily sequences, yielding 4,015 samples in total. The final dataset has a shape of $(4{,}015 \times 24 \times 1)$, where each sample represents the daily energy consumption profile of one region, and the region identifier is used as the condition label.

\subsubsection{CA Dataset}
The California dataset is obtained from the California Independent System Operator (CAISO)~\cite{caiso_load_data}, which operates and monitors California’s bulk electric power system. It contains publicly available historical hourly energy consumption measurements collected through CAISO’s energy management system (EMS). The dataset provides region-level energy consumption for four regions corresponding to the CAISO load zones: Pacific Gas and Electric (PGE), Southern California Edison (SCE), San Diego Gas and Electric (SDGE), and Valley Electric Association (VEA). The system-wide CAISO aggregate is excluded from our analysis. The dataset covers the full calendar year of 2024, and we organize the measurements into region-level daily sequences, each consisting of 24 hourly observations. After preprocessing, each region contains 365 complete daily sequences, yielding 1,460 samples in total. The final dataset has a shape of $(1{,}460 \times 24 \times 1)$, where each sample represents the daily energy consumption profile of one region, and the region identifier is used as the condition label.

\subsection{Baseline Description} \label{Appendix:Baseline}
We compare \m with 11 baseline methods from seven different categories. An overview of these baselines is provided in Table~\ref{tab:baseline_statistics}. The implementation details are summarized as follows:

\begin{table*}[htbp]
\centering
\caption{Overview of baseline methods compared in this paper.}
\label{tab:baseline_statistics}
\renewcommand{\arraystretch}{1.12}
\small
\begin{tabular}{llll}
\toprule
Baseline & Category & Venue & Link \\
\midrule
TimeGAN~\cite{yoon2019time} & GAN & NeurIPS'19 &
\href{https://github.com/AlexanderVNikitin/tsgm/blob/main/tsgm/models/timeGAN.py}{Official Implementation} \\
\arrayrulecolor{gray!60}\cmidrule(lr){1-4}\arrayrulecolor{black}
TimeVAE~\cite{desai2021timevae} & VAE & ArXiv'21 &
\href{https://github.com/wangyz1999/timeVAE-pytorch}{Official Implementation} \\
koVAE~\cite{naimangenerative} & VAE & ICLR'24 &
\href{https://github.com/azencot-group/KoVAE}{Official Implementation} \\
\arrayrulecolor{gray!60}\cmidrule(lr){1-4}\arrayrulecolor{black}
F-Flow~\cite{alaa2021generative} & Flow & ICLR'21 &
\href{https://github.com/ahmedmalaa/Fourier-flows}{Official Implementation} \\
\arrayrulecolor{gray!60}\cmidrule(lr){1-4}\arrayrulecolor{black}
DiffWave~\cite{kongdiffwave} & Diffusion & ICLR'21 &
\href{https://github.com/lmnt-com/diffwave}{Official Implementation} \\
Diffusion-TS~\cite{yuan2024diffusionts} & Diffusion & ICLR'24 &
\href{https://github.com/Y-debug-sys/Diffusion-TS}{Official Implementation} \\
\arrayrulecolor{gray!60}\cmidrule(lr){1-4}\arrayrulecolor{black}
SDForger~\cite{rousseau2025forging} & LLM & NeurIPS'25 &
\href{https://github.com/IBM/fms-dgt/tree/main/fms_dgt/public/databuilders/time_series}{Official Implementation} \\
\arrayrulecolor{gray!60}\cmidrule(lr){1-4}\arrayrulecolor{black}
FIDE~\cite{galib2024fide} & Extreme-Aware & NeurIPS'24 &
\href{https://github.com/galib19/FIDE}{Official Implementation} \\
HeavyDiff~\cite{pandey2025heavy} & Extreme-Aware & ICLR'25 &
\href{https://github.com/Y-debug-sys/Diffusion-TS}{Official Implementation} \\
\arrayrulecolor{gray!60}\cmidrule(lr){1-4}\arrayrulecolor{black}
CENTS~\cite{fuest2025cents} & Electricity-Specific & ArXiv'25 & \href{https://github.com/DAI-Lab/Cents}{Official Implementation} \\
Cond-Diff~\cite{fu2024creating} & Electricity-Specific & Energy and Buildings'24 &
\href{https://github.com/buds-lab/energy-diffusion}{Official Implementation} \\
\bottomrule
\end{tabular}
\renewcommand{\arraystretch}{1.0}
\end{table*}

\begin{itemize}
    \item \textbf{TimeGAN}~\cite{yoon2019time}: We implement TimeGAN using the TSGM library~\cite{nikitin2024tsgm} for improved reproducibility and compatibility, as the original codebase relies on an outdated TensorFlow environment. The sequence length and feature dimension are inferred automatically from the input data. We use a GRU-based architecture with a hidden dimension of 24, three recurrent layers, a batch size of 128, and $\gamma=1.0$. Separate Adam optimizers with a learning rate of $10^{-3}$ are used for the model components, following the original reconstruction, supervised, and adversarial objectives. The model is trained for 100 epochs using the standard embedding, supervised, and joint training stages. During generation, uniformly sampled noise is passed through the trained generator in batches, and the synthetic sequences are transformed back to the original scale through inverse min-max normalization.

    \item \textbf{TimeVAE}~\cite{desai2021timevae}: We implement TimeVAE using a recent PyTorch reimplementation of the original model. TimeVAE adopts a convolutional variational autoencoder, where the encoder maps each input sequence to a latent representation and the decoder reconstructs it through global-level, polynomial-trend, and residual components. The model is optimized using a weighted combination of reconstruction loss and KL divergence. We use a latent dimension of 8, hidden dimensions of ([50,100,200]), a reconstruction weight of 3.0, a batch size of 16, and a second-order polynomial trend component with residual connections enabled and seasonality disabled. Each dataset is split into 90\% training and 10\% validation sets, with min-max normalization fitted only on the training data. The model is trained for 200 epochs using Adam with a random seed. During generation, latent variables are sampled from a standard Gaussian distribution, decoded into synthetic sequences, and transformed back to the original scale, with the number of generated samples matching the training-set size.

    \item \textbf{koVAE}~\cite{naimangenerative}: We reproduce koVAE using the authors' official implementation. koVAE extends the variational autoencoder by replacing the conventional static latent prior with a Koopman-inspired dynamical prior, where latent evolution is modeled through a linear transition operator to capture temporal regularities. We adopt the regular setting with bidirectional GRU encoders and decoders, using a three-layer encoder with hidden dimension 20, a latent dimension of 16, batch normalization, and a sigmoid output layer. All channels are independently normalized to \([0,1]\) using min--max scaling and transformed back to the original scale after generation. The training objective combines reconstruction, KL regularization, and one-step Koopman prior prediction losses with weights \(1.0\), \(0.007\), and \(0.005\), respectively. The model is trained for 100 epochs using Adam with a learning rate of \(7\times10^{-4}\), a batch size of 64, and random seed 10. During generation, latent trajectories are sampled and decoded into synthetic sequences, with the number of generated samples matching the training-set size.

    \item \textbf{F-Flow}~\cite{alaa2021generative}: We reproduce F-Flow using the authors' official implementation. F-Flow is a normalizing-flow-based generative model that transforms time series into the Fourier domain and learns an explicit likelihood over spectral coefficients through invertible flows, enabling it to capture global temporal and periodic patterns. All channels are independently normalized to \([0,1]\), after which each sequence is flattened and transformed into the frequency domain; when the flattened length is even, a zero is prepended to satisfy the Fourier transform requirement. We use five flow layers with a hidden dimension of 200 and standardize the spectral coefficients using dataset-level statistics. The model is trained for 1,000 epochs using Adam with a learning rate of \(10^{-3}\) and exponential decay. During generation, Gaussian samples are mapped through the inverse flow and inverse Fourier transform, reshaped to the original sequence format, and transformed back to the original scale, with the number of generated samples matching the training-set size.

    \item \textbf{DiffWave}~\cite{kongdiffwave}: We adapt the authors' official implementation from speech waveform synthesis to conditional univariate time-series generation by replacing the audio pipeline with direct \texttt{.npy} loading for sequences of shape ((N,T,1)). DiffWave progressively denoises Gaussian noise using a one-dimensional convolutional network, with the region index provided as the conditioning input in place of the original spectrogram~\cite{cheng2025misleader}. The data are standardized using dataset-level mean and standard deviation statistics, and shorter sequences are right-padded when necessary. We retain the original architecture with 30 residual layers, 64 residual channels, a dilation cycle length of 10, and 50 diffusion steps. The model is trained using Adam with a learning rate of \(2\times10^{-4}\) and an \(L_1\) noise-prediction loss. During generation, Gaussian noise is refined through conditional reverse diffusion and transformed back to the original scale, with the number of generated samples matching the training-set size.
    
    \item \textbf{Diffusion-TS}~\cite{yuan2024diffusionts}: We use the official implementation of Diffusion-TS and retain its original diffusion training and sampling procedure. Diffusion-TS reconstructs the clean sequence from noisy inputs through an interpretable Transformer-based denoiser that explicitly decomposes predictions into trend and seasonality components. The input sequences are normalized to \([-1,1]\), and the model is configured with two encoder layers, two decoder layers, a hidden dimension of 64, and four attention heads. The diffusion process uses 1,000 steps with a cosine noise schedule and an \(L_1\) reconstruction loss. We train the model for 10,000 optimization steps using Adam with a base learning rate of \(10^{-5}\), a batch size of 64, and gradient accumulation every two steps. EMA is applied with a decay of 0.995 and an update interval of 10, together with a warmup-based learning-rate scheduler. During generation, the EMA model performs the reverse diffusion process to synthesize sequences with the same size as the training set.

    \item \textbf{SDForger}~\cite{rousseau2025forging}: We follow the original SDForger framework, which transforms raw time series into compact functional embeddings, formulates synthesis as structured text generation with a large language model~\cite{li2026llmclinicalgraphstructure}, and decodes the generated embeddings back into time-series samples. We use the multisample setting with FastICA, where the embedding dimension is selected automatically using a variance retention target of 0.7. The resulting embeddings are used to fine-tune an autoregressive language model, and generated embeddings are recovered through inverse embedding and inverse normalization. We set diversity threshold to 0 and generate the same number of synthetic samples as in the training set.

    \item \textbf{FIDE}~\cite{galib2024fide}: We adapt the original FIDE implementation to our benchmark. FIDE is an extreme-aware conditional diffusion model that improves tail preservation through frequency-domain inflation and block-maximum conditioning. Each training sequence is paired with its block maximum, and a generalized extreme value (GEV) distribution is fitted to the observed maxima. The denoiser is trained with the standard DDPM objective and an additional GEV-based regularization term, while block maxima sampled from the fitted GEV distribution guide reverse diffusion during generation. We use a hidden dimension of 64, a batch size of 2,000, and train the model for 400 epochs. The diffusion process contains 100 steps with a linear noise schedule from \(\beta_{\mathrm{start}}=10^{-4}\) to \(\beta_{\mathrm{end}}=0.2\), together with correlated Gaussian-process noise using \(\sigma=0.05\). Following the original frequency-enhancement strategy, the top 20\% high-frequency components are amplified by a factor of 1.1 before training and inversely transformed after generation.
    
    \item \textbf{HeavyDiff}~\cite{pandey2025heavy}: Since no official implementation of HeavyDiff is publicly available and the original method is not designed specifically for time-series generation, we adapt its core heavy-tailed diffusion mechanism within Diffusion-TS. Specifically, Gaussian noise in both the forward diffusion and reverse sampling processes is replaced with Student-(t) noise generated through the Gaussian scale-mixture formulation \(\epsilon=z/\sqrt{\kappa/\nu}\), where \(z\sim\mathcal{N}(0,I)\), \(\kappa\sim\chi_{\nu}^{2}\), and \(\nu\) controls the tail heaviness. To reduce the variance shift and improve training stability, we apply the correction \(\epsilon\leftarrow\epsilon\sqrt{(\nu-2)/\nu}\) for \(\nu>2\). In the main experiments, we set \(\nu=2.5\) and enable variance correction by default, while retaining all other Diffusion-TS configurations.

    \item \textbf{CENTS}~\cite{fuest2025cents}: We adapt the core context-aware generation mechanism of CENTS to the Diffusion-TS backbone while retaining the original diffusion training and sampling procedure. Specifically, the context condition reuses the regional information available in our benchmark, including the region identifier, median household income, and median property value. The region identifier is represented through a learnable embedding, while the two numerical attributes are normalized and projected into the same latent space. These representations are concatenated and compressed by a multilayer perceptron into a unified context embedding with a dimension of 64, which replaces the original region condition and is injected into the Diffusion-TS denoiser. Following CENTS, auxiliary reconstruction heads recover the region identifier and regional attributes from the compressed context embedding, encouraging it to retain information from all contextual variables. The model is jointly optimized using the original Diffusion-TS reconstruction objective and the context reconstruction loss, with the latter weighted by 0.1. For efficient method-level reproduction, we reuse the dataset-level normalization adopted by Diffusion-TS and do not implement the context-dependent normalizer designed primarily for unseen contextual combinations. All remaining model architecture, diffusion, optimization, and generation settings are kept identical to those of Diffusion-TS.

    \item \textbf{Cond-Diff}~\cite{fu2024creating}: We adapt the original conditional diffusion framework for synthetic energy meter data to our benchmark. Cond-Diff reshapes each one-dimensional consumption sequence into a two-dimensional temporal matrix and applies a metadata-conditioned U-Net to capture both intra-day and inter-day patterns. For our four-hour-resolution monthly sequences, each sample of length 180 is reshaped into a ($30\times6$) matrix, where the two dimensions represent days and intervals within each day, respectively. The context condition reuses the regional information in our benchmark, including the region identifier, median household income, and median property value. The region identifier is represented by a learnable embedding, while the two numerical attributes are normalized and projected into latent representations. These features are concatenated into a unified context vector and injected together with the diffusion-step embedding into the residual blocks of the U-Net. The model is trained using the standard DDPM noise-prediction objective with Gaussian noise, without additional context reconstruction or anomaly-specific losses. We use a base hidden dimension of 64, 500 diffusion steps, a batch size of 64, and train the model for 10,000 optimization steps using Adam with a learning rate of ($10^{-4}$). The input sequences are normalized to ([-1,1]), and the generated matrices are reshaped back into one-dimensional sequences after reverse diffusion. The number of generated samples is set equal to the size of the training set.        
\end{itemize}

\subsection{Metric Description} \label{Appendix:Metric}
To comprehensively evaluate generation performance, we employ 10 metrics across three complementary dimensions: overall generation fidelity, anomaly preservation fidelity, and downstream quality. The detailed definitions and implementation procedures of these metrics are provided below.

For \textbf{Overall Generation Fidelity}, we assess the overall similarity between real and generated data distributions using four distance-based metrics following the evaluation setting of TSGBench~\cite{ang2023tsgbench}, as defined below:

\begin{itemize}
    \item \textbf{Temporal Wasserstein Distance (T-Wass.)}: T-Wass. measures the discrepancy between real and generated consumption distributions at each time interval. For the \(t\)-th interval, we compute the 1-Wasserstein distance between the real values \(\{x_i^t\}_{i=1}^{N}\) and the generated values \(\{\bar{x}_i^t\}_{i=1}^{N}\):
    \begin{equation}
    W_1^{(t)}
    =
    W_1\!\left(
    \{x_i^t\}_{i=1}^{N},
    \{\bar{x}_i^t\}_{i=1}^{N}
    \right).
    \end{equation}
    The final score is averaged over all \(K\) time intervals:
    \begin{equation}
    \mathrm{T\text{-}Wass.}
    =
    \frac{1}{K}
    \sum_{t=1}^{K}
    W_1^{(t)}.
    \end{equation}
    A lower value indicates greater similarity between the temporal distributions of real and generated consumption data.

    \item \textbf{Differential Wasserstein Distance (D-Wass.)}: D-Wass. measures the discrepancy between the real and generated distributions of interval-to-interval consumption changes, capturing whether their temporal dynamics are consistent. For the \(t\)-th transition, we first compute the consumption increments:
    \begin{equation}
    \Delta x_i^t = x_i^{t+1}-x_i^t,
    \qquad
    \Delta \bar{x}_i^t = \bar{x}_i^{t+1}-\bar{x}_i^t,
    \end{equation}
    where \(t=1,\ldots,K-1\). We then compute the 1-Wasserstein distance between the real and generated increment distributions:
    \begin{equation}
    D_1^{(t)}
    =
    W_1\!\left(
    \{\Delta x_i^t\}_{i=1}^{N},
    \{\Delta \bar{x}_i^t\}_{i=1}^{N}
    \right).
    \end{equation}
    The final score is averaged over all \(K-1\) consecutive time transitions:
    \begin{equation}
    \mathrm{D\text{-}Wass.}
    =
    \frac{1}{K-1}
    \sum_{t=1}^{K-1}
    D_1^{(t)}.
    \end{equation}
    A lower value indicates greater similarity between the temporal distributions of real and generated consumption data.

    \item \textbf{Slot-wise Wasserstein Distance (S-Wass.)}: S-Wass. measures the discrepancy between real and generated consumption distributions at the same within-day time slot across different days. Let \(S\) denote the number of time slots per day. For the \(s\)-th slot, we collect all time intervals assigned to that slot:
    \begin{equation}
    \mathcal{I}_s
    =
    \left\{
    t \mid t \bmod S = s
    \right\},
    \qquad
    s=0,\ldots,S-1.
    \end{equation}
    We then compute the 1-Wasserstein distance between the pooled real and generated values:
    \begin{equation}
    S_1^{(s)}
    =
    W_1\!\left(
    \{x_i^t\}_{i=1,\ldots,N;\,t\in\mathcal{I}_s},
    \{\bar{x}_i^t\}_{i=1,\ldots,N;\,t\in\mathcal{I}_s}
    \right).
    \end{equation}
    The final score is averaged over all \(S\) within-day time slots:
    \begin{equation}
    \mathrm{S\text{-}Wass.}
    =
    \frac{1}{S}
    \sum_{s=0}^{S-1}
    S_1^{(s)}.
    \end{equation}
    A lower value indicates greater similarity between the recurring within-day distributions of real and generated consumption data.

    \item \textbf{Maximum Mean Discrepancy (MMD)}: MMD measures the discrepancy between real and generated consumption distributions at the trajectory level, capturing differences in the overall sequence structure. Each sequence is treated as a \(K\)-dimensional vector, and similarity between two sequences is computed using the radial basis function (RBF) kernel:
    \begin{equation}
    k(\mathbf{x},\mathbf{y})
    =
    \exp\!\left(
    -\frac{\|\mathbf{x}-\mathbf{y}\|_2^2}{2\sigma^2}
    \right),
    \end{equation}
    where \(\sigma\) denotes the kernel bandwidth. Given \(N\) real sequences \(\{\mathbf{x}_i\}_{i=1}^{N}\) and \(N\) generated sequences \(\{\bar{\mathbf{x}}_i\}_{i=1}^{N}\), the squared MMD is computed as:
    \begin{equation}
    \begin{aligned}
    \mathrm{MMD}^2
    ={}&
    \frac{1}{N^2}
    \sum_{i=1}^{N}\sum_{j=1}^{N}
    k(\mathbf{x}_i,\mathbf{x}_j)
    +
    \frac{1}{N^2}
    \sum_{i=1}^{N}\sum_{j=1}^{N}
    k(\bar{\mathbf{x}}_i,\bar{\mathbf{x}}_j)
    \\
    &-
    \frac{2}{N^2}
    \sum_{i=1}^{N}\sum_{j=1}^{N}
    k(\mathbf{x}_i,\bar{\mathbf{x}}_j).
    \end{aligned}
    \end{equation}
    The final score is obtained as:
    \begin{equation}
    \mathrm{MMD}
    =
    \sqrt{\max\!\left(\mathrm{MMD}^2,0\right)}.
    \end{equation}
    A lower value indicates greater similarity between the trajectory-level distributions of real and generated consumption data.
\end{itemize}

For \textbf{Anomaly Preservation Fidelity}, we evaluate how well the generated data preserve anomalous events in the real data using four metrics that capture distinct characteristics: anomaly event rate, anomaly event count, anomaly event energy, and aggregated anomaly-point similarity. 

Before defining these metrics, we identify anomalous points using the residual-based formulation introduced in Section~\ref{sec:Anomalous_event_defintion}. For the generated data, we first transform each generated normalized sequence back to its original consumption scale. We then compute the residuals of both real and generated sequences relative to their corresponding regional background patterns. Specifically, the residuals are defined as
\begin{equation}
e_{n,m}^{k}
=
x_{n,m}^{k}-b_{n}^{k},
\qquad
\bar{e}_{n,m}^{k}
=
\bar{x}_{n,m}^{k}-\bar{b}_{n}^{k},
\end{equation}
where $\bar{x}_{n,m}^{k}$ denotes the inverse-normalized generated consumption and $b_n^k$ and $\bar{b}_n^k$ denote the regional background values of the real and generated data, respectively. An interval is identified as anomalous when the absolute residual exceeds the anomaly threshold $\delta$:
\begin{equation}
a_{n,m}^{k}
=
\mathbb{I}\!\left(
|e_{n,m}^{k}|>\delta
\right),
\qquad
\bar{a}_{n,m}^{k}
=
\mathbb{I}\!\left(
|\bar{e}_{n,m}^{k}|>\delta
\right).
\end{equation}
Here, $\mathbb{I}(\cdot)$ denotes the indicator function, which equals 1 when the specified condition is satisfied and 0 otherwise. Based on these residuals and anomaly indicators, we define the following four metrics to compare anomalous characteristics between real and generated consumption data:
\begin{itemize}
    \item \textbf{Anomaly Rate Distance (A-Rate)}: A-Rate measures the discrepancy between real and generated data in the proportion of anomalous intervals within each consumption sequence. For each real sequence $\mathbf{x}_{n,m}$, its anomaly rate is computed as
    \begin{equation}
    q_{n,m}
    =
    \frac{1}{K}
    \sum_{k=1}^{K}
    a_{n,m}^{k},
    \end{equation}
    where $a_{n,m}^{k}$ indicates whether the $k$-th interval is anomalous. Similarly, the anomaly rate of each generated sequence is
    \begin{equation}
    \bar{q}_{n,m}
    =
    \frac{1}{K}
    \sum_{k=1}^{K}
    \bar{a}_{n,m}^{k}.
    \end{equation}
    We then compute the 1-Wasserstein distance between the anomaly-rate distributions of the real and generated sequences:
    \begin{equation}
    \mathrm{A\text{-}Rate}
    =
    W_1\!\left(
    \{q_{n,m}\},
    \{\bar{q}_{n,m}\}
    \right).
    \end{equation}
    A lower value indicates greater similarity in the proportion of time intervals identified as anomalous between real and generated data.

    \item \textbf{Anomaly Event Count Distance (A-Count)}: A-Count measures the discrepancy between real and generated data in the number of anomalous events within each consumption sequence. An anomalous event is defined as a consecutive segment of anomalous intervals. For each real sequence, the event count is computed as
    \begin{equation}
    c_{n,m}
    =
    \sum_{k=1}^{K}
    a_{n,m}^{k}
    \left(
    1-a_{n,m}^{k-1}
    \right),
    \end{equation}
    where we set $a_{n,m}^{0}=0$, such that each transition from a normal interval to an anomalous interval marks the start of a new event. Similarly, the event count for each generated sequence is
    \begin{equation}
    \bar{c}_{n,m}
    =
    \sum_{k=1}^{K}
    \bar{a}_{n,m}^{k}
    \left(
    1-\bar{a}_{n,m}^{k-1}
    \right),
    \end{equation}
    with $\bar{a}_{n,m}^{0}=0$. We then compute the 1-Wasserstein distance between the event-count distributions of the real and generated sequences:
    \begin{equation}
    \mathrm{A\text{-}Count}
    =
    W_1\!\left(
    \{c_{n,m}\},
    \{\bar{c}_{n,m}\}
    \right).
    \end{equation}
    A lower value indicates greater similarity in anomalous event frequency between real and generated data.

    \item \textbf{Anomaly Energy Distance (A-Energy)}: A-Energy measures the discrepancy between real and generated data in the overall magnitude of anomalous deviations within each consumption sequence. For each real sequence, its anomaly energy is computed as
    \begin{equation}
    g_{n,m}
    =
    \sum_{k=1}^{K}
    \left(e_{n,m}^{k}\right)^2
    a_{n,m}^{k},
    \end{equation}
    where only residuals at anomalous intervals contribute to the energy. Similarly, the anomaly energy of each generated sequence is
    \begin{equation}
    \bar{g}_{n,m}
    =
    \sum_{k=1}^{K}
    \left(\bar{e}_{n,m}^{k}\right)^2
    \bar{a}_{n,m}^{k}.
    \end{equation}
    We then compute the 1-Wasserstein distance between the anomaly-energy distributions of the real and generated sequences:
    \begin{equation}
    \mathrm{A\text{-}Energy}
    =
    W_1\!\left(
    \{g_{n,m}\},
    \{\bar{g}_{n,m}\}
    \right).
    \end{equation}
    A lower value indicates greater similarity in anomalous deviation magnitude between real and generated data.

    \item \textbf{Anomaly Tail Distance (A-Tail)}: A-Tail measures the discrepancy between real and generated data in the magnitude distribution of anomalous residuals. Unlike the preceding sequence-level metrics, A-Tail pools the absolute residuals from all anomalous intervals:
    \begin{equation}
    \mathcal{A}
    =
    \left\{
    |e_{n,m}^{k}|
    \;\middle|\;
    a_{n,m}^{k}=1
    \right\},
    \end{equation}
    and similarly for the generated data:
    \begin{equation}
    \bar{\mathcal{A}}
    =
    \left\{
    |\bar{e}_{n,m}^{k}|
    \;\middle|\;
    \bar{a}_{n,m}^{k}=1
    \right\}.
    \end{equation}
    We then compute the 1-Wasserstein distance between the pooled anomalous-residual distributions:
    \begin{equation}
    \mathrm{A\text{-}Tail}
    =
    W_1\!\left(
    \mathcal{A},
    \bar{\mathcal{A}}
    \right).
    \end{equation}
    Unlike the preceding household-level metrics, A-Tail pools anomalous residuals across all households to evaluate the city-level distribution of anomaly generation. A lower value therefore indicates greater similarity in the overall distribution of anomalous residual magnitudes between real and generated data.
\end{itemize}

For \textbf{Downstream Quality}, we consider two representative downstream tasks: anomaly detection and time-series prediction. Following the train-on-synthetic, test-on-real strategy, we train downstream models exclusively on generated data and evaluate them on real data. This setting assesses whether the generated data preserve task-relevant patterns and can serve as an effective substitute for real training data. 

For both downstream tasks, we construct training samples using a sliding-window strategy. Given a consumption sequence, the input at interval $k$ is defined as
\begin{equation}
\mathbf{X}_{n,m}^{k}
=
\left[
x_{n,m}^{k-L},
\ldots,
x_{n,m}^{k-1}
\right]
\in
\mathbb{R}^{L}.
\end{equation}
where $L$ denotes the length of the lookback window. Models are trained exclusively on windows extracted from generated data and evaluated on windows from real data. We use Transformer-based models for both tasks and report the area under the precision--recall curve (PR-AUC), which is well suited to the imbalanced labels considered in our evaluation:
\begin{equation}
\mathrm{PR\text{-}AUC}
=
\sum_{q}
\left(
R_q-R_{q-1}
\right)P_q,
\end{equation}
where $P_q$ and $R_q$ denote precision and recall at the $q$-th decision threshold, respectively. A higher PR-AUC indicates better downstream performance on real data.

\begin{itemize}
    \item \textbf{Anomaly Detection PR-AUC (Det-PRAUC)}: Det-PRAUC evaluates whether a model trained on generated data can identify the starting points of anomalous events in real consumption sequences. The start of an anomalous event occurs when the current interval is anomalous while the preceding interval is normal. Accordingly, the detection label at interval $k$ is defined as
    \begin{equation}
    y_{n,m}^{k,\mathrm{det}}
    =
    \mathbb{I}\!\left(
    |e_{n,m}^{k}|>\delta
    \;\land\;
    |e_{n,m}^{k-1}|\leq\delta
    \right),
    \end{equation}
    where $\delta$ is the anomaly threshold defined in Section~\ref{sec:Anomalous_event_defintion}. For each interval $k$, the model takes the preceding $L_{\mathrm{det}}$ intervals as input, where $L_{\mathrm{det}}=24$ for FL1 and FL2 and $L_{\mathrm{det}}=6$ for NY and CA:
    \begin{equation}
    \mathbf{X}_{n,m}^{k,\mathrm{det}}
    =
    \left[
    x_{n,m}^{k-L_{\mathrm{det}}},
    \ldots,
    x_{n,m}^{k-1}
    \right].
    \end{equation}
    A Transformer-based binary classifier is trained on windows extracted from generated data and evaluated on real data, producing the probability $\hat{y}_{n,m}^{k,\mathrm{det}}$ that an anomalous event starts at interval $k$. The final metric is computed as
    \begin{equation}
    \mathrm{Det\text{-}PRAUC}
    =
    \mathrm{PR\text{-}AUC}
    \left(
    \left\{y_{n,m}^{k,\mathrm{det}}\right\},
    \left\{\hat{y}_{n,m}^{k,\mathrm{det}}\right\}
    \right).
    \end{equation}
    A higher value indicates that the generated data preserve patterns useful for detecting anomaly onsets in real data.

    \item \textbf{Anomaly Prediction PR-AUC (Pred-PRAUC)}: Pred-PRAUC evaluates whether a model trained on generated data can predict anomalous intervals in real consumption sequences one step ahead. The prediction label at interval $k$ is defined according to whether the next interval is anomalous:
    \begin{equation}
    y_{n,m}^{k,\mathrm{pred}}
    =
    \mathbb{I}\!\left(
    |e_{n,m}^{k+1}|>\delta
    \right),
    \end{equation}
    where $\delta$ is the anomaly threshold defined in Section~\ref{sec:Anomalous_event_defintion}. For each interval $k$, the model takes the preceding $L_{\mathrm{pred}}$ intervals as input, where $L_{\mathrm{pred}}=48$ for FL1 and FL2 and $L_{\mathrm{pred}}=6$ for NY and CA.
    \begin{equation}
    \mathbf{X}_{n,m}^{k,\mathrm{pred}}
    =
    \left[
    x_{n,m}^{k-L_{\mathrm{pred}}+1},
    \ldots,
    x_{n,m}^{k}
    \right].
    \end{equation}
    A Transformer-based binary classifier is trained on windows extracted from generated data and evaluated on real data, producing the probability $\hat{y}_{n,m}^{k,\mathrm{pred}}$ that the next interval is anomalous. The final metric is computed as
    \begin{equation}
    \mathrm{Pred\text{-}PRAUC}
    =
    \mathrm{PR\text{-}AUC}
    \left(
    \left\{y_{n,m}^{k,\mathrm{pred}}\right\},
    \left\{\hat{y}_{n,m}^{k,\mathrm{pred}}\right\}
    \right).
    \end{equation}
    A higher value indicates that the generated data preserve temporal patterns useful for predicting future anomalies in real data.
\end{itemize}

\subsection{More Results on Generation Fidelity and Downstream Utility}
\label{appendix:more_result}
As shown in Tables~\ref{tab:FL2}, \ref{tab:NY}, and \ref{tab:CA}, we report additional results on the FL2, NY, and CA datasets across overall generation fidelity, anomaly preservation fidelity, and downstream utility. The results are consistent with those observed on FL1. Specifically, \m ranks among the top two methods on all ten metrics for each dataset, achieving the best performance on seven metrics for FL2 and eight metrics for both NY and CA. It consistently performs best on all four anomaly preservation fidelity metrics while maintaining top-two performance in overall generation fidelity and downstream utility. Compared with the strongest baseline on each metric, \m improves anomaly preservation fidelity by an average of 24.00\% and downstream utility by 3.54\% across the three datasets. These results demonstrate the \textbf{robustness and generalizability} of \m across different anomalous events, geographic regions, and temporal granularities.

\subsection{More Results on Anomaly-Preserving Performance}
\label{appendix:anomaly_preserving_performance}
\begin{figure*}[htbp]
    \centering

    \begin{subfigure}[t]{0.235\textwidth}
        \centering
        \includegraphics[width=\linewidth]{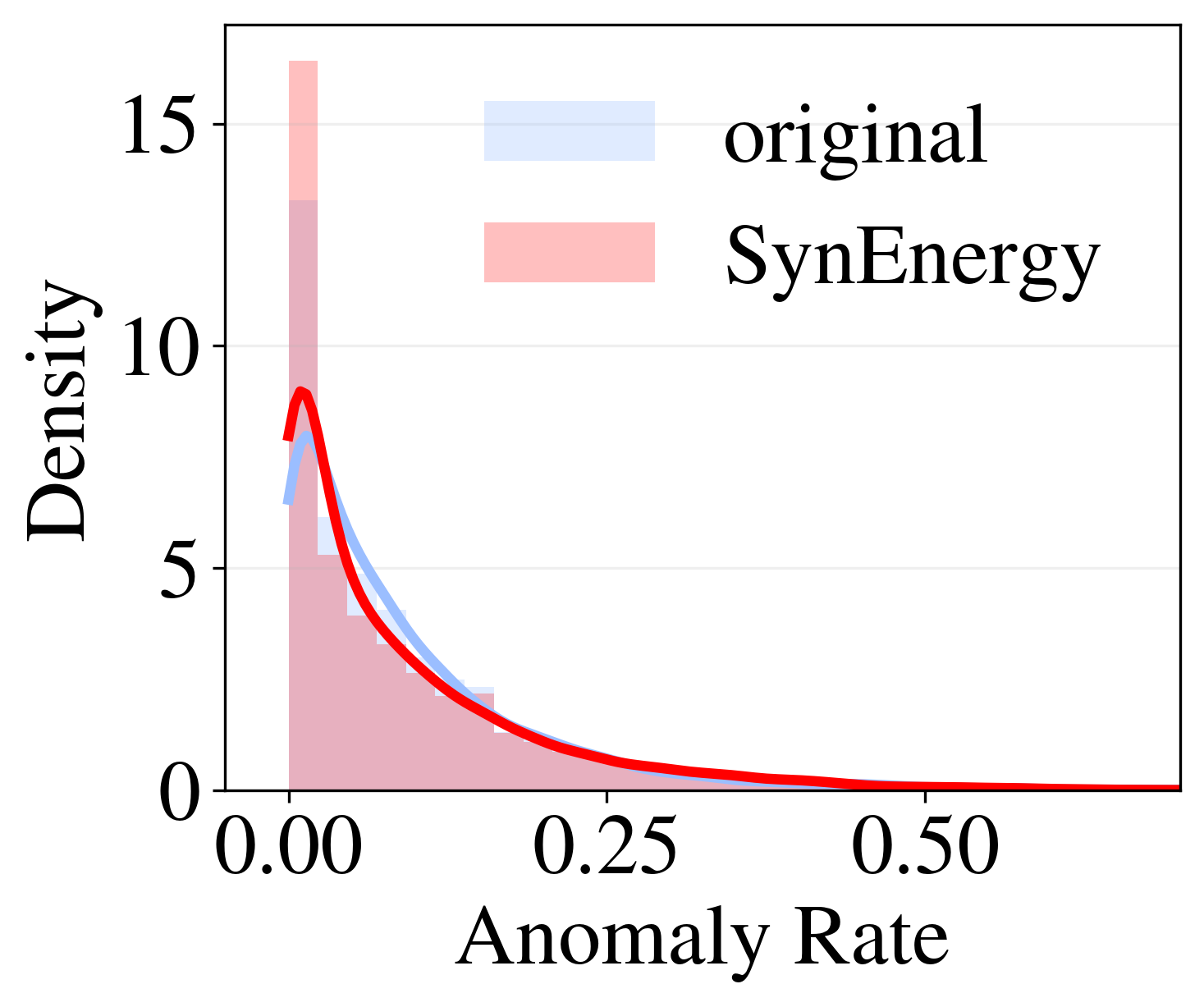}
        \caption{SynEnergy.}
        \label{fig:anomaly_rate_synenergy}
    \end{subfigure}
    \hfill
    \begin{subfigure}[t]{0.235\textwidth}
        \centering
        \includegraphics[width=\linewidth]{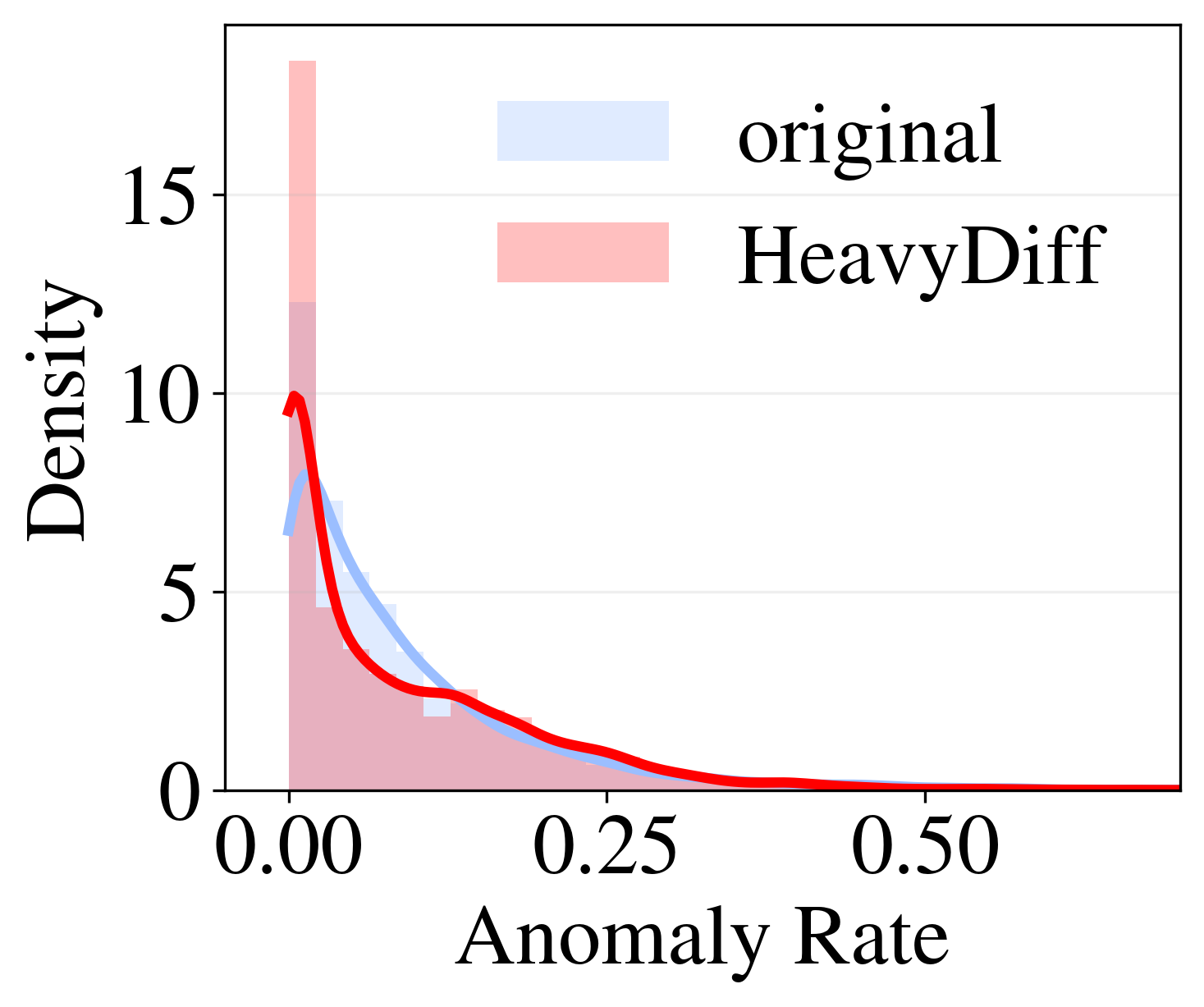}
        \caption{HeavyDiff.}
        \label{fig:anomaly_rate_HeavyDiff}
    \end{subfigure}
    \hfill
    \begin{subfigure}[t]{0.235\textwidth}
        \centering
        \includegraphics[width=\linewidth]{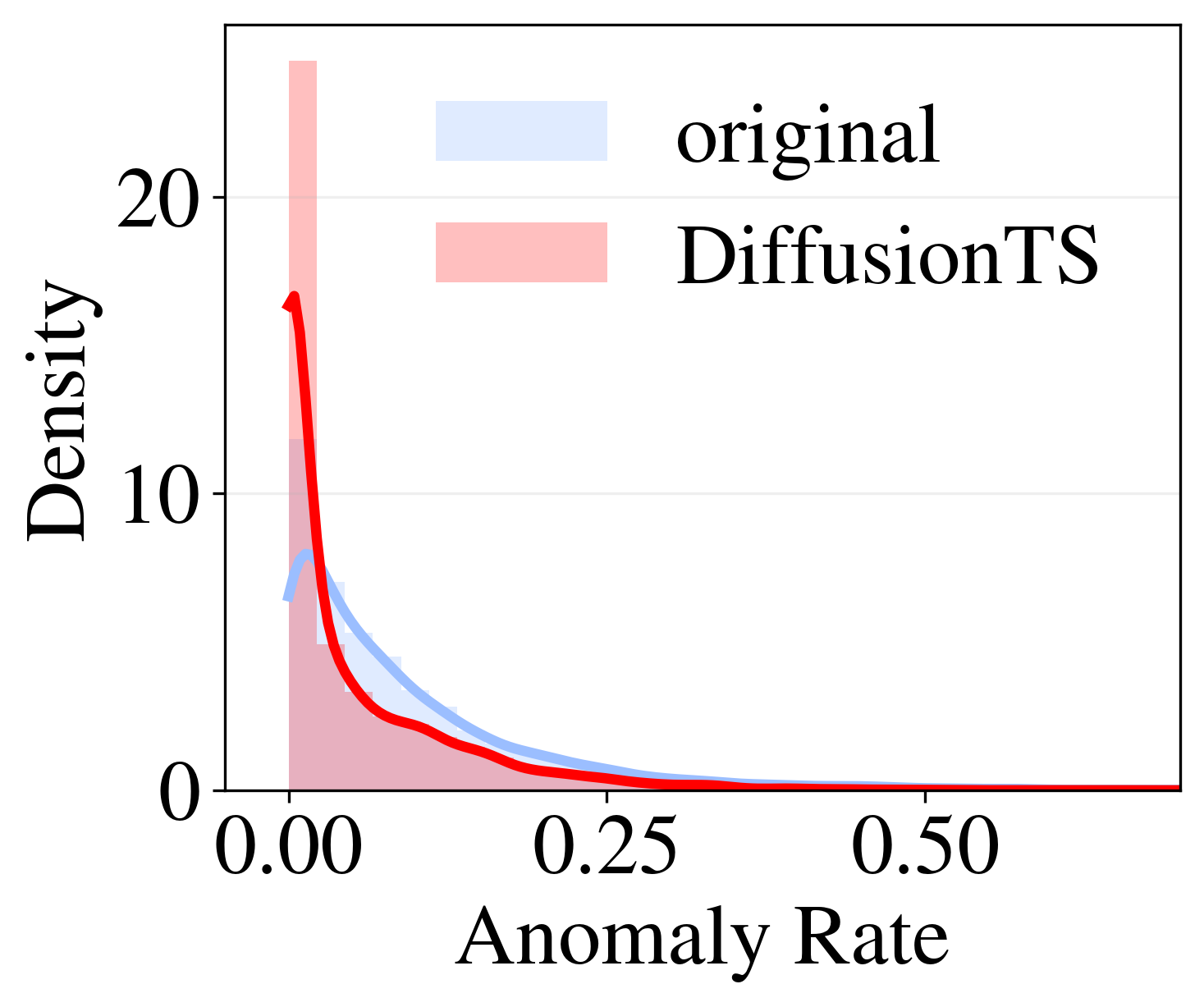}
        \caption{DiffusionTS.}
        \label{fig:anomaly_rate_DiffusionTS}
    \end{subfigure}
    \hfill
    \begin{subfigure}[t]{0.235\textwidth}
        \centering
        \includegraphics[width=\linewidth]{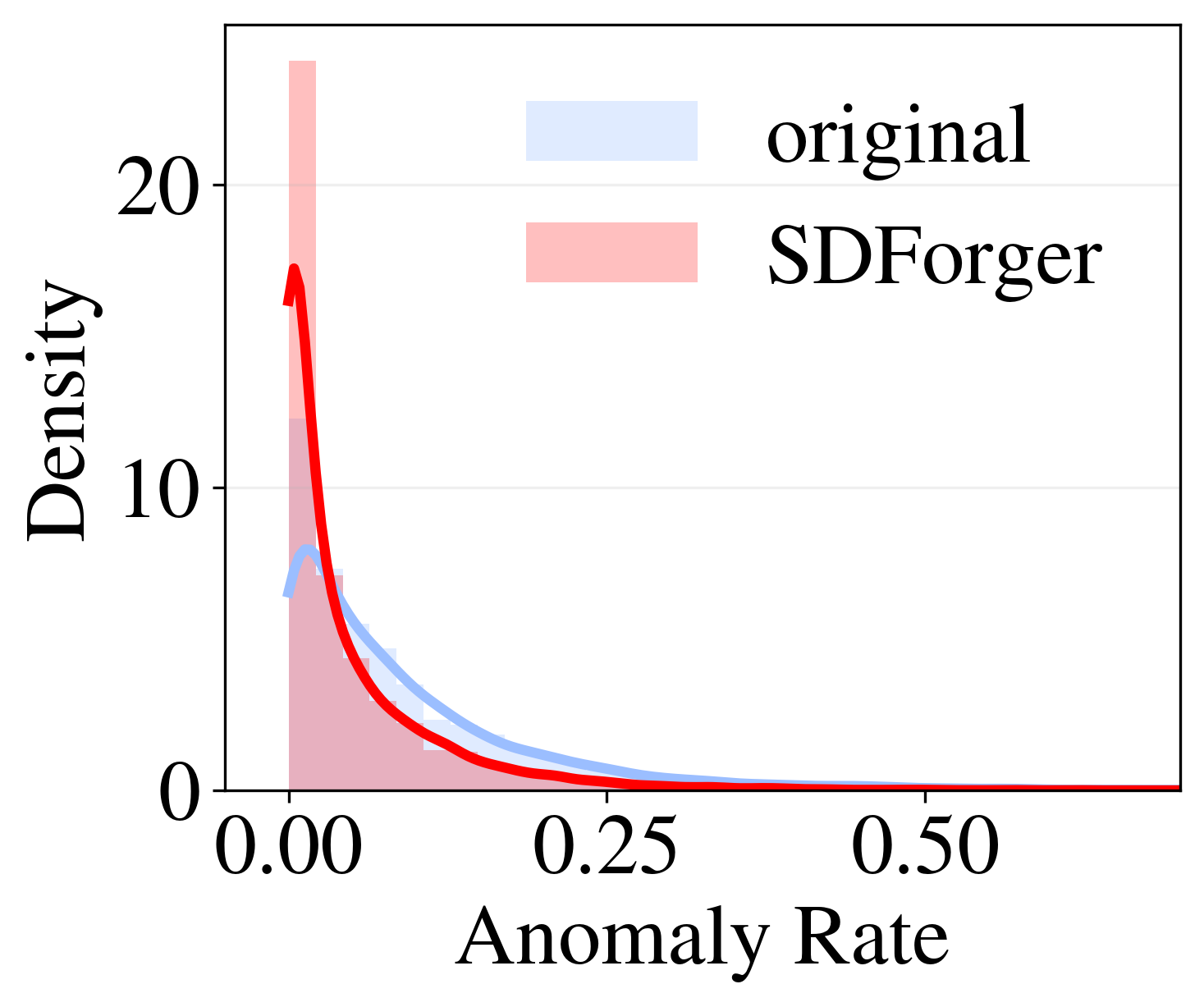}
        \caption{SDForger.}
        \label{fig:anomaly_rate_SDForger}
    \end{subfigure}

    \caption{Comparison of Sample-Level Anomaly-Rate Distributions.}
    \label{fig:comparison_anomaly_rate_distributions}
\end{figure*}

\begin{figure*}[htbp]
    \centering

    \begin{subfigure}[t]{0.235\textwidth}
        \centering
        \includegraphics[width=\linewidth]{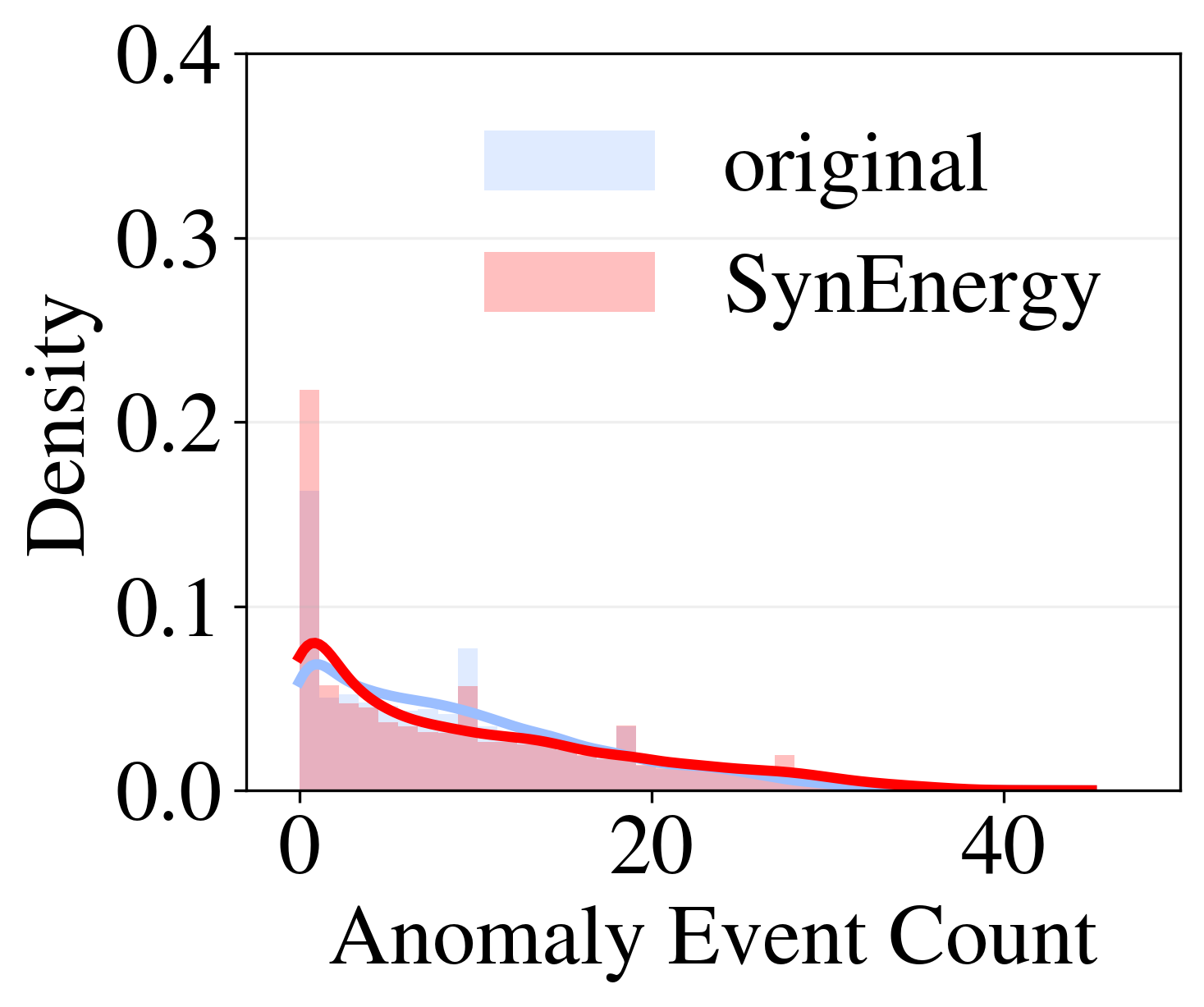}
        \caption{SynEnergy.}
        \label{fig:anomaly_event_count_synenergy}
    \end{subfigure}
    \hfill
    \begin{subfigure}[t]{0.235\textwidth}
        \centering
        \includegraphics[width=\linewidth]{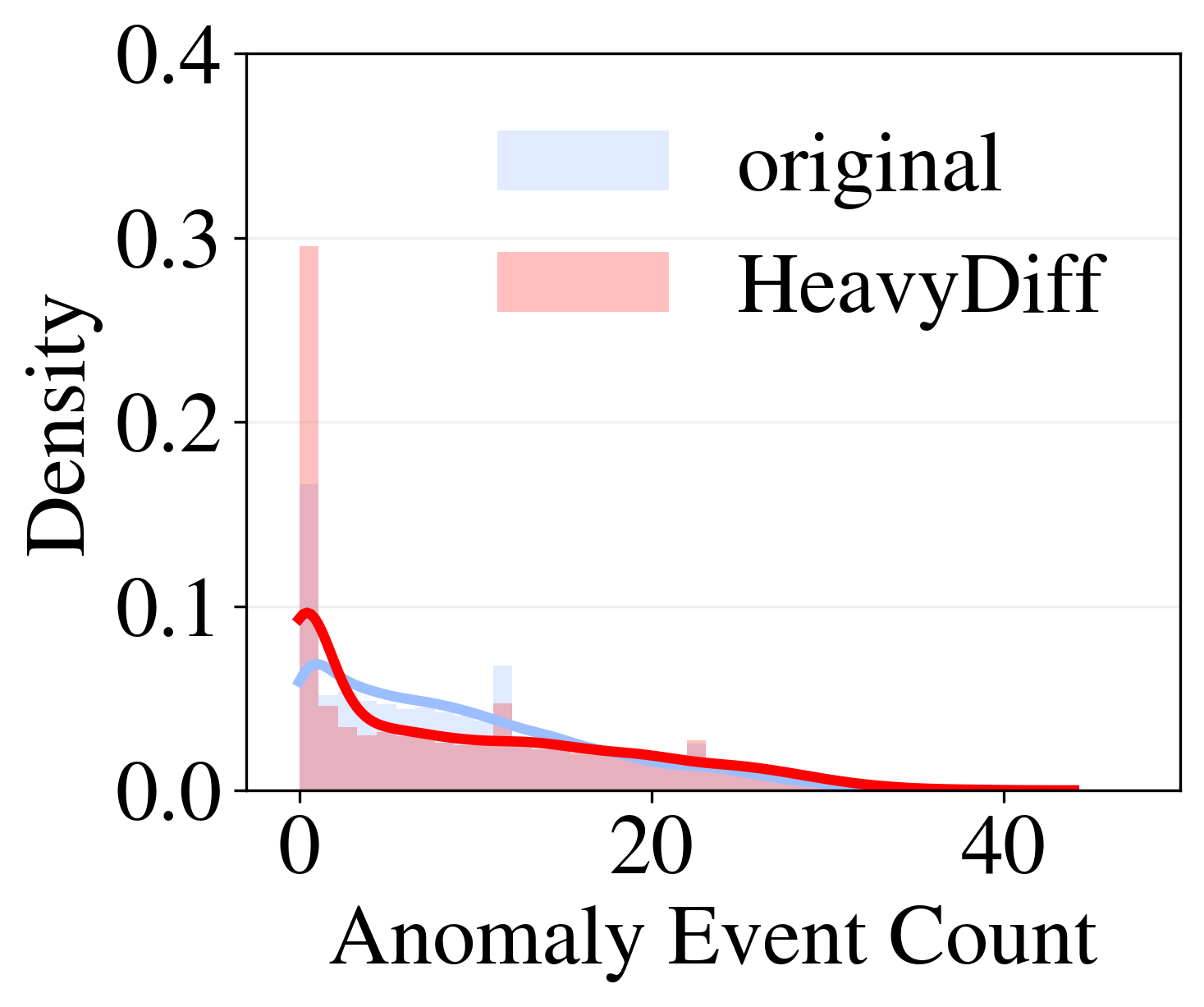}
        \caption{HeavyDiff.}
        \label{fig:anomaly_event_count_HeavyDiff}
    \end{subfigure}
    \hfill
    \begin{subfigure}[t]{0.235\textwidth}
        \centering
        \includegraphics[width=\linewidth]{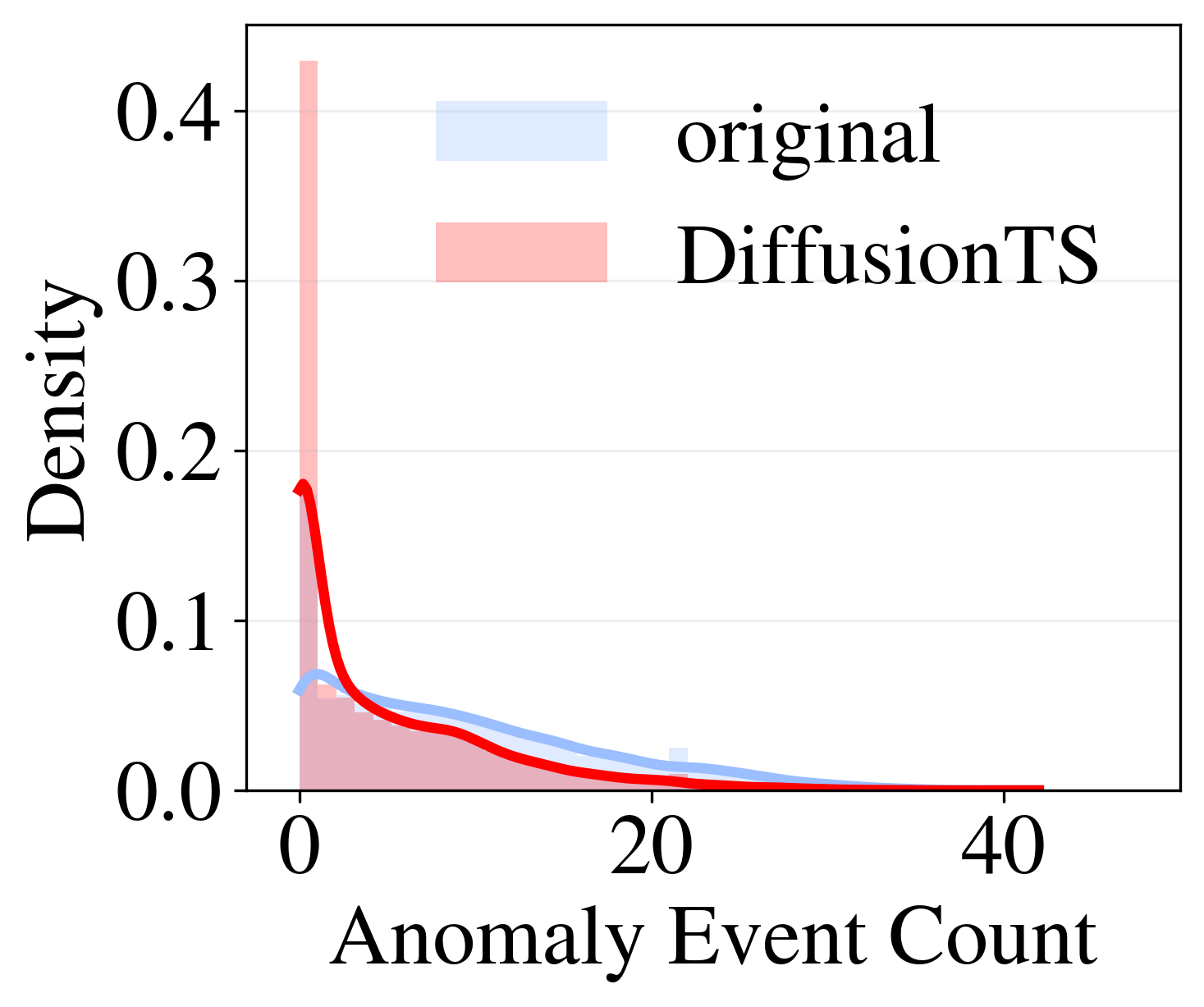}
        \caption{DiffusionTS.}
        \label{fig:anomaly_event_count_DiffusionTS}
    \end{subfigure}
    \hfill
    \begin{subfigure}[t]{0.235\textwidth}
        \centering
        \includegraphics[width=\linewidth]{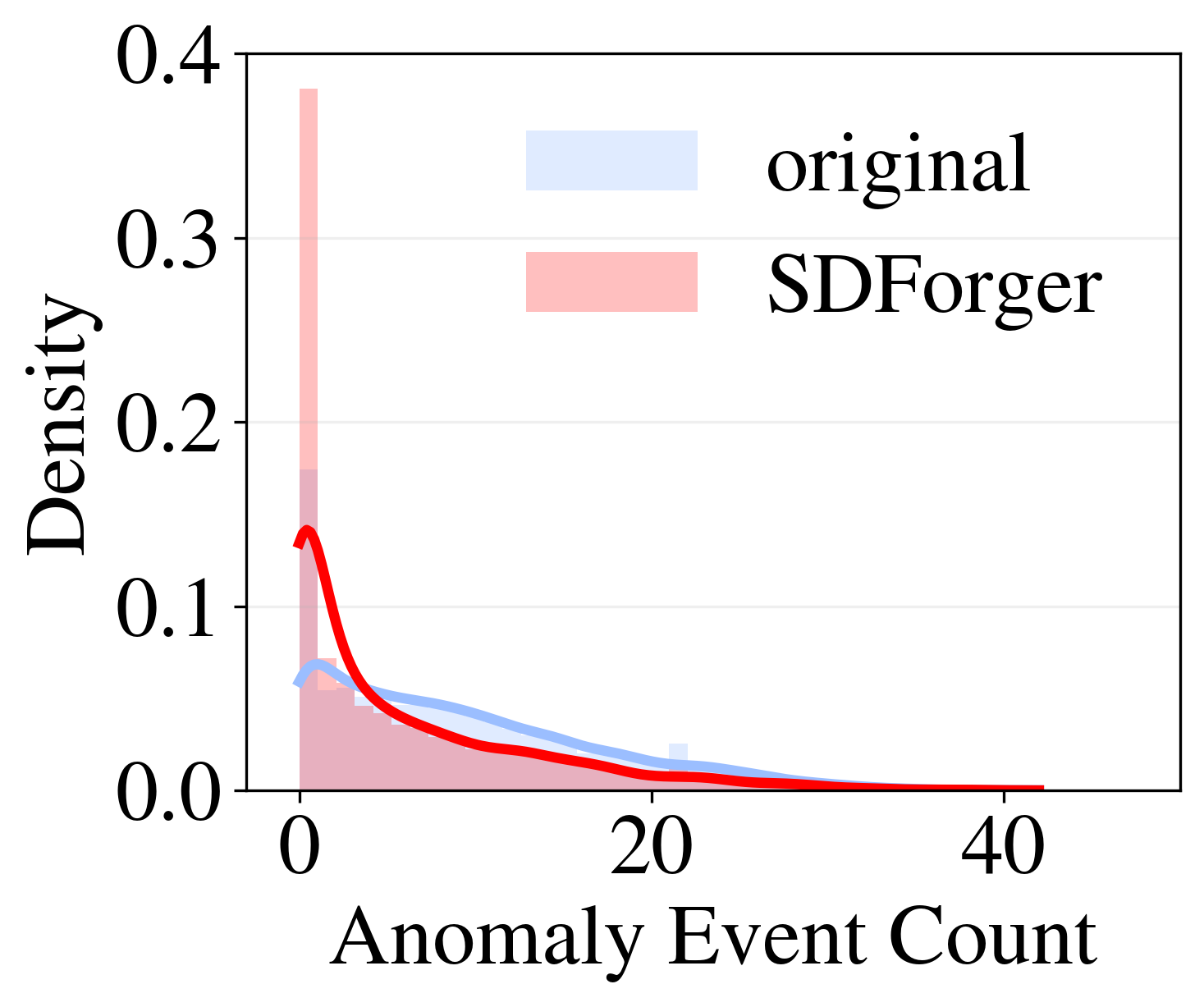}
        \caption{SDForger.}
        \label{fig:anomaly_event_count_SDForger}
    \end{subfigure}

    \caption{Comparison of Sample-Level Anomalous Event Count Distributions.}
    \label{fig:comparison_anomaly_event_count_distributions}
\end{figure*}

\begin{figure*}[htbp]
    \centering

    \begin{subfigure}[t]{0.235\textwidth}
        \centering
        \includegraphics[width=\linewidth]{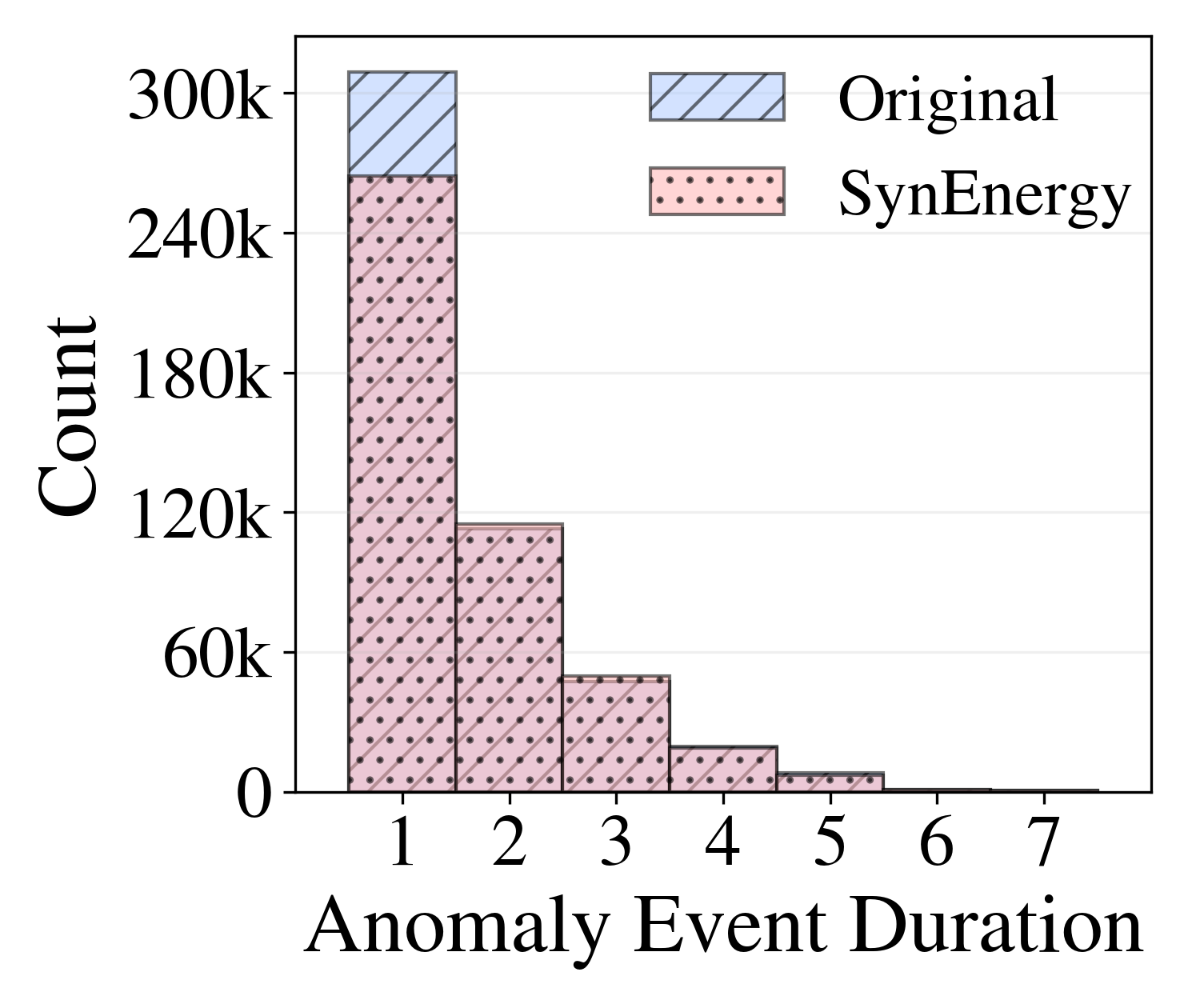}
        \caption{SynEnergy.}
        \label{fig:anomaly_duration_synenergy}
    \end{subfigure}
    \hfill
    \begin{subfigure}[t]{0.235\textwidth}
        \centering
        \includegraphics[width=\linewidth]{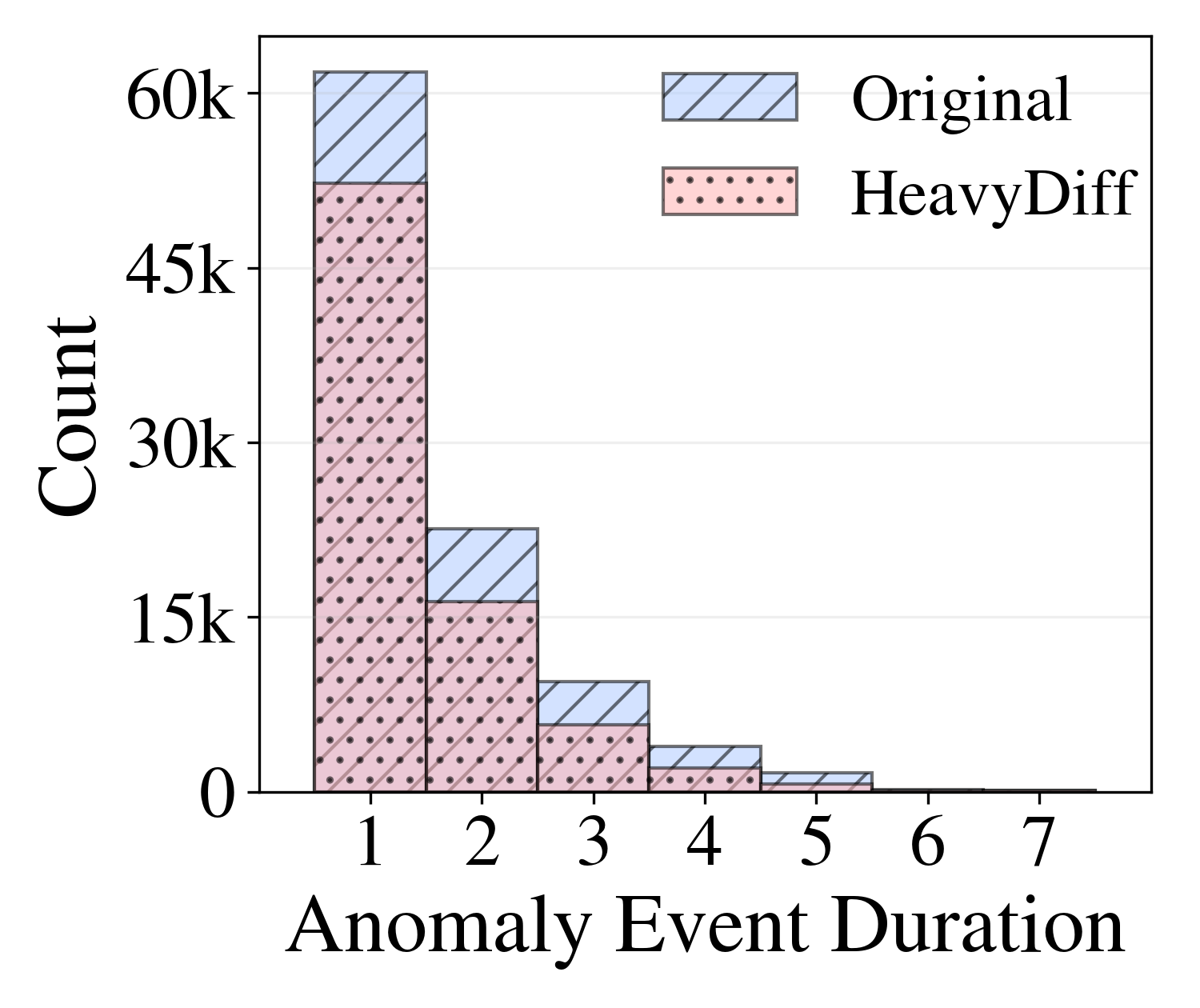}
        \caption{HeavyDiff.}
        \label{fig:anomaly_duration_HeavyDiff}
    \end{subfigure}
    \hfill
    \begin{subfigure}[t]{0.235\textwidth}
        \centering
        \includegraphics[width=\linewidth]{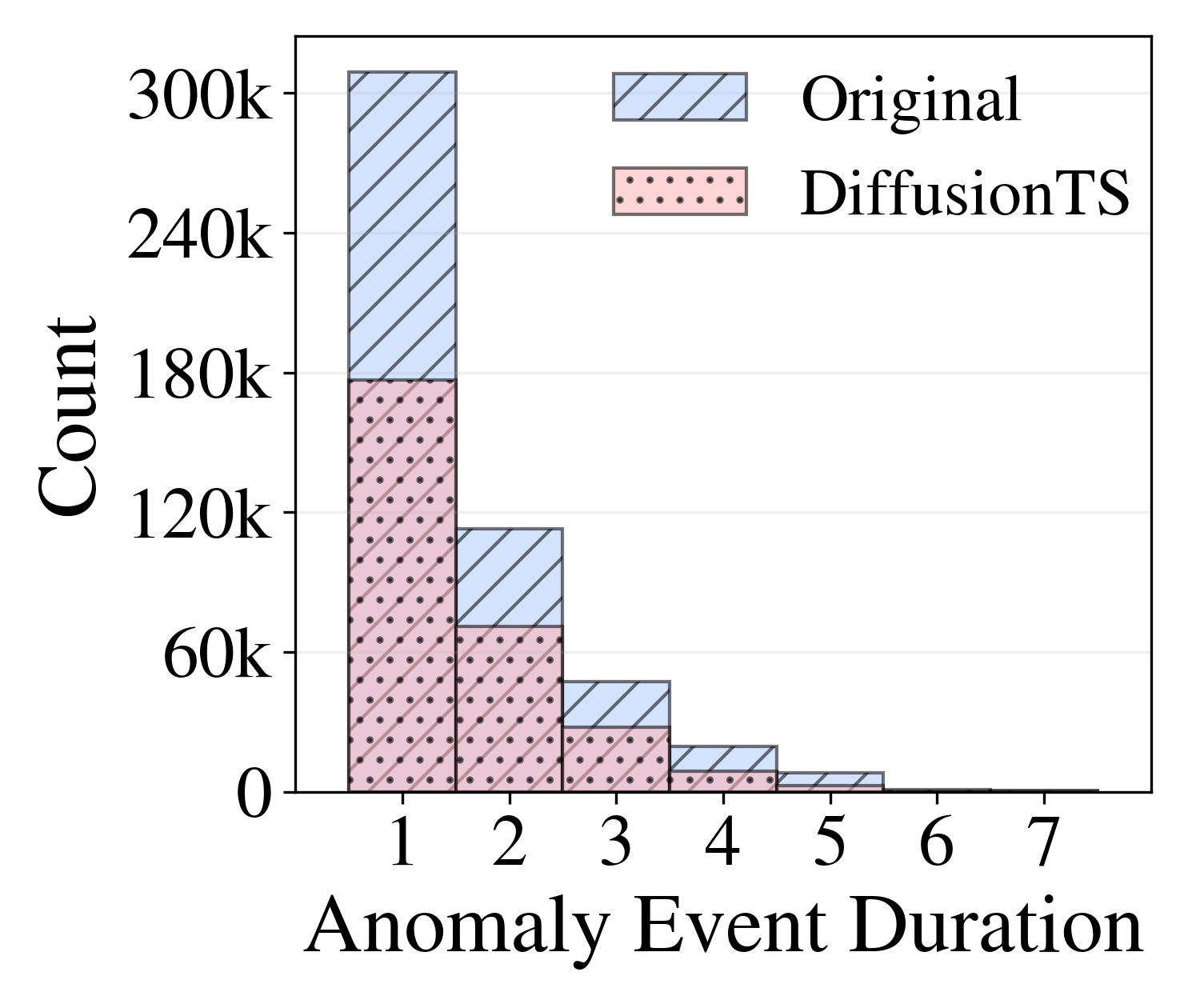}
        \caption{DiffusionTS.}
        \label{fig:anomaly_duration_DiffusionTS}
    \end{subfigure}
    \hfill
    \begin{subfigure}[t]{0.235\textwidth}
        \centering
        \includegraphics[width=\linewidth]{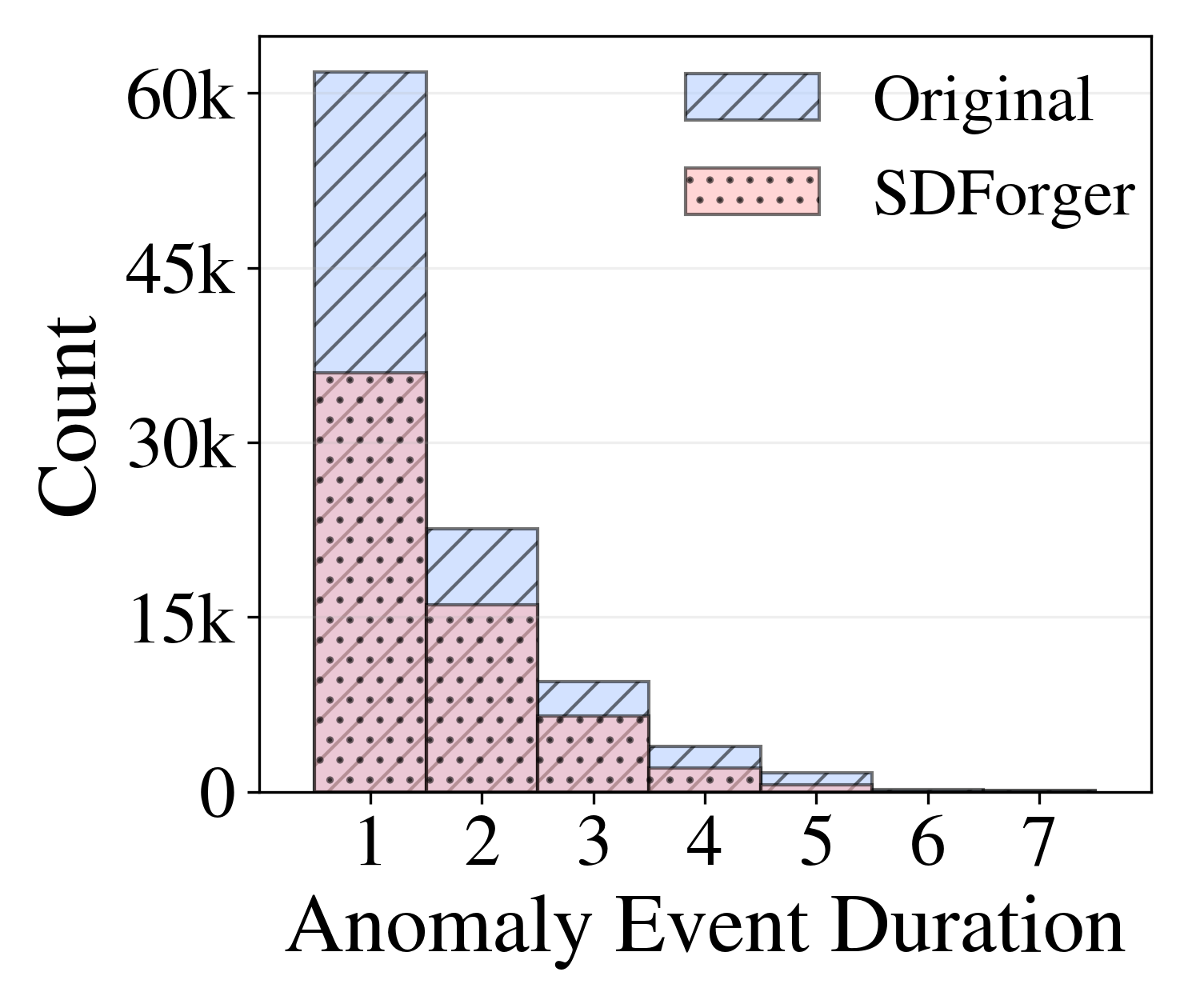}
        \caption{SDForger.}
        \label{fig:anomaly_duration_SDForger}
    \end{subfigure}

    \caption{Comparison of Sample-Level Anomalous Event Duration Distributions.}
    \label{fig:comparison_anomaly_event_duration_distributions}
\end{figure*}

\begin{figure*}[htbp]
    \centering

    \begin{subfigure}[t]{0.235\textwidth}
        \centering
        \includegraphics[width=\linewidth]{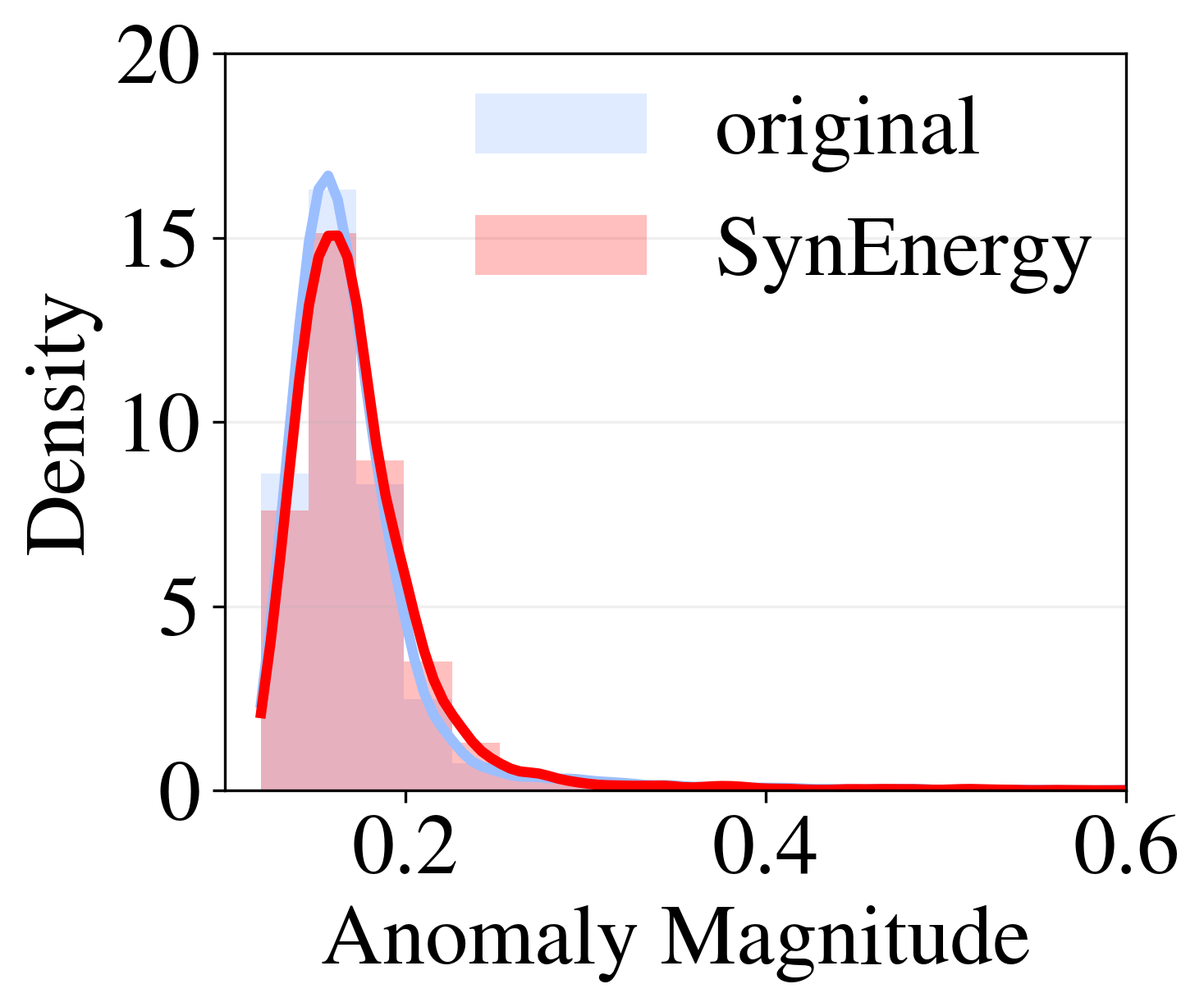}
        \caption{SynEnergy.}
        \label{fig:anomaly_magnitude_synenergy}
    \end{subfigure}
    \hfill
    \begin{subfigure}[t]{0.235\textwidth}
        \centering
        \includegraphics[width=\linewidth]{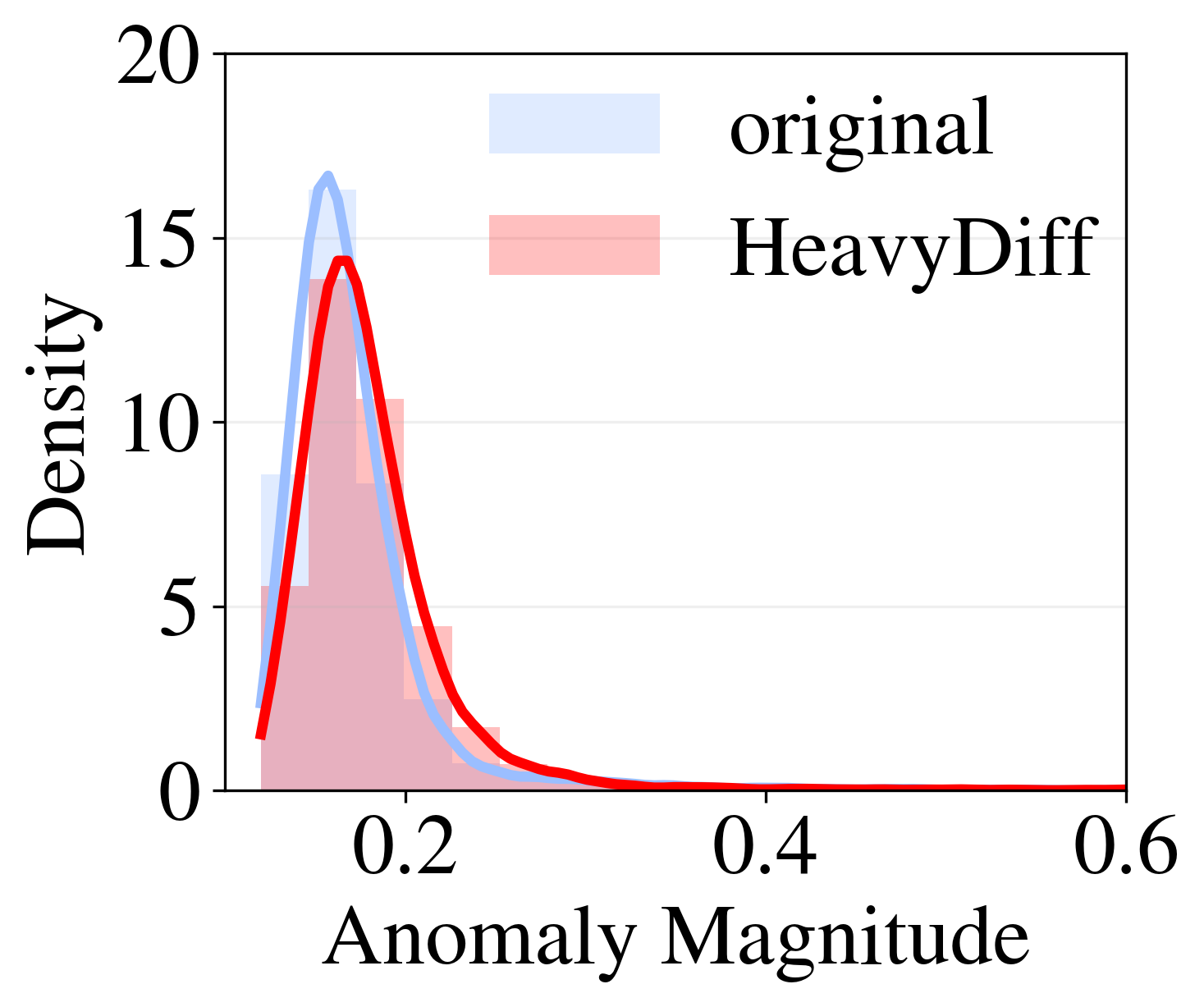}
        \caption{HeavyDiff.}
        \label{fig:anomaly_magnitude_HeavyDiff}
    \end{subfigure}
    \hfill
    \begin{subfigure}[t]{0.235\textwidth}
        \centering
        \includegraphics[width=\linewidth]{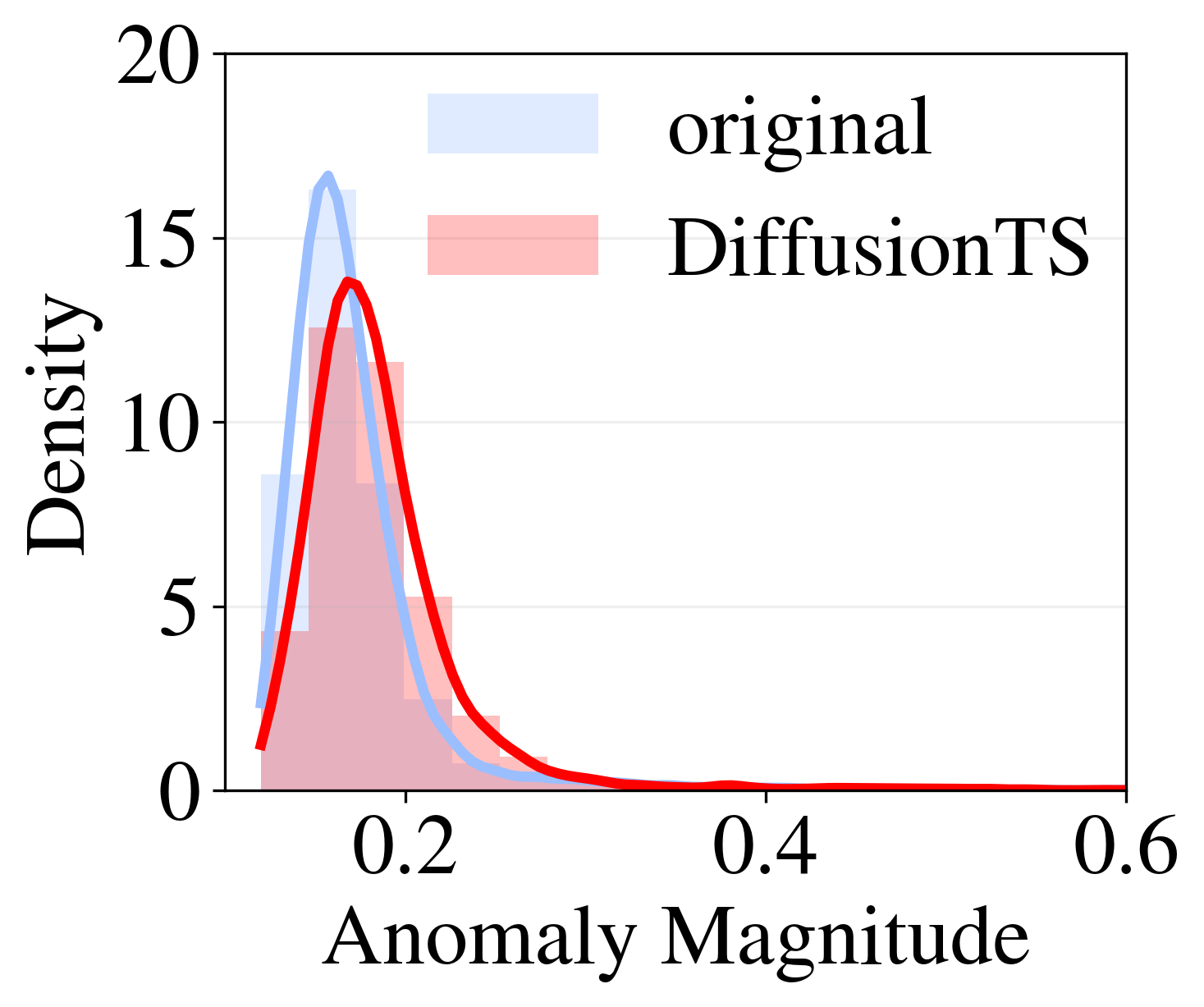}
        \caption{DiffusionTS.}
        \label{fig:anomaly_magnitude_DiffusionTS}
    \end{subfigure}
    \hfill
    \begin{subfigure}[t]{0.235\textwidth}
        \centering
        \includegraphics[width=\linewidth]{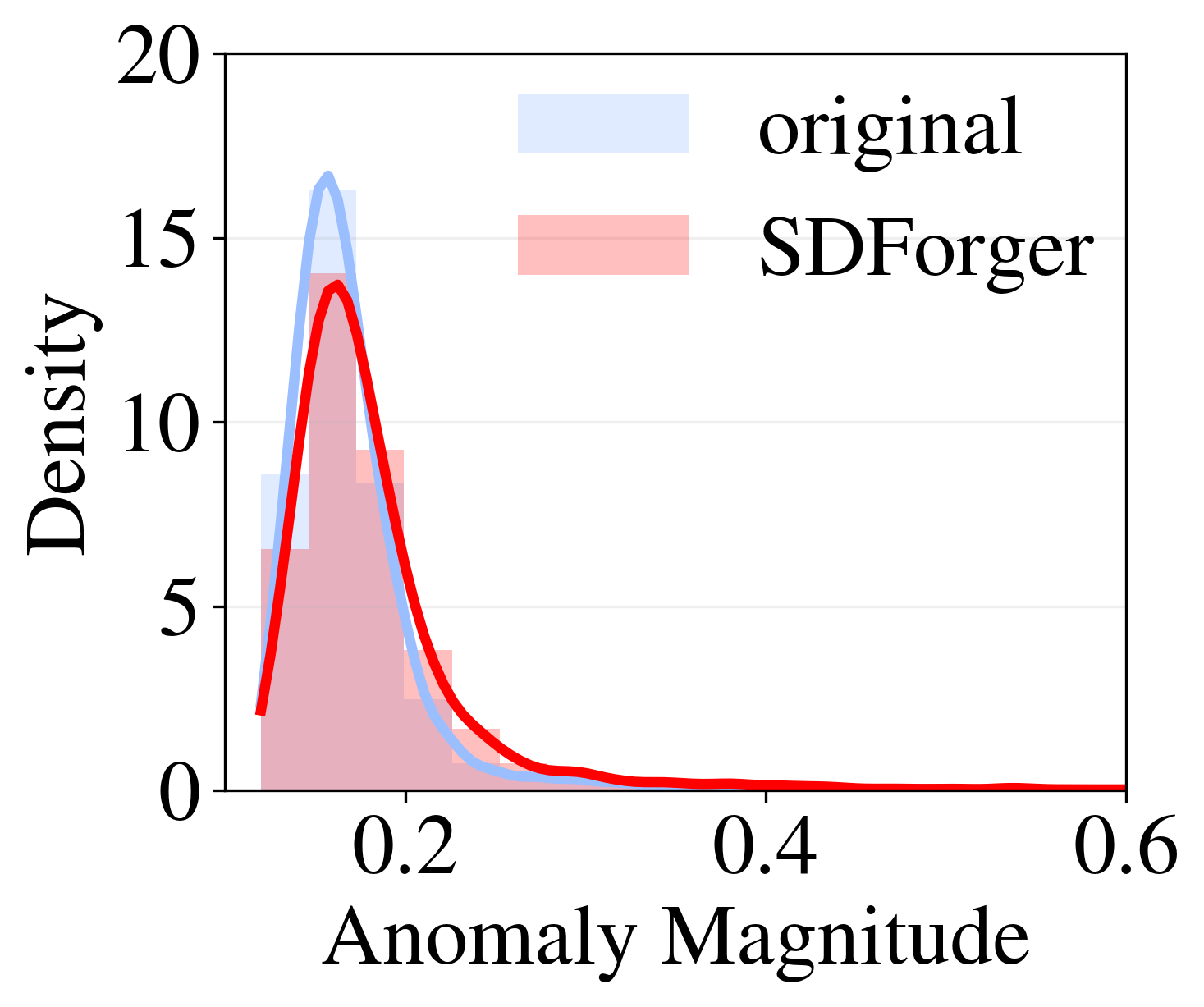}
        \caption{SDForger.}
        \label{fig:anomaly_magnitude_SDForger}
    \end{subfigure}

    \caption{Comparison of Sample-Level Anomaly Magnitude Distributions.}
    \label{fig:comparison_anomaly_magnitude_distributions}
\end{figure*}

\begin{figure*}[htbp]
    \centering

    \begin{subfigure}[t]{0.235\textwidth}
        \centering
        \includegraphics[width=\linewidth]{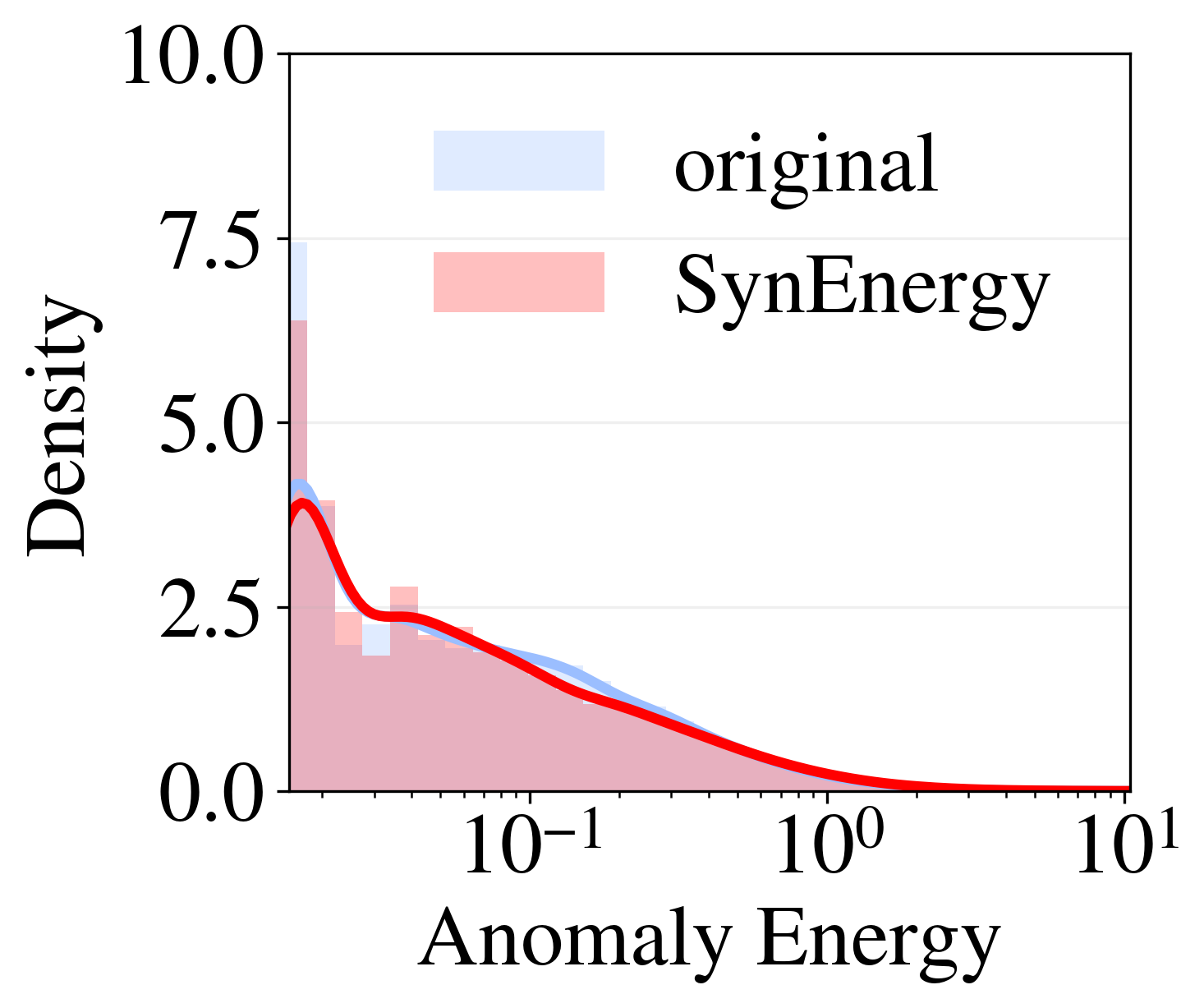}
        \caption{SynEnergy.}
        \label{fig:anomaly_energy_synenergy}
    \end{subfigure}
    \hfill
    \begin{subfigure}[t]{0.235\textwidth}
        \centering
        \includegraphics[width=\linewidth]{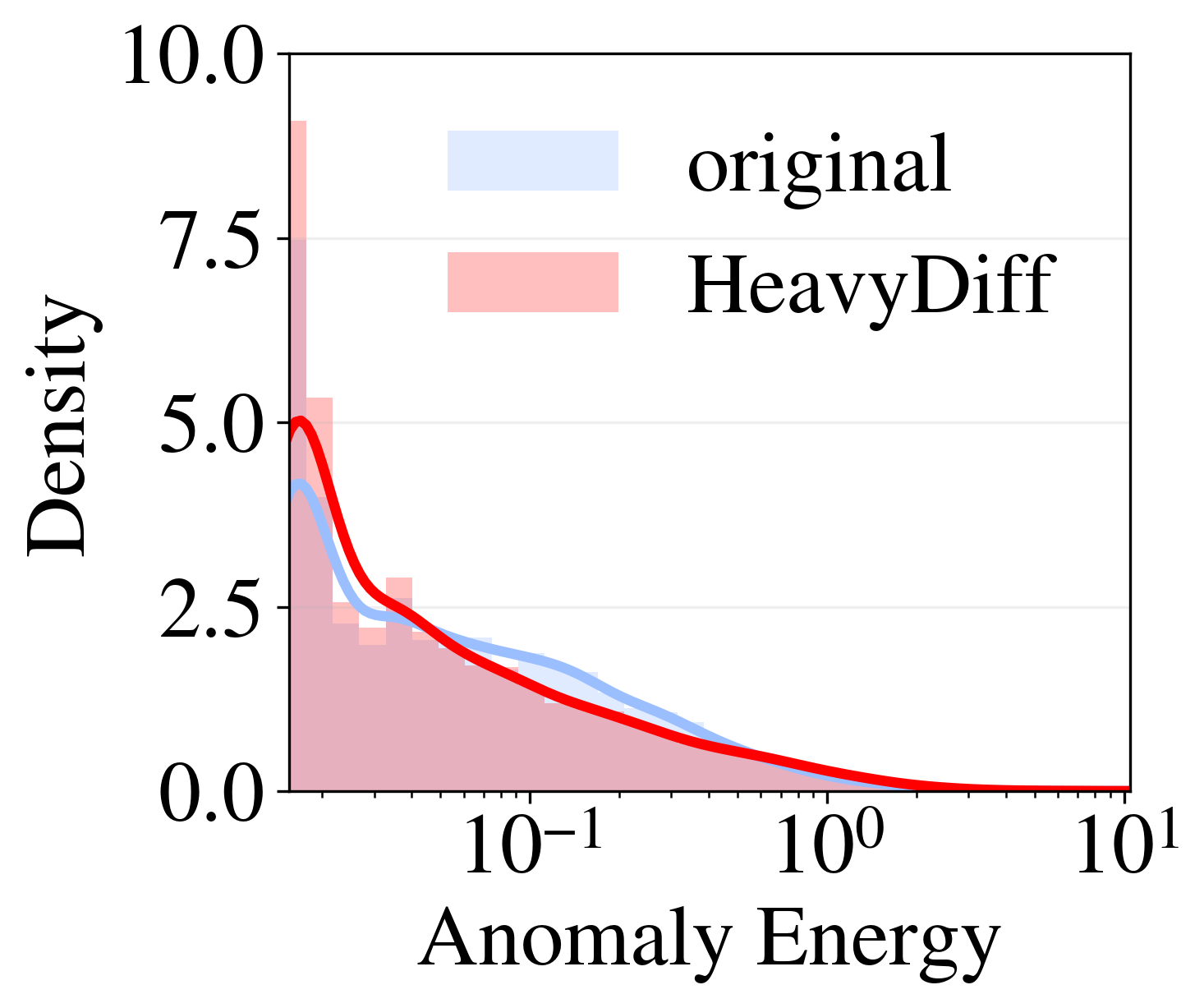}
        \caption{HeavyDiff.}
        \label{fig:anomaly_energy_HeavyDiff}
    \end{subfigure}
    \hfill
    \begin{subfigure}[t]{0.235\textwidth}
        \centering
        \includegraphics[width=\linewidth]{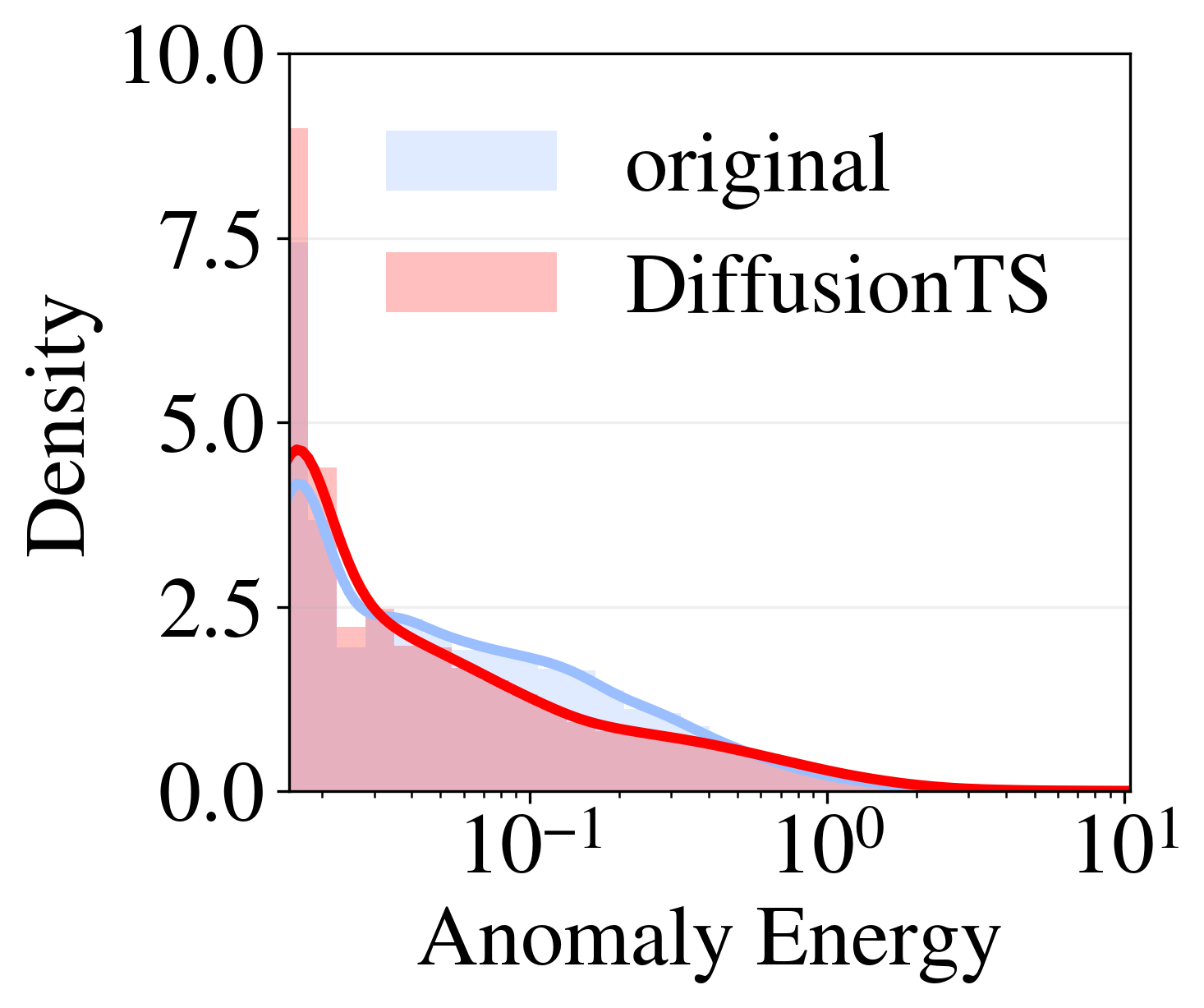}
        \caption{DiffusionTS.}
        \label{fig:anomaly_energy_DiffusionTS}
    \end{subfigure}
    \hfill
    \begin{subfigure}[t]{0.235\textwidth}
        \centering
        \includegraphics[width=\linewidth]{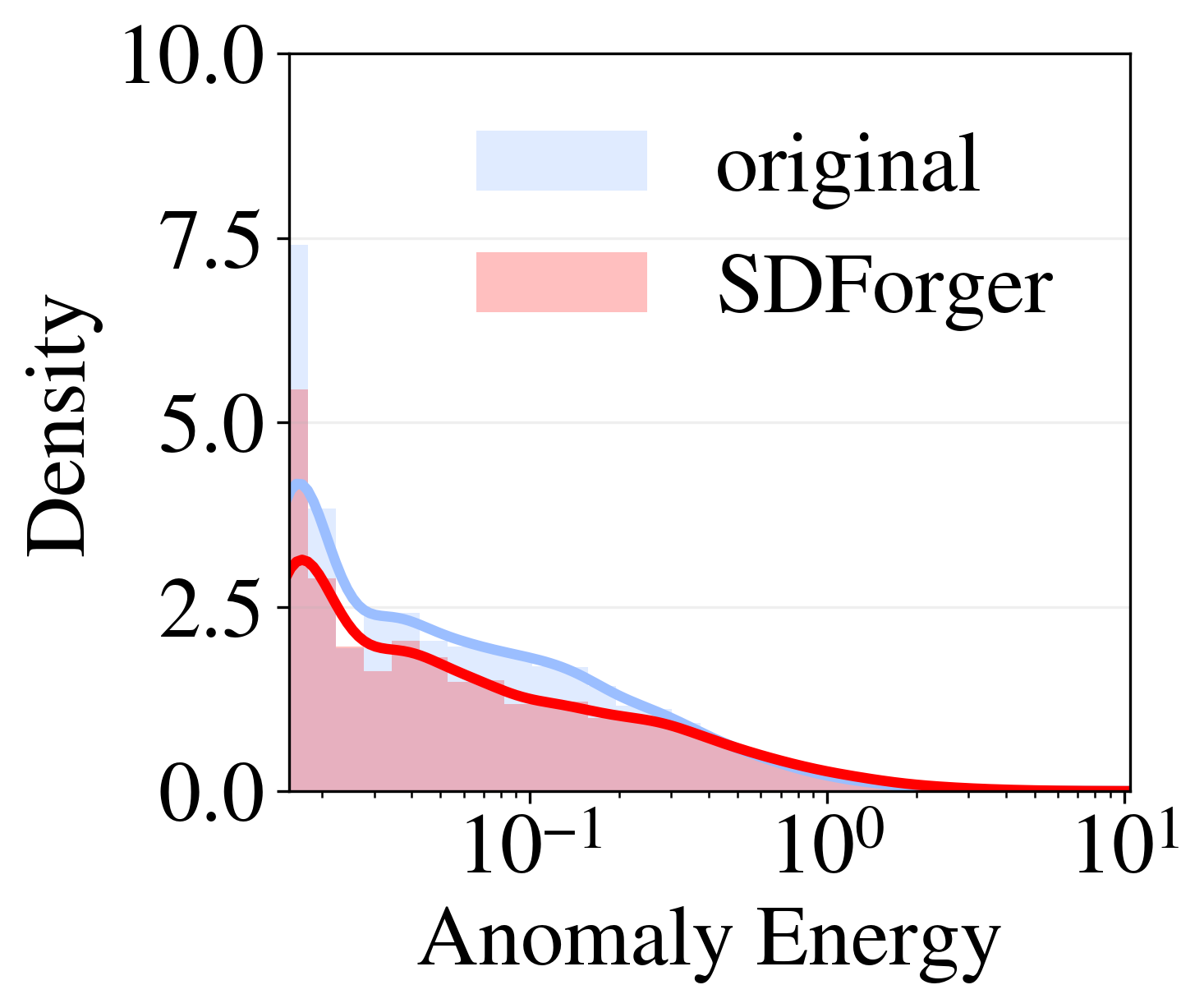}
        \caption{SDForger.}
        \label{fig:anomaly_energy_SDForger}
    \end{subfigure}

    \caption{Comparison of Sample-Level Anomaly Energy Distributions.}
    \label{fig:comparison_anomaly_energy_distributions}
\end{figure*}

\begin{figure*}[htbp]
    \centering

    \begin{subfigure}[t]{0.235\textwidth}
        \centering
        \includegraphics[width=\linewidth]{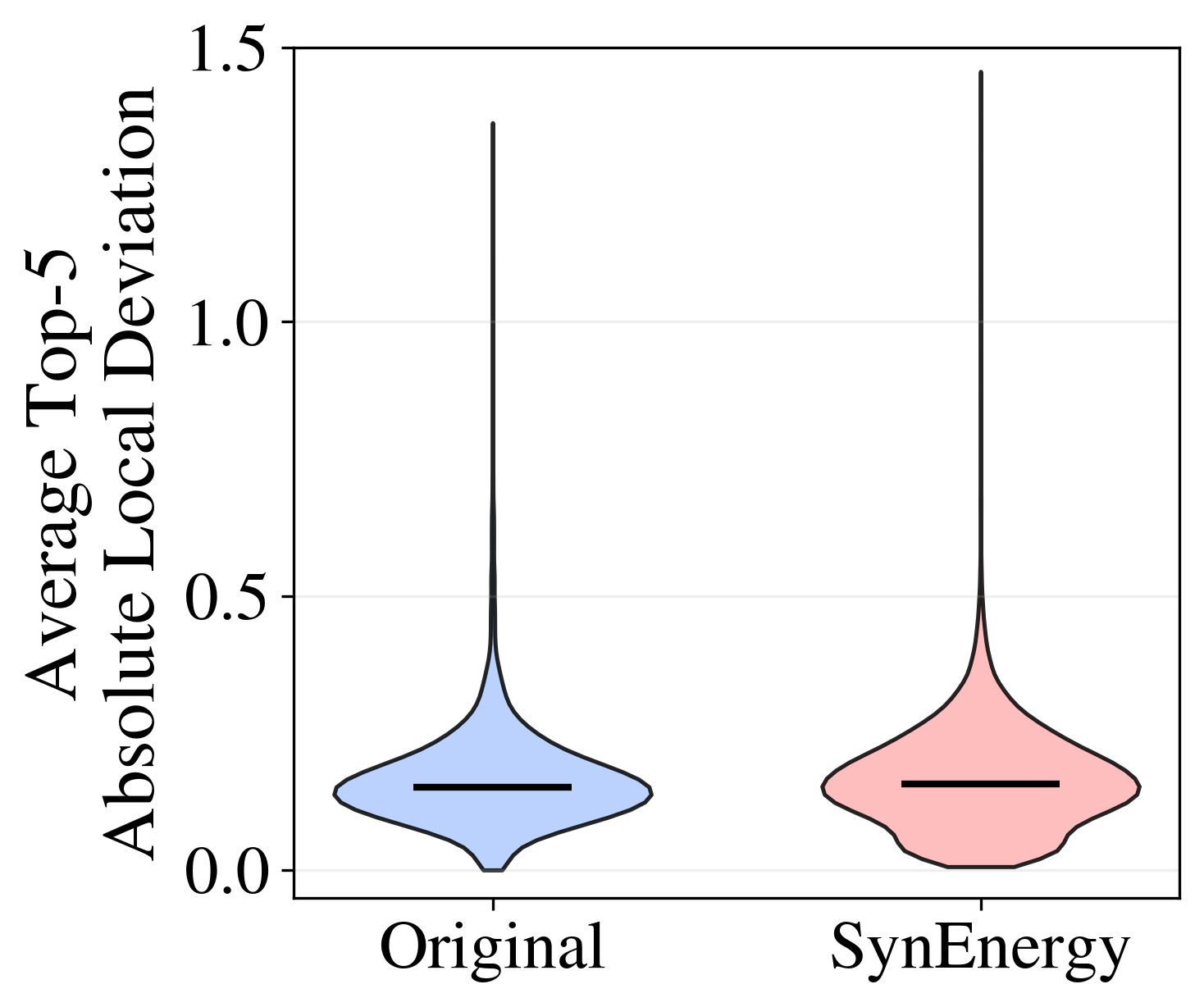}
        \caption{SynEnergy.}
        \label{fig:anomaly_deviation_violinplot_synenergy}
    \end{subfigure}
    \hfill
    \begin{subfigure}[t]{0.235\textwidth}
        \centering
        \includegraphics[width=\linewidth]{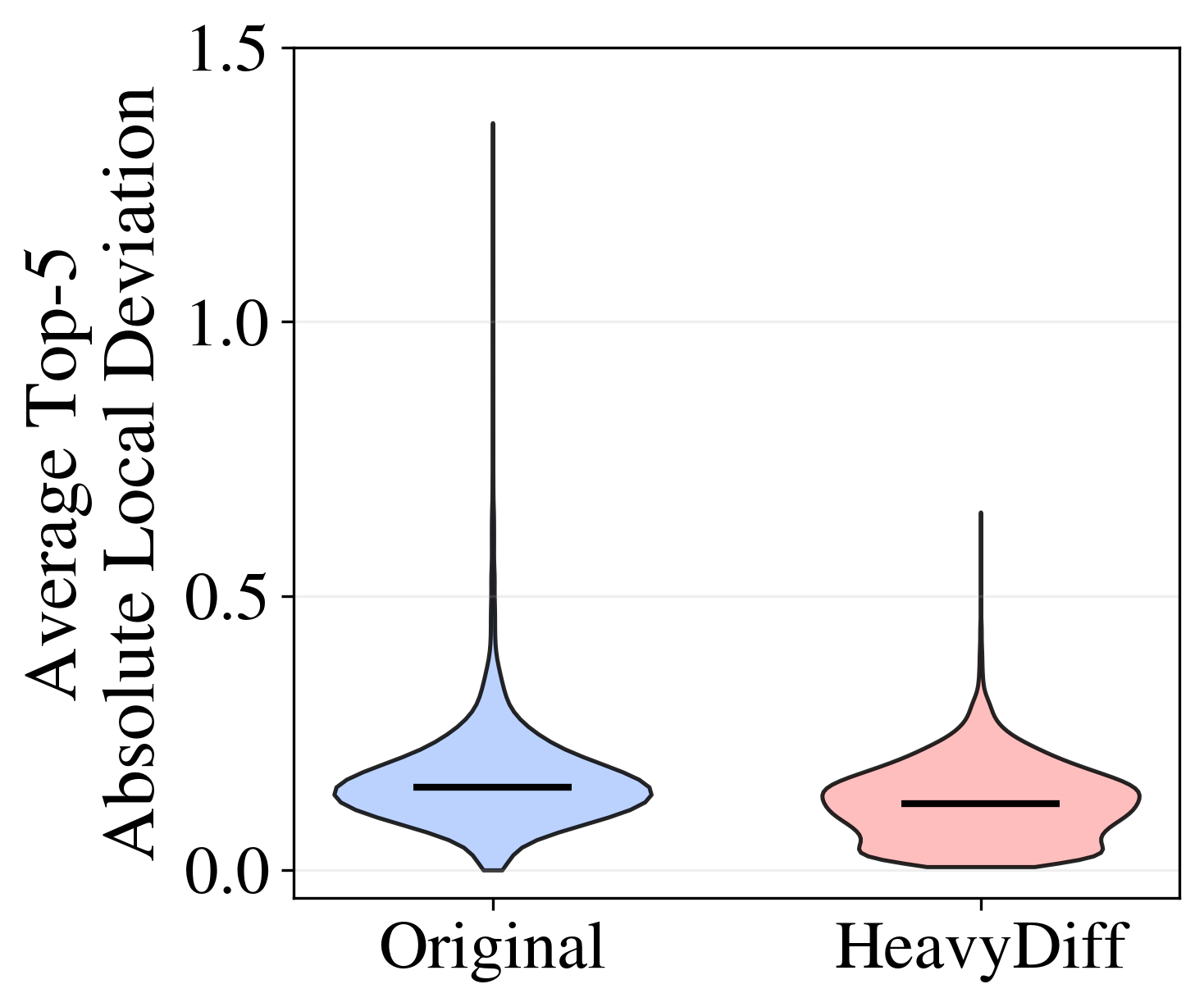}
        \caption{HeavyDiff.}
        \label{fig:anomaly_deviation_violinplot_HeavyDiff}
    \end{subfigure}
    \hfill
    \begin{subfigure}[t]{0.235\textwidth}
        \centering
        \includegraphics[width=\linewidth]{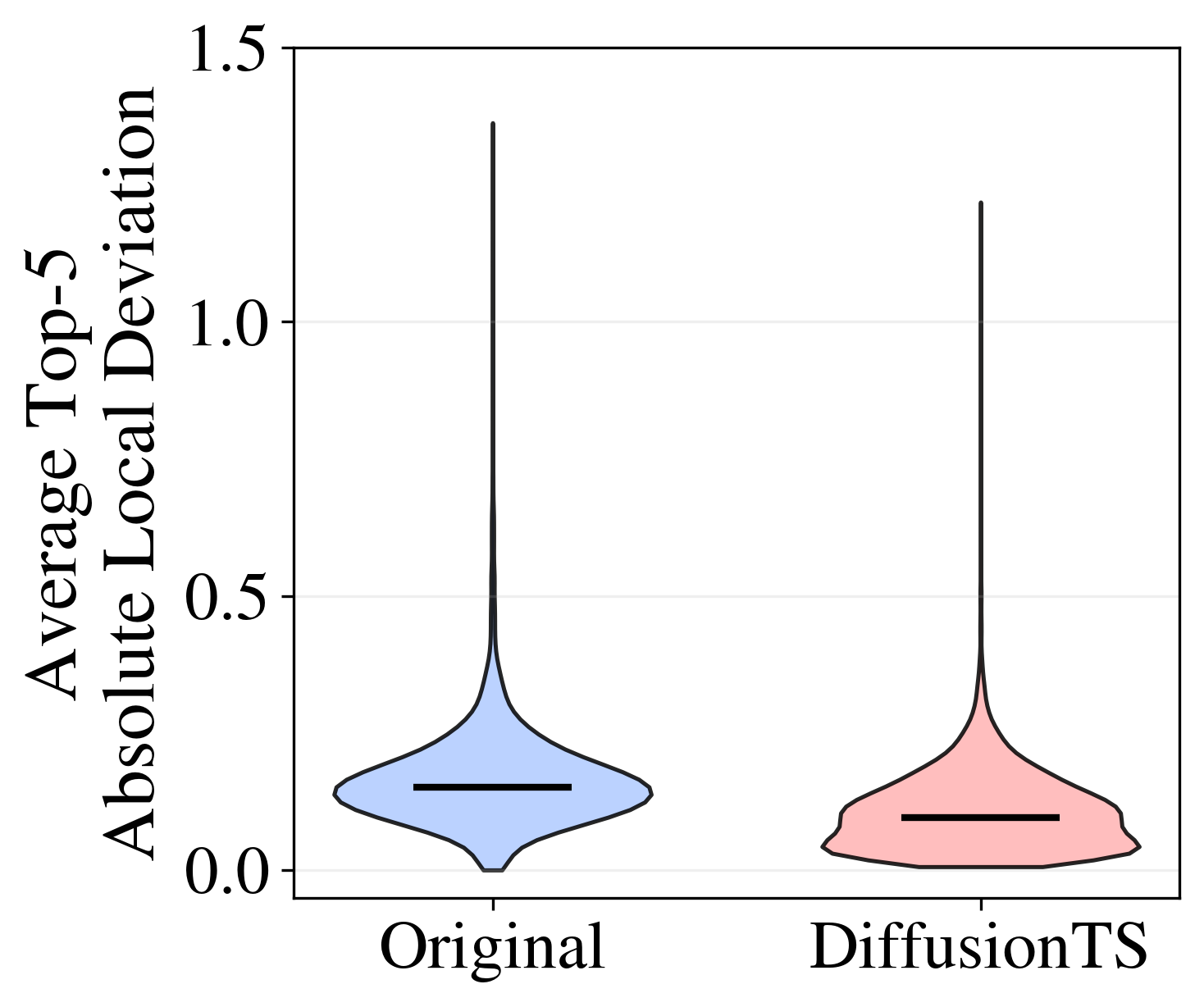}
        \caption{DiffusionTS.}
        \label{fig:anomaly_deviation_violinplot_DiffusionTS}
    \end{subfigure}
    \hfill
    \begin{subfigure}[t]{0.235\textwidth}
        \centering
        \includegraphics[width=\linewidth]{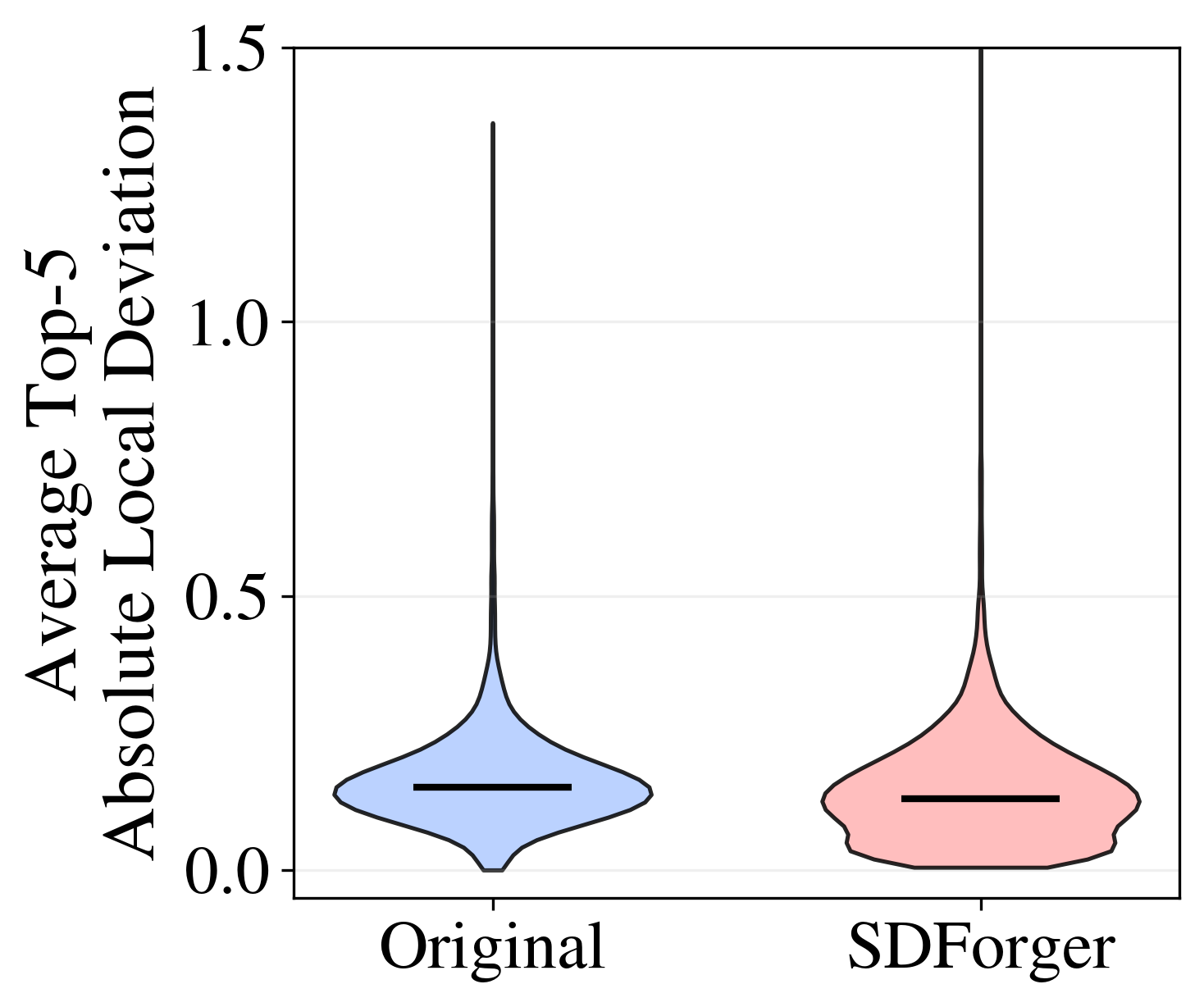}
        \caption{SDForger.}
        \label{fig:anomaly_deviation_violinplot_SDForger}
    \end{subfigure}

    \caption{Violin-Plot Comparison of Sample-Level Anomaly Deviation Distributions.}
    \label{fig:comparison_deviation_violinplot_distributions}
\end{figure*}

As shown in Figures~\ref{fig:comparison_anomaly_rate_distributions}, \ref{fig:comparison_anomaly_event_count_distributions}, \ref{fig:comparison_anomaly_event_duration_distributions}, \ref{fig:comparison_anomaly_magnitude_distributions}, \ref{fig:comparison_anomaly_energy_distributions}, and \ref{fig:comparison_deviation_violinplot_distributions}, we provide additional sample-level distribution comparisons from six perspectives: anomaly rate, anomalous event count, event duration, event magnitude, anomaly energy, and anomaly deviation. For each perspective, we compare \m with the three strongest baselines, namely HeavyDiff, Diffusion-TS, and SDForger. Across all comparisons, the distributions generated by \m consistently align more closely with the original data than those produced by the baselines.

\subsection{Generation Scalability and Granularity Analysis}\label{appendix:scalability}
As shown in Tables~\ref{tab:scalability_household_number} and~\ref{tab:temporal_granularity}, \m maintains a clear advantage under both small-scale and large-scale settings, without exhibiting systematic performance degradation as the number of households increases. Its advantage is particularly stable on anomaly preservation metrics, indicating that the proposed anomaly semantic modeling remains effective despite substantial changes in sample size. Similarly, across different temporal intervals, \m effectively captures both fine-grained fluctuations and aggregated consumption patterns; notably, at the one-hour granularity, it achieves the best results on all four overall generation fidelity metrics and three of the four anomaly fidelity metrics. Although individual downstream results vary across settings, \m remains highly competitive overall, confirming its applicability to datasets with different scales and temporal resolutions.

\subsection{Sensitivity to the Anomaly Threshold \texorpdfstring{$\delta$}{delta}}
\label{appendix:threshold}
As shown in Table~\ref{tab:threshold_sensitivity}, moderate threshold settings provide the most balanced performance across the three evaluation dimensions. In particular, the top-5\% setting yields the strongest anomaly preservation fidelity, while the top-10\% setting maintains a favorable balance between overall fidelity, anomaly preservation, and downstream utility. A highly restrictive threshold at 1\% identifies only a small number of extreme residuals and therefore provides insufficient coverage of diverse anomalous patterns, whereas the broader 20\% setting introduces more regular variations into the anomaly set and weakens the discriminative anomaly guidance. These results support the use of the top-10\% threshold as the default setting for \m.

\subsection{Ablation Study for \m}\label{appendix:ablation}
As shown in Table~\ref{tab:ablation}, both ablated variants exhibit substantial performance degradation on the FL1 and FL2 datasets, particularly on the anomaly preservation metrics. Replacing HG-ASL with a simple encoder reduces the performance of \m to a level comparable to Diffusion-TS, indicating that directly encoding anomalous residuals is insufficient to capture their complex structures. Moreover, the small performance gap between \m w/o HG-ASL and \m w/o AG suggests that anomaly information provides limited guidance without an effective semantic representation. These results confirm that both the anomaly semantics learned by HG-ASL and their dedicated injection into the diffusion process through AS-Diff are essential to the effectiveness of \m.

\subsection{Computational Efficiency Analysis}\label{appendix:efficiency}
As shown in Table~\ref{tab:efficiency}, \m exhibits practical computational efficiency across all four datasets. Based on the average per-step runtime, training for 10,000 optimization steps takes approximately 20 minutes on FL1 and FL2 and 8.3 minutes on NY and CA. Similarly, generating 10,000 samples takes approximately 31.7 minutes on FL1 and FL2 and 13.3 minutes on NY and CA. These results demonstrate that \m supports efficient training and large-scale synthetic data generation across datasets with different sizes and sequence configurations.

\begin{table}[htbp]
\centering
\small
\renewcommand{\arraystretch}{1.15}
\setlength{\tabcolsep}{5pt}

\vspace{-6pt}
\caption{Training and sampling efficiency of \m across the four datasets.}
\label{tab:efficiency}
\begin{tabular}{lcc}
\toprule
\textbf{Dataset}
& \makecell{\textbf{Training Time}\\\textbf{(s/step)}}
& \makecell{\textbf{Sampling Time}\\\textbf{(s/sample)}} \\
\midrule

\textbf{FL1} & 0.12 & 0.19 \\
\textbf{FL2} & 0.12 & 0.19 \\
\textbf{NY}  & 0.05 & 0.08 \\
\textbf{CA}  & 0.05 & 0.08 \\

\bottomrule
\end{tabular}
\end{table}

\section{Related Work}\label{appendix:related_work}

\subsection{Synthetic Energy Data Generation}
Synthetic energy data~\cite{yu2026trustenergy,yu2026energymamba} generation has emerged as an important solution to the scarcity and privacy concerns associated with fine-grained consumption records. Research in this area has progressed from statistical and behavior-based modeling to deep generative approaches, including Generative Adversarial Networks (GANs)~\cite{li2026neurogripretrievalaugmentedgraphrefinement} and diffusion models. Early studies relied on statistical distributions, stochastic processes, and domain-specific behavioral assumptions to generate energy consumption profiles~\cite{kang2023systematic}. For example, Thorve et al.~\cite{thorve2023high} combined population surveys with building-level physical properties to model energy consumption, while Yuan et al.~\cite{yuan2023synthetic} integrated empirical consumption data with probabilistic models of distributed energy resources. Although these approaches incorporate useful domain knowledge, their reliance on predefined distributions and behavioral assumptions limits their ability to capture complex and volatile consumption patterns.

With the development of deep generative models, GAN-based methods have been introduced to learn energy consumption distributions directly from data. Hu et al.~\cite{hu2023multiload} proposed MultiLoad-GAN to generate groups of consumption profiles while preserving spatio-temporal correlations, whereas Razghandi et al.~\cite{razghandi2023smart} combined VAE and GAN architectures to generate synthetic smart-home electricity data for energy management applications. More recently, diffusion models have attracted increasing attention due to their stable training and effective distribution modeling. Fu et al.~\cite{fu2024creating} developed a conditional diffusion model that incorporates building metadata for synthetic energy data generation, while Fuest et al.~\cite{fuest2025cents} proposed CENTS to synthesize energy consumption time series under rare and previously unseen scenarios. Despite this progress, existing methods primarily optimize overall distributional and temporal fidelity. Rare but important anomalous events are typically treated as part of the overall data distribution rather than explicitly modeled and preserved, limiting their representation in generated energy consumption data and motivating the design of \m.

\subsection{Anomaly-Aware Time-series Generation}
Anomaly-aware time-series generation is an emerging research direction that aims to improve the representation of rare, extreme, and heavy-tailed patterns in synthetic data. Early studies mainly focused on modeling extreme-value distributions. Allouche et al.~\cite{allouche2022ev} incorporated extreme value theory into GANs to improve tail-event generation, while Hasan et al.~\cite{hasan2022modeling} introduced structurally constrained neural networks to capture multivariate extreme-value dependencies. Finzi et al.~\cite{finzi2023user} further enabled diffusion models to generate trajectories satisfying user-defined rare-event conditions in physical dynamical systems.

More recent studies have modified the diffusion process to better model heavy-tailed distributions. Shariatian et al.~\cite{shariatian2025denoising} replaced Gaussian perturbations with L\'evy-based noise, whereas Pandey et al.~\cite{pandey2025heavy} developed Student-\(t\)-based diffusion and flow models to improve tail coverage and rare-event generation. Extending these ideas to time-series data, Galib et al.~\cite{galib2024fide} proposed FIDE, which combines frequency-domain enhancement with extreme-value conditioning to preserve extreme observations during diffusion generation. Jiang et al.~\cite{jiang2026e4gen} further introduced event-level controls for generating temporally structured extreme events with explicit timing and pattern guidance. However, most of these methods are designed for general time series and do not account for the characteristics of energy consumption data. In particular, they do not jointly model its strong temporal periodicity and spatial correlations across regions, which are essential for preserving city-scale anomalous events such as widespread outages and heatwave-driven demand surges.

\subsection{Diffusion-based Time-series Generation}
Diffusion models have emerged as an effective framework for time-series generation because of their stable training and ability to model complex temporal distributions. Diffusion-TS~\cite{yuan2024diffusionts} introduced an encoder--decoder Transformer with explicit trend--seasonality decomposition and direct clean-sequence reconstruction for interpretable time-series generation. TSDiff~\cite{kollovieh2023predict} trained an unconditional diffusion model that supports generation, forecasting, and refinement through self-guided sampling. Bilo\v{s} et al.~\cite{bilovs2023modeling} formulated diffusion in function space to model continuous temporal processes and accommodate irregularly sampled observations.

For conditional generation, Time Weaver~\cite{narasimhan2024time} incorporated categorical, continuous, and time-varying metadata to guide the synthesis of context-specific time series. Other studies have focused on temporal representation and generation efficiency. Yan et al.~\cite{yan2024probabilistic} proposed a decomposable denoising diffusion model that combines efficient probability paths with sequence modeling of local and global dependencies. ImagenTime~\cite{naiman2024utilizing} transformed time series into image representations, enabling vision diffusion models to handle sequences with varying lengths and temporal structures. More recently, PaD-TS~\cite{li2025population} introduced population-aware training to preserve dataset-level value distributions and cross-variable dependencies. Despite their effective temporal modeling capabilities, these methods mainly optimize general temporal and distributional fidelity. They lack a dedicated design for extracting, representing, and injecting anomaly semantics into the diffusion process, making it difficult to explicitly control and preserve anomalous patterns during generation. This limitation is particularly important for city-scale energy data, where anomalies exhibit distinct temporal structures and coordinated impacts across regions.

\clearpage
\begin{table*}[!t]
\centering
\footnotesize
\renewcommand{\arraystretch}{1.0}
\setlength{\tabcolsep}{2pt}

\caption{Overall generation fidelity, anomaly preservation fidelity, and downstream quality on the FL2 dataset. Best results are highlighted in bold, and second-best results are underlined. $\uparrow$ and $\downarrow$ indicate that higher and lower values are better, respectively.}%\vspace{-3pt}
\label{tab:FL2}
\begin{tabular}{cclcccccccccc}
\toprule
\multirow{2}{*}{\textbf{Dataset}} 
& \multirow{2}{*}{\textbf{Type}} 
& \multirow{2}{*}{\textbf{Method}} 
& \multicolumn{4}{c}{\textbf{Overall Generation Fidelity}} 
& \multicolumn{4}{c}{\textbf{Anomaly Preservation Fidelity}} 
& \multicolumn{2}{c}{\textbf{Downstream Quality}} \\
\cmidrule(lr){4-7} \cmidrule(lr){8-11} \cmidrule(lr){12-13}
& & 
& \textbf{T-Wass.} $\downarrow$
& \textbf{D-Wass.} $\downarrow$
& \textbf{S-Wass.} $\downarrow$
& \textbf{MMD} $\downarrow$
& \textbf{A-Rate.} $\downarrow$
& \textbf{A-Count.} $\downarrow$
& \textbf{A-Energy.} $\downarrow$
& \textbf{A-Tail.} $\downarrow$
& \textbf{Det-PRAUC.} $\uparrow$
& \textbf{Pred-PRAUC.} $\uparrow$ \\
\midrule
\multirow{13}{*}{\makecell{\textbf{FL2}\\2019\\-05}}
& \textbf{GAN}
& TimeGAN (2019)~\cite{yoon2019time}
& 0.0988 & 0.0437 & 0.0751 & 0.8240 & 0.1535 & 10.083 & 1.9939 & 0.0347 & 0.0398 & 0.1771 \\

\arrayrulecolor{gray!60}\cmidrule(lr){2-13}\arrayrulecolor{black}

& \multirow{2}{*}{\textbf{VAE}}
& TimeVAE (2021)~\cite{desai2021timevae}
& 0.1232 & 0.0481 & 0.0883 & 1.1374 & 0.1372 & 8.3885 & 1.6751 & 0.0290 & 0.0475 & 0.1766 \\

&
& koVAE (2024)~\cite{naimangenerative}
& 0.1446 & 0.0632 & 0.1389 & 1.1773 & 0.1680 & 10.108 & 2.0046 & 0.0328 & 0.0462 & 0.1535 \\

\arrayrulecolor{gray!60}\cmidrule(lr){2-13}\arrayrulecolor{black}

& \textbf{Flow}
& F-Flow (2021)~\cite{alaa2021generative}
& 0.2159 & 0.0807 & 0.1974 &  1.3851 & 0.2003 & 12.137  & 2.3341  &  0.0415 & 0.0399 & 0.1312\\

\arrayrulecolor{gray!60}\cmidrule(lr){2-13}\arrayrulecolor{black}

& \multirow{2}{*}{\textbf{Diffusion}}
& DiffWave (2021)~\cite{kongdiffwave}
& 0.1998 & 0.0643 & 0.1547 & 1.0114 & 0.1590 & 6.3996 & 1.5078 & 0.0255 & 0.0501 & 0.1893 \\

&
& Diffusion-TS (2024)~\cite{yuan2024diffusionts}
& 0.0381 & 0.0238 & 0.0375 & \textbf{0.1345} & 0.0464 & 4.6972 & 0.6892 & 0.0240 & \underline{0.0572} & \textbf{0.2421}  \\

\arrayrulecolor{gray!60}\cmidrule(lr){2-13}\arrayrulecolor{black}

& \textbf{LLM}
& SDForger (2025)~\cite{rousseau2025forging}
 & 0.0494 & 0.0200 & 0.0454 & 0.2396 & \underline{0.0394} & 4.6064 & \underline{0.3375} & 0.0190 & 0.0555 & 0.2191 \\

\arrayrulecolor{gray!60}\cmidrule(lr){2-13}\arrayrulecolor{black}

& \multirow{2}{*}{\makecell{\textbf{Anomaly-}\\\textbf{Aware}}}
& FIDE (2024)~\cite{galib2024fide}
& 0.0474 & 0.0253 & 0.4722 & 0.1998 & 0.0453 & 5.1132 & 0.3572 & 0.0298 & \underline{0.0572} & 0.2292 \\

&
& HeavyDiff (2025)~\cite{pandey2025heavy}
& \underline{0.0369} & \textbf{0.0111} & \underline{0.0347} & 0.1576 & 0.0569 & \underline{2.5332} & 0.3823 & \underline{0.0170} & 0.0560 & 0.2253 \\

\arrayrulecolor{gray!60}\cmidrule(lr){2-13}\arrayrulecolor{black}

& \multirow{2}{*}{\makecell{\textbf{Elec.-}\\\textbf{Specific}}}
& Cond-Diff (2024)~\cite{fu2024creating}
& 0.1720 & 0.0589 & 0.0944 & 0.7342 & 0.1158 & 7.7585 & 1.5633 & 0.0289 & 0.0519 & 0.1993 \\

&
& CENTS (2025)~\cite{fuest2025cents}
& 0.0454 & 0.0288 & 0.0433 & 0.2735 & 0.0547 & 5.5358 & 0.4270 & 0.0273 & 0.0538 & 0.2159 \\

\arrayrulecolor{gray!60}\cmidrule(lr){2-13}\arrayrulecolor{black}

& \textbf{Ours}
& \textbf{\m}
& \textbf{0.0209} & \underline{0.0175} & \textbf{0.0182} & \underline{0.1482} & \textbf{0.0268} & \textbf{2.3590} & \textbf{0.3280} & \textbf{0.0126} & \textbf{0.0618} & \underline{0.2411}  \\
\bottomrule
\end{tabular}
\end{table*}
\begin{table*}[!t]
\centering
\footnotesize
\renewcommand{\arraystretch}{1.0}
\setlength{\tabcolsep}{2pt}

\caption{Overall generation fidelity, anomaly preservation fidelity, and downstream quality on the NY dataset. Best results are highlighted in bold, and second-best results are underlined. $\uparrow$ and $\downarrow$ indicate that higher and lower values are better, respectively.}%\vspace{-3pt}
\label{tab:NY}
\begin{tabular}{cclcccccccccc}
\toprule
\multirow{2}{*}{\textbf{Dataset}} 
& \multirow{2}{*}{\textbf{Type}} 
& \multirow{2}{*}{\textbf{Method}} 
& \multicolumn{4}{c}{\textbf{Overall Generation Fidelity}} 
& \multicolumn{4}{c}{\textbf{Anomaly Preservation Fidelity}} 
& \multicolumn{2}{c}{\textbf{Downstream Quality}} \\
\cmidrule(lr){4-7} \cmidrule(lr){8-11} \cmidrule(lr){12-13}
& & 
& \textbf{T-Wass.} $\downarrow$
& \textbf{D-Wass.} $\downarrow$
& \textbf{S-Wass.} $\downarrow$
& \textbf{MMD} $\downarrow$
& \textbf{A-Rate.} $\downarrow$
& \textbf{A-Count.} $\downarrow$
& \textbf{A-Energy.} $\downarrow$
& \textbf{A-Tail.} $\downarrow$
& \textbf{Det-PRAUC.} $\uparrow$
& \textbf{Pred-PRAUC.} $\uparrow$ \\
\midrule
\multirow{13}{*}{\makecell{\textbf{NY}\\2024}}
& \textbf{GAN}
& TimeGAN (2019)~\cite{yoon2019time}
& 0.0636 & 0.0073 & 0.0541 & 0.1174 & 0.1185 & 0.5923 & 0.1328 & 0.0597 & 0.0213 & 0.5834 \\

\arrayrulecolor{gray!60}\cmidrule(lr){2-13}\arrayrulecolor{black}

& \multirow{2}{*}{\textbf{VAE}}
& TimeVAE (2021)~\cite{desai2021timevae}
& 0.0772 & 0.0097 & 0.0643 & 0.1019 & 0.1299 & 0.5192 & 0.1296 & 0.0440 & 0.0258 & 0.5648 \\

&
& koVAE (2024)~\cite{naimangenerative}
& 0.0885 & 0.0082 & 0.0665 & 0.1223 & 0.1551 & 0.5635 & 0.1476 & 0.0603 & 0.0249 & 0.5536 \\

\arrayrulecolor{gray!60}\cmidrule(lr){2-13}\arrayrulecolor{black}

& \textbf{Flow}
& F-Flow (2021)~\cite{alaa2021generative}
& 0.1142 & 0.0089 & 0.0748 & 0.1307 & 0.1771 & 0.5229 & 0.1473 & 0.0558 & 0.0201 & 0.4829 \\

\arrayrulecolor{gray!60}\cmidrule(lr){2-13}\arrayrulecolor{black}

& \multirow{2}{*}{\textbf{Diffusion}}
& DiffWave (2021)~\cite{kongdiffwave}
& 0.0992 & 0.0075 & 0.0642 & 0.0997 & 0.1036 & 0.5587 & 0.1391 & 0.0434 & 0.0216 & 0.5363 \\

&
& Diffusion-TS (2024)~\cite{yuan2024diffusionts}
& \underline{0.0164} & \underline{0.0017} & 0.0262 & 0.0648 &  0.0493 & 0.0786 & \underline{0.0532} & \underline{0.0077} & 0.0278 & \underline{0.6109} \\

\arrayrulecolor{gray!60}\cmidrule(lr){2-13}\arrayrulecolor{black}

& \textbf{LLM}
& SDForger (2025)~\cite{rousseau2025forging}
& 0.0220 & \underline{0.0017} & \textbf{0.0120} & \underline{0.0397} & 0.0475 & 0.2349 & 0.0605 & 0.0293 & \underline{0.0303} & 0.6106 \\

\arrayrulecolor{gray!60}\cmidrule(lr){2-13}\arrayrulecolor{black}

& \multirow{2}{*}{\makecell{\textbf{Anomaly-}\\\textbf{Aware}}}
& FIDE (2024)~\cite{galib2024fide}
& 0.0273 & 0.0025 & 0.0153 & 0.0894 & 0.0655 & 0.3238 & 0.0899 & 0.0284 & 0.0276 & 0.6002 \\

&
& HeavyDiff (2025)~\cite{pandey2025heavy}
& 0.0252 & 0.0027 & 0.0267 & 0.0702 & \underline{0.0306} & \underline{0.0726} & 0.0763 & 0.0194 & 0.0290 & \textbf{0.6153} \\

\arrayrulecolor{gray!60}\cmidrule(lr){2-13}\arrayrulecolor{black}

& \multirow{2}{*}{\makecell{\textbf{Elec.-}\\\textbf{Specific}}}
& Cond-Diff (2024)~\cite{fu2024creating}
& 0.0554 & 0.0072 & 0.0617 & 0.0858 & 0.1667 & 0.4319 & 0.1005 & 0.0334 & 0.0256 & 0.5892  \\
&
& CENTS (2025)~\cite{fuest2025cents}
& 0.0230 & 0.0028 & 0.0278 & 0.0535 & 0.0547 & 0.1893 & 0.0762 & 0.0205 & 0.0280 & 0.5729  \\

\arrayrulecolor{gray!60}\cmidrule(lr){2-13}\arrayrulecolor{black}

& \textbf{Ours}
& \textbf{\m}
& \textbf{0.0148} & \textbf{0.0016} & \underline{0.0136} & \textbf{0.0387} & \textbf{0.0244} & \textbf{0.0696} & \textbf{0.0175} & \textbf{0.0028} & \textbf{0.0365} & \underline{0.6109} \\
\bottomrule
\end{tabular}
\vspace{-5pt}
\end{table*}
\begin{table*}[!t]
\centering
\footnotesize
\renewcommand{\arraystretch}{1.0}
\setlength{\tabcolsep}{2pt}

\caption{Overall generation fidelity, anomaly preservation fidelity, and downstream quality on the CA dataset. Best results are highlighted in bold, and second-best results are underlined. $\uparrow$ and $\downarrow$ indicate that higher and lower values are better, respectively.}%\vspace{-3pt}
\label{tab:CA}
\begin{tabular}{cclcccccccccc}
\toprule
\multirow{2}{*}{\textbf{Dataset}} 
& \multirow{2}{*}{\textbf{Type}} 
& \multirow{2}{*}{\textbf{Method}} 
& \multicolumn{4}{c}{\textbf{Overall Generation Fidelity}} 
& \multicolumn{4}{c}{\textbf{Anomaly Preservation Fidelity}} 
& \multicolumn{2}{c}{\textbf{Downstream Quality}} \\
\cmidrule(lr){4-7} \cmidrule(lr){8-11} \cmidrule(lr){12-13}
& & 
& \textbf{T-Wass.} $\downarrow$
& \textbf{D-Wass.} $\downarrow$
& \textbf{S-Wass.} $\downarrow$
& \textbf{MMD} $\downarrow$
& \textbf{A-Rate.} $\downarrow$
& \textbf{A-Count.} $\downarrow$
& \textbf{A-Energy.} $\downarrow$
& \textbf{A-Tail.} $\downarrow$
& \textbf{Det-PRAUC.} $\uparrow$
& \textbf{Pred-PRAUC.} $\uparrow$ \\
\midrule
\multirow{13}{*}{\makecell{\textbf{CA}\\2024}}
& \textbf{GAN}
& TimeGAN (2019)~\cite{yoon2019time}
& 0.0772 & 0.0058 & 0.1012 & 0.0853 & 0.1541 & 0.2388 & 0.2417 & 0.0445 & 0.0561 & 0.7273 \\

\arrayrulecolor{gray!60}\cmidrule(lr){2-13}\arrayrulecolor{black}

& \multirow{2}{*}{\textbf{VAE}}
& TimeVAE (2021)~\cite{desai2021timevae}
& 0.0801 & 0.0061 & 0.1056 & 0.0929 & 0.1398 & 0.2465 & 0.2799 & 0.0462 & 0.0547 & 0.7198  \\

&
& koVAE (2024)~\cite{naimangenerative}
& 0.0786 & 0.0055 & 0.1138 & 0.0797 & 0.1472 & 0.2513 & 0.2586 & 0.0488 & 0.0594 & 0.7331  \\

\arrayrulecolor{gray!60}\cmidrule(lr){2-13}\arrayrulecolor{black}

& \textbf{Flow}
& F-Flow (2021)~\cite{alaa2021generative}
& 0.0947 & 0.0071 & 0.1295 & 0.1086 & 0.1784 & 0.2968 & 0.3267 & 0.0573 & 0.0478 & 0.6816  \\

\arrayrulecolor{gray!60}\cmidrule(lr){2-13}\arrayrulecolor{black}

& \multirow{2}{*}{\textbf{Diffusion}}
& DiffWave (2021)~\cite{kongdiffwave}
& 0.0814 & 0.0060 & 0.1097 & 0.0838 & 0.1516 & 0.2442 & 0.2674 & 0.0457 & 0.0578 & 0.7286  \\

&
& Diffusion-TS (2024)~\cite{yuan2024diffusionts}
& 0.0213 & 0.0038 & 0.0253 & 0.0266 & 0.0191 & 0.1983 & 0.1748 & 0.0330 & 0.0643 & \underline{0.7626}  \\

\arrayrulecolor{gray!60}\cmidrule(lr){2-13}\arrayrulecolor{black}

& \textbf{LLM}
& SDForger (2025)~\cite{rousseau2025forging}
& 0.0386 & 0.0042 & 0.0456 & 0.0400 & 0.1006 & \underline{0.1021} & 0.2599 & 0.0163 & \textbf{0.0716} & 0.7554 \\

\arrayrulecolor{gray!60}\cmidrule(lr){2-13}\arrayrulecolor{black}

& \multirow{2}{*}{\makecell{\textbf{Anomaly-}\\\textbf{Aware}}}
& FIDE (2024)~\cite{galib2024fide}
& 0.0311 & 0.0040 & 0.0519 & 0.0663 & 0.0826 & 0.2027 & 0.2185 & 0.0244 & 0.0605 & 0.7124 \\

&
& HeavyDiff (2025)~\cite{pandey2025heavy}
& \textbf{0.0193} & \underline{0.0031} & \textbf{0.0174} & \underline{0.0053} & \underline{0.0120} & 0.1333 & \underline{0.1441} & \underline{0.0104} & 0.0626 & 0.7503 \\

\arrayrulecolor{gray!60}\cmidrule(lr){2-13}\arrayrulecolor{black}

& \multirow{2}{*}{\makecell{\textbf{Elec.-}\\\textbf{Specific}}}
& Cond-Diff (2024)~\cite{fu2024creating}
& 0.0787 & 0.0054 & 0.1062 & 0.0708 & 0.1267 & 0.1968 & 0.2579 & 0.0438 & 0.0595 & 0.7356 \\
&
& CENTS (2025)~\cite{fuest2025cents}
& 0.0348 & 0.0044 & 0.0457 & 0.0528 & 0.0713 & 0.1586 & 0.2364 & 0.0201 & 0.0658 & 0.7347  \\

\arrayrulecolor{gray!60}\cmidrule(lr){2-13}\arrayrulecolor{black}

& \textbf{Ours}
& \textbf{\m}
& \underline{0.0154} & \textbf{0.0030} & \underline{0.0185} & \textbf{0.0041} & \textbf{0.0117} & \textbf{0.0993} & \textbf{0.1351} & \textbf{0.0048} & \underline{0.0666} & \textbf{0.7692} \\
\bottomrule
\end{tabular}
\vspace{-5pt}
\end{table*}
\clearpage

\begin{table*}[!t]
\centering
\footnotesize
\renewcommand{\arraystretch}{1.0}
\setlength{\tabcolsep}{1.5pt}

\caption{Scalability analysis on the FL1 dataset across different numbers of household samples.}
\label{tab:scalability_household_number}
\begin{tabular}{cclcccccccccc}
\toprule
\multirow{2}{*}{\makecell{\textbf{Household}\\\textbf{Scale}}}
& \multirow{2}{*}{\textbf{Type}}
& \multirow{2}{*}{\textbf{Method}}
& \multicolumn{4}{c}{\textbf{Overall Generation Fidelity}}
& \multicolumn{4}{c}{\textbf{Anomaly Preservation Fidelity}}
& \multicolumn{2}{c}{\textbf{Downstream Quality}} \\
\cmidrule(lr){4-7}
\cmidrule(lr){8-11}
\cmidrule(lr){12-13}
& &
& \textbf{T-Wass.} $\downarrow$
& \textbf{D-Wass.} $\downarrow$
& \textbf{S-Wass.} $\downarrow$
& \textbf{MMD} $\downarrow$
& \textbf{A-Rate} $\downarrow$
& \textbf{A-Count} $\downarrow$
& \textbf{A-Energy} $\downarrow$
& \textbf{A-Tail} $\downarrow$
& \textbf{Det-PRAUC} $\uparrow$
& \textbf{Pred-PRAUC} $\uparrow$ \\
\midrule

% ==================== 50,000 households ====================
\multirow{4}{*}{\textbf{50,000}}
& \textbf{Diffusion}
& Diffusion-TS (2024)~\cite{yuan2024diffusionts}
& 0.0273
& \underline{0.0124}
& \underline{0.0270}
& \underline{0.1347}
& 0.0318
& 2.1497
& 0.4048
& 0.0189
& 0.0634
& 0.3443 \\

\arrayrulecolor{gray!60}
\cmidrule(lr){2-13}
\arrayrulecolor{black}

& \textbf{LLM}
& SDForger (2025)~\cite{rousseau2025forging}
& 0.0385
& 0.0201
& 0.0382
& 0.2332
& 0.0453
& 4.1887
& 0.4664
& 0.0169
& 0.0655
& 0.3231 \\

\arrayrulecolor{gray!60}
\cmidrule(lr){2-13}
\arrayrulecolor{black}

& \textbf{Anomaly}
& HeavyDiff (2025)~\cite{pandey2025heavy}
& \textbf{0.0201}
& \textbf{0.0114}
& 0.0293
& 0.1508
& \underline{0.0176}
& \underline{2.0125}
& \underline{0.1456}
& \underline{0.0089}
& \underline{0.0658}
& \textbf{0.3837} \\

\arrayrulecolor{gray!60}
\cmidrule(lr){2-13}
\arrayrulecolor{black}

& \textbf{Ours}
& \textbf{\m}
& \underline{0.0275}
& 0.0132
& \textbf{0.0264}
& \textbf{0.1244}
& \textbf{0.0148}
& \textbf{1.9422}
& \textbf{0.1290}
& \textbf{0.0071}
& \textbf{0.0677}
& \underline{0.3516} \\

% ==================== Double separator ====================
\midrule
\midrule

% ==================== 10,000 households ====================
\multirow{4}{*}{\textbf{10,000}}
& \textbf{Diffusion}
& Diffusion-TS (2024)~\cite{yuan2024diffusionts}
& 0.0356
& 0.0181
& 0.0313
& 0.2049
& 0.0427
& 3.9424
& 0.4483
& \underline{0.0118}
& \underline{0.0608}
& 0.3354 \\

\arrayrulecolor{gray!60}
\cmidrule(lr){2-13}
\arrayrulecolor{black}

& \textbf{LLM}
& SDForger (2025)~\cite{rousseau2025forging}
& 0.0386
& 0.0156
& 0.0345
& 0.2200
& 0.0377
& 3.0494
& 0.4218
& 0.0125
& 0.0607
& 0.3106 \\

\arrayrulecolor{gray!60}
\cmidrule(lr){2-13}
\arrayrulecolor{black}

& \textbf{Anomaly}
& HeavyDiff (2025)~\cite{pandey2025heavy}
& \underline{0.0234}
& \textbf{0.0109}
& \textbf{0.0206}
& \underline{0.1662}
& \underline{0.0161}
& \underline{1.8122}
& \underline{0.2631}
& \textbf{0.0107}
& 0.0591
& \underline{0.3411} \\

\arrayrulecolor{gray!60}
\cmidrule(lr){2-13}
\arrayrulecolor{black}

& \textbf{Ours}
& \textbf{\m}
& \textbf{0.0228}
& \underline{0.0122}
& \underline{0.0217}
& \textbf{0.1569}
& \textbf{0.0135}
& \textbf{1.6491}
& \textbf{0.2008}
& \textbf{0.0107}
& \textbf{0.0623}
& \textbf{0.3529} \\

% ==================== Double separator ====================
\midrule
\midrule

% ==================== 1,000 households ====================
\multirow{4}{*}{\textbf{1,000}}
& \textbf{Diffusion}
& Diffusion-TS (2024)~\cite{yuan2024diffusionts}
& 0.0323
& \underline{0.0136}
& 0.0320
& 0.1748
& 0.0333
& 2.5777
& 0.4027
& 0.0132
& 0.0608
& 0.3439 \\

\arrayrulecolor{gray!60}
\cmidrule(lr){2-13}
\arrayrulecolor{black}

& \textbf{LLM}
& SDForger (2025)~\cite{rousseau2025forging}
& 0.0294
& 0.0178
& 0.0290
& \underline{0.1727}
& 0.0482
& 4.6105
& 0.4757
& \underline{0.0127}
& 0.0613
& \textbf{0.3617} \\

\arrayrulecolor{gray!60}
\cmidrule(lr){2-13}
\arrayrulecolor{black}

& \textbf{Anomaly}
& HeavyDiff (2025)~\cite{pandey2025heavy}
& \underline{0.0269}
& 0.0137
& \underline{0.0259}
& 0.1928
& \underline{0.0221}
& \underline{2.2613}
& \textbf{0.1710}
& 0.0158
& \underline{0.0614}
& 0.3319 \\

\arrayrulecolor{gray!60}
\cmidrule(lr){2-13}
\arrayrulecolor{black}

& \textbf{Ours}
& \textbf{\m}
& \textbf{0.0178}
& \textbf{0.0106}
& \textbf{0.0158}
& \textbf{0.1287}
& \textbf{0.0172}
& \textbf{1.3911}
& \underline{0.1937}
& \textbf{0.0078}
& \textbf{0.0619}
& \underline{0.3608} \\

% ==================== Double separator ====================
\midrule
\midrule

% ==================== 100 households ====================
\multirow{4}{*}{\textbf{100}}
& \textbf{Diffusion}
& Diffusion-TS (2024)~\cite{yuan2024diffusionts}
& 0.0568
& 0.0313
& 0.0551
& 0.2148
& 0.0762
& 5.94
& \underline{0.9128}
& \textbf{0.0037}
& 0.1052
& 0.5159 \\

\arrayrulecolor{gray!60}
\cmidrule(lr){2-13}
\arrayrulecolor{black}

& \textbf{LLM}
& SDForger (2025)~\cite{rousseau2025forging}
& 0.0716
& 0.0343
& 0.0709
& 0.2296
& 0.0808
& 4.76
& 1.7518
& 0.0297
& 0.1067
& 0.4990 \\

\arrayrulecolor{gray!60}
\cmidrule(lr){2-13}
\arrayrulecolor{black}

& \textbf{Anomaly}
& HeavyDiff (2025)~\cite{pandey2025heavy}
& \underline{0.0410}
& \textbf{0.0263}
& \underline{0.0247}
& \underline{0.1520}
& \underline{0.0366}
& \underline{2.83}
& 1.0814
& 0.0326
& \underline{0.1215}
& \textbf{0.5661} \\

\arrayrulecolor{gray!60}
\cmidrule(lr){2-13}
\arrayrulecolor{black}

& \textbf{Ours}
& \textbf{\m}
& \textbf{0.0320}
& \underline{0.0266}
& \textbf{0.0209}
& \textbf{0.1265}
& \textbf{0.0149}
& \textbf{1.38}
& \textbf{0.3550}
& \underline{0.0042}
& \textbf{0.1256}
& \underline{0.5360} \\

\bottomrule
\end{tabular}
\end{table*}

\begin{table*}[!h]
\centering
\footnotesize
\renewcommand{\arraystretch}{1.0}
\setlength{\tabcolsep}{1.5pt}

\caption{Temporal granularity analysis on the FL1 dataset using different time intervals.}
\label{tab:temporal_granularity}
\begin{tabular}{cclcccccccccc}
\toprule
\multirow{2}{*}{\makecell{\textbf{Temporal}\\\textbf{Granularity}}}
& \multirow{2}{*}{\textbf{Type}}
& \multirow{2}{*}{\textbf{Method}}
& \multicolumn{4}{c}{\textbf{Overall Generation Fidelity}}
& \multicolumn{4}{c}{\textbf{Anomaly Preservation Fidelity}}
& \multicolumn{2}{c}{\textbf{Downstream Quality}} \\
\cmidrule(lr){4-7}
\cmidrule(lr){8-11}
\cmidrule(lr){12-13}
& &
& \textbf{T-Wass.} $\downarrow$
& \textbf{D-Wass.} $\downarrow$
& \textbf{S-Wass.} $\downarrow$
& \textbf{MMD} $\downarrow$
& \textbf{A-Rate.} $\downarrow$
& \textbf{A-Count.} $\downarrow$
& \textbf{A-Energy.} $\downarrow$
& \textbf{A-Tail.} $\downarrow$
& \textbf{Det-PRAUC.} $\uparrow$
& \textbf{Pred-PRAUC.} $\uparrow$ \\
\midrule

% ==================== One-day interval ====================
\multirow{4}{*}{\textbf{One Day}}
& \textbf{Diffusion}
& Diffusion-TS (2024)~\cite{yuan2024diffusionts}
& \underline{0.0295}
& 0.0242
& \textbf{0.0220}
& 0.0530
& 0.0208
& \textbf{0.1880}
& 0.3274
& 0.0166
& 0.1097
& 0.7323 \\

\arrayrulecolor{gray!60}
\cmidrule(lr){2-13}
\arrayrulecolor{black}

& \textbf{LLM}
& SDForger (2025)~\cite{rousseau2025forging}
& 0.0383
& 0.0183
& 0.0324
& 0.0700
& 0.0386
& 0.5150
& \underline{0.1165}
& \underline{0.0040}
& \underline{0.1118}
& \underline{0.7343} \\

\arrayrulecolor{gray!60}
\cmidrule(lr){2-13}
\arrayrulecolor{black}

& \textbf{Anomaly}
& HeavyDiff (2025)~\cite{pandey2025heavy}
& 0.0300
& \underline{0.0179}
& 0.0254
& \underline{0.0495}
& \underline{0.0170}
& 0.2150
& 0.1260
& 0.0103
& 0.1079
& \textbf{0.7358} \\

\arrayrulecolor{gray!60}
\cmidrule(lr){2-13}
\arrayrulecolor{black}

& \textbf{Ours}
& \textbf{\m}
& \textbf{0.0269}
& \textbf{0.0174}
& \underline{0.0223}
& \textbf{0.0479}
& \textbf{0.0166}
& \underline{0.2120}
& \textbf{0.1102}
& \textbf{0.0025}
& \textbf{0.1179}
& 0.7312 \\
% ==================== Double separator ====================
\midrule
\midrule

% ==================== Four-hours interval ====================
\multirow{4}{*}{\textbf{Four Hours}}
& \textbf{Diffusion}
& Diffusion-TS (2024)~\cite{yuan2024diffusionts}
& 0.0356
& 0.0181
& 0.0313
& 0.2049
& 0.0427
& 3.9424
& 0.4483
& \underline{0.0118}
& \underline{0.0608}
& 0.3354 \\

\arrayrulecolor{gray!60}
\cmidrule(lr){2-13}
\arrayrulecolor{black}

& \textbf{LLM}
& SDForger (2025)~\cite{rousseau2025forging}
& 0.0386
& 0.0156
& 0.0345
& 0.2200
& 0.0377
& 3.0494
& 0.4218
& 0.0125
& 0.0607
& 0.3106 \\

\arrayrulecolor{gray!60}
\cmidrule(lr){2-13}
\arrayrulecolor{black}

& \textbf{Anomaly}
& HeavyDiff (2025)~\cite{pandey2025heavy}
& \underline{0.0234}
& \textbf{0.0109}
& \textbf{0.0206}
& \underline{0.1662}
& \underline{0.0161}
& \underline{1.8122}
& \underline{0.2631}
& \textbf{0.0107}
& 0.0591
& \underline{0.3411} \\

\arrayrulecolor{gray!60}
\cmidrule(lr){2-13}
\arrayrulecolor{black}

& \textbf{Ours}
& \textbf{\m}
& \textbf{0.0228}
& \underline{0.0122}
& \underline{0.0217}
& \textbf{0.1569}
& \textbf{0.0135}
& \textbf{1.6491}
& \textbf{0.2008}
& \textbf{0.0107}
& \textbf{0.0623}
& \textbf{0.3529} \\

% ==================== Double separator ====================
\midrule
\midrule

% ==================== One-hour interval ====================
\multirow{4}{*}{\textbf{One Hour}}
& \textbf{Diffusion}
& Diffusion-TS (2024)~\cite{yuan2024diffusionts}
& 0.0296
& \underline{0.0084}
& 0.0280
& 0.2533
& 0.0152
& 3.2630
& 1.4755
& 0.1150
& 0.0443
& 0.3211 \\

\arrayrulecolor{gray!60}
\cmidrule(lr){2-13}
\arrayrulecolor{black}

& \textbf{LLM}
& SDForger (2025)~\cite{rousseau2025forging}
& 0.0264
& 0.0099
& 0.0257
& \underline{0.2527}
& 0.0136
& 3.5690
& 1.0967
& 0.0384
& 0.0599
& 0.2811 \\

\arrayrulecolor{gray!60}
\cmidrule(lr){2-13}
\arrayrulecolor{black}

& \textbf{Anomaly}
& HeavyDiff (2025)~\cite{pandey2025heavy}
& \underline{0.0167}
& 0.0087
& \underline{0.0157}
& 0.3055
& \textbf{0.0043}
& \underline{2.4480}
& \underline{0.6812}
& \underline{0.0289}
& \textbf{0.0694}
& \underline{0.4295} \\

\arrayrulecolor{gray!60}
\cmidrule(lr){2-13}
\arrayrulecolor{black}

& \textbf{Ours}
& \textbf{\m}
& \textbf{0.0138}
& \textbf{0.0050}
& \textbf{0.0107}
& \textbf{0.1744}
& \underline{0.0065}
& \textbf{1.5550}
& \textbf{0.5247}
& \textbf{0.0257}
& \underline{0.0682}
& \textbf{0.4341} \\

\bottomrule
\end{tabular}
\end{table*}

\begin{table*}[!h]
\centering
\footnotesize
\renewcommand{\arraystretch}{1.0}
\setlength{\tabcolsep}{1.5pt}

\caption{Sensitivity analysis of \m to the anomaly threshold $\delta$ on the FL1 dataset.}
\label{tab:threshold_sensitivity}
\begin{tabular}{cclcccccccccc}
\toprule
\multirow{2}{*}{\makecell{\textbf{Threshold}\\$\boldsymbol{\delta}$}}
& \multirow{2}{*}{\textbf{Type}}
& \multirow{2}{*}{\textbf{Method}}
& \multicolumn{4}{c}{\textbf{Overall Generation Fidelity}}
& \multicolumn{4}{c}{\textbf{Anomaly Preservation Fidelity}}
& \multicolumn{2}{c}{\textbf{Downstream Quality}} \\
\cmidrule(lr){4-7}
\cmidrule(lr){8-11}
\cmidrule(lr){12-13}
& &
& \textbf{T-Wass.} $\downarrow$
& \textbf{D-Wass.} $\downarrow$
& \textbf{S-Wass.} $\downarrow$
& \textbf{MMD} $\downarrow$
& \textbf{A-Rate} $\downarrow$
& \textbf{A-Count} $\downarrow$
& \textbf{A-Energy} $\downarrow$
& \textbf{A-Tail} $\downarrow$
& \textbf{Det-PRAUC} $\uparrow$
& \textbf{Pred-PRAUC} $\uparrow$ \\
\midrule

% ==================== Top 20% ====================

\multirow{4}{*}{
    \makecell{
        \textbf{Top 20\%}\\
        $\boldsymbol{\delta=0.073}$
    }
}
& \textbf{Diffusion}
& Diffusion-TS (2024)~\cite{yuan2024diffusionts}
& \underline{0.0255}
& \underline{0.0130}
& \underline{0.0246}
& \textbf{0.1160}
& 0.0290
& 2.3365
& 0.3812
& 0.0148
& 0.0608
& 0.3152 \\

\arrayrulecolor{gray!60}
\cmidrule(lr){2-13}
\arrayrulecolor{black}

& \textbf{LLM}
& SDForger (2025)~\cite{rousseau2025forging}
& 0.0439
& 0.0214
& 0.0438
& 0.2669
& 0.0508
& 4.7475
& 0.5392
& 0.0174
& \textbf{0.0615}
& 0.3246 \\

\arrayrulecolor{gray!60}
\cmidrule(lr){2-13}
\arrayrulecolor{black}

& \textbf{Anomaly}
& HeavyDiff (2025)~\cite{pandey2025heavy}
& 0.0303
& 0.0131
& 0.0300
& 0.1893
& \underline{0.0162}
& \underline{1.3375}
& \underline{0.2336}
& \textbf{0.0053}
& 0.0589
& \underline{0.3483} \\

\arrayrulecolor{gray!60}
\cmidrule(lr){2-13}
\arrayrulecolor{black}

& \textbf{Ours}
& \textbf{\m}
& \textbf{0.0235}
& \textbf{0.0106}
& \textbf{0.0222}
& \underline{0.1365}
& \textbf{0.0098}
& \textbf{1.0101}
& \textbf{0.2168}
& \underline{0.0087}
& \underline{0.0612}
& \textbf{0.3727} \\

% Double horizontal separator
\midrule
\midrule

% ==================== Top 10% ====================
\multirow{4}{*}{
    \makecell{
        \textbf{Top 10\%}\\
        $\boldsymbol{\delta=0.120}$
    }
}
& \textbf{Diffusion}
& Diffusion-TS (2024)~\cite{yuan2024diffusionts}
& 0.0356
& 0.0181
& 0.0313
& 0.2049
& 0.0427
& 3.9424
& 0.4483
& \underline{0.0118}
& \underline{0.0608}
& 0.3354 \\

\arrayrulecolor{gray!60}
\cmidrule(lr){2-13}
\arrayrulecolor{black}

& \textbf{LLM}
& SDForger (2025)~\cite{rousseau2025forging}
& 0.0386
& 0.0156
& 0.0345
& 0.2200
& 0.0377
& 3.0494
& 0.4218
& 0.0125
& 0.0607
& 0.3106 \\

\arrayrulecolor{gray!60}
\cmidrule(lr){2-13}
\arrayrulecolor{black}

& \textbf{Anomaly}
& HeavyDiff (2025)~\cite{pandey2025heavy}
& \underline{0.0234}
& \textbf{0.0109}
& \textbf{0.0206}
& \underline{0.1662}
& \underline{0.0161}
& \underline{1.8122}
& \underline{0.2631}
& \textbf{0.0107}
& 0.0591
& \underline{0.3411} \\

\arrayrulecolor{gray!60}
\cmidrule(lr){2-13}
\arrayrulecolor{black}

& \textbf{Ours}
& \textbf{\m}
& \textbf{0.0228}
& \underline{0.0122}
& \underline{0.0217}
& \textbf{0.1569}
& \textbf{0.0135}
& \textbf{1.6491}
& \textbf{0.2008}
& \textbf{0.0107}
& \textbf{0.0623}
& \textbf{0.3529} \\

% Double horizontal separator
\midrule
\midrule

% ==================== Top 5% ====================

\multirow{4}{*}{
    \makecell{
        \textbf{Top 5\%}\\
        $\boldsymbol{\delta=0.156}$
    }
}
& \textbf{Diffusion}
& Diffusion-TS (2024)~\cite{yuan2024diffusionts}
& 0.0385
& 0.0148
& 0.0384
& 0.2083
& 0.0408
& 3.2475
& 0.4788
& 0.0179
& 0.0601
& 0.3242 \\

\arrayrulecolor{gray!60}
\cmidrule(lr){2-13}
\arrayrulecolor{black}

& \textbf{LLM}
& SDForger (2025)~\cite{rousseau2025forging}
& 0.0403
& 0.0217
& 0.0402
& 0.2516
& 0.0459
& 4.5615
& 0.4440
& 0.0068
& 0.0621
& \textbf{0.3383} \\

\arrayrulecolor{gray!60}
\cmidrule(lr){2-13}
\arrayrulecolor{black}

& \textbf{Anomaly}
& HeavyDiff (2025)~\cite{pandey2025heavy}
& \underline{0.0285}
& \underline{0.0140}
& \underline{0.0274}
& \underline{0.1819}
& \underline{0.0198}
& \underline{2.1759}
& \underline{0.1914}
& \underline{0.0060}
& \underline{0.0625}
& 0.3238 \\

\arrayrulecolor{gray!60}
\cmidrule(lr){2-13}
\arrayrulecolor{black}

& \textbf{Ours}
& \textbf{\m}
& \textbf{0.0261}
& \textbf{0.0078}
& \textbf{0.0258}
& \textbf{0.1521}
& \textbf{0.0099}
& \textbf{0.6221}
& \textbf{0.1264}
& \textbf{0.0050}
& \textbf{0.0633}
& \underline{0.3339} \\

% Double horizontal separator
\midrule
\midrule

% ==================== Top 1% ====================

\multirow{4}{*}{
    \makecell{
        \textbf{Top 1\%}\\
        $\boldsymbol{\delta=0.279}$
    }
}
& \textbf{Diffusion}
& Diffusion-TS (2024)~\cite{yuan2024diffusionts}
& 0.0416
& \underline{0.0159}
& 0.0415
& 0.2420
& 0.0364
& 3.0055
& 0.3773
& 0.0068
& 0.0612
& 0.3206 \\

\arrayrulecolor{gray!60}
\cmidrule(lr){2-13}
\arrayrulecolor{black}

& \textbf{LLM}
& SDForger (2025)~\cite{rousseau2025forging}
& \textbf{0.0270}
& 0.0172
& \underline{0.0265}
& \underline{0.1534}
& 0.0401
& 3.8225
& 0.4171
& 0.0103
& \textbf{0.0626}
& \textbf{0.3668} \\

\arrayrulecolor{gray!60}
\cmidrule(lr){2-13}
\arrayrulecolor{black}

& \textbf{Anomaly}
& HeavyDiff (2025)~\cite{pandey2025heavy}
& 0.0324
& \textbf{0.0154}
& 0.0321
& 0.2137
& \underline{0.0259}
& \textbf{2.3855}
& \underline{0.2367}
& \underline{0.0047}
& 0.0576
& 0.3054 \\

\arrayrulecolor{gray!60}
\cmidrule(lr){2-13}
\arrayrulecolor{black}

& \textbf{Ours}
& \textbf{\m}
& \underline{0.0277}
& 0.0165
& \textbf{0.0233}
& \textbf{0.1488}
& \textbf{0.0250}
& \underline{2.9263}
& \textbf{0.2156}
& \textbf{0.0045}
& \underline{0.0618}
& \underline{0.3585} \\

\bottomrule
\end{tabular}
\end{table*}

\begin{table*}[!h]
\centering
\footnotesize
\renewcommand{\arraystretch}{1.0}
\setlength{\tabcolsep}{2pt}

\caption{Ablation results of \m after removing HG-ASL (\m w/o HG-ASL) and Anomaly Guidance (\m w/o AG) on the FL1 and FL2 datasets.}
\label{tab:ablation}
\begin{tabular}{clcccccccccc}
\toprule
\multirow{2}{*}{\textbf{Dataset}}
& \multirow{2}{*}{\textbf{Method}}
& \multicolumn{4}{c}{\textbf{Overall Generation Fidelity}}
& \multicolumn{4}{c}{\textbf{Anomaly Preservation Fidelity}}
& \multicolumn{2}{c}{\textbf{Downstream Quality}} \\
\cmidrule(lr){3-6}
\cmidrule(lr){7-10}
\cmidrule(lr){11-12}
&
& \textbf{T-Wass.} $\downarrow$
& \textbf{D-Wass.} $\downarrow$
& \textbf{S-Wass.} $\downarrow$
& \textbf{MMD} $\downarrow$
& \textbf{A-Rate.} $\downarrow$
& \textbf{A-Count.} $\downarrow$
& \textbf{A-Energy.} $\downarrow$
& \textbf{A-Tail.} $\downarrow$
& \textbf{Det-PRAUC.} $\uparrow$
& \textbf{Pred-PRAUC.} $\uparrow$ \\
\midrule

% ==================== FL1 ====================
\multirow{3}{*}{\makecell{\textbf{FL1}\\2018\\-10}}
& \m w/o HG-ASL
& 0.0355
& 0.0189
& 0.0341
& 0.2084
& 0.0422
& 3.5634
& 0.3888
& 0.0125
& 0.0612
& 0.3368 \\

& \m w/o AG
& 0.0371
& 0.0192
& 0.0335
& 0.2185
& 0.0477
& 3.9460
& 0.4582
& 0.0129
& 0.0588
& 0.3361 \\

\arrayrulecolor{gray!60}
\cmidrule(lr){2-12}
\arrayrulecolor{black}

& \textbf{\m}
& \textbf{0.0228}
& \textbf{0.0122}
& \textbf{0.0217}
& \textbf{0.1569}
& \textbf{0.0135}
& \textbf{1.6491}
& \textbf{0.2008}
& \textbf{0.0107}
& \textbf{0.0623}
& \textbf{0.3529} \\

% ==================== Double separator ====================
\midrule
\midrule

% ==================== FL2 ====================
\multirow{3}{*}{\makecell{\textbf{FL2}\\2019\\-05}}
& \m w/o HG-ASL
& 0.0372
& 0.0298
& 0.0351
& 0.1781
& 0.0443
& 4.6271
& 0.6593
& 0.02577
& 0.0594
& 0.2365 \\

& \m w/o AG
& 0.0395
& 0.0272
& 0.0365
& 0.1545
& 0.0453
& 4.5827
& 0.7055
& 0.0290
& 0.0571
& \textbf{0.2418} \\

\arrayrulecolor{gray!60}
\cmidrule(lr){2-12}
\arrayrulecolor{black}

& \textbf{\m}
& \textbf{0.0209}
& \textbf{0.0175}
& \textbf{0.0182}
& \textbf{0.1482}
& \textbf{0.0268}
& \textbf{2.3590}
& \textbf{0.3280}
& \textbf{0.0126}
& \textbf{0.0618}
& 0.2411 \\
\bottomrule
\end{tabular}
\vspace{-5pt}
\end{table*}

\end{document}